%% file: main.tex
\documentclass[runningheads]{llncs}

\usepackage{eccv}

\usepackage{eccvabbrv}

\usepackage{graphicx}
\usepackage{booktabs}
\usepackage{tikz}
\usepackage{comment}
\usepackage{amsmath,amssymb} 
\usepackage{color}
\usepackage{cite}
\usepackage{siunitx}
\usepackage{multirow}
\usepackage{tabularx}
\usepackage{amsmath}
\usepackage{gensymb}
\usepackage{xr}
\usepackage{wrapfig}
\usepackage{makecell}

\usepackage{pgfplots} 
\pgfplotsset{compat=1.18}
\usepgfplotslibrary{units}

\usepackage[accsupp]{axessibility}  

\usepackage{hyperref}

\usepackage{orcidlink}

\begin{document}

\title{SRPose: Two-view Relative Pose Estimation\protect\\ with Sparse Keypoints} 


\author{Rui Yin \inst{1} \and
Yulun Zhang \inst{2} \and
Zherong Pan \inst{3} \and
Jianjun Zhu \inst{1}  \and \\
Cheng Wang \inst{1}  \and
Biao Jia  \inst{1\thanks{indicates corresponding author.}} }

\authorrunning{R.~Yin et al.}

\institute{Hanglok-Tech, China \and 
MoE Key Lab of Artificial Intelligence, Shanghai Jiao Tong University, China \and
LightSpeed Studios, Tencent America}

\maketitle

\begin{abstract}
Two-view pose estimation is essential for map-free visual relocalization and object pose tracking tasks. However, traditional matching methods suffer from time-consuming robust estimators, while deep learning-based pose regressors only cater to camera-to-world pose estimation, lacking generalizability to different image sizes and camera intrinsics. In this paper, we propose SRPose, a sparse keypoint-based framework for two-view relative pose estimation in camera-to-world and object-to-camera scenarios.  SRPose consists of a sparse keypoint detector, an intrinsic-calibration position encoder, and promptable prior knowledge-guided attention layers. Given two RGB images of a fixed scene or a moving object, SRPose estimates the relative camera or 6D object pose transformation. Extensive experiments demonstrate that SRPose achieves competitive or superior performance compared to state-of-the-art methods in terms of accuracy and speed, showing generalizability to both scenarios. It is robust to different image sizes and camera intrinsics, and can be deployed with low computing resources.
Project page: \href{https://frickyinn.github.io/srpose}{https://frickyinn.github.io/srpose}.

\end{abstract}

\vspace{-5mm}
\input{introduction.tex}

\vspace{-5mm}
\input{related.tex}

\vspace{-5mm}
\input{method.tex}

\vspace{-5mm}
\input{experiment.tex}

\vspace{-5mm}
\input{conclusion.tex}

\section*{Acknowledgements}
\vspace{-1mm}
This work was supported by Shanghai Municipal Science and Technology Major Project (2021SHZDZX0102) and the Fundamental Research Funds for the Central Universities.


%
%
\bibliographystyle{splncs04}
\bibliography{reference}

\clearpage

\appendix
\input{appendix}

\end{document}

%% file: introduction.tex
\section{Introduction}
\label{sec:intro}
\vspace{-2mm}

\renewcommand\floatpagefraction{.9}
\renewcommand\topfraction{.9}
\renewcommand\bottomfraction{.9}
\renewcommand\textfraction{.1}
\setcounter{totalnumber}{50}
\setcounter{topnumber}{50}
\setcounter{bottomnumber}{50}

\begin{figure}[htbp]
	\centering
    \begin{tabular}{ccc}
        \begin{subfigure}{0.28\linewidth}
            \centering
            \includegraphics[width=\linewidth]{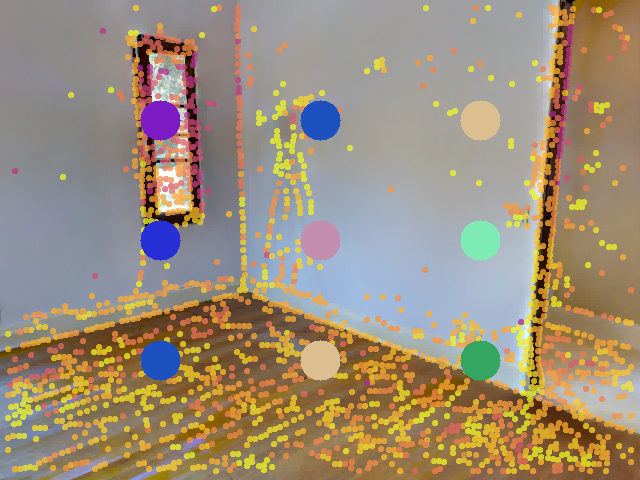}
            \caption{Reference}
            \label{scene-r}
        \end{subfigure} &
        \begin{subfigure}{0.28\linewidth}
            \centering
            \includegraphics[width=\linewidth]{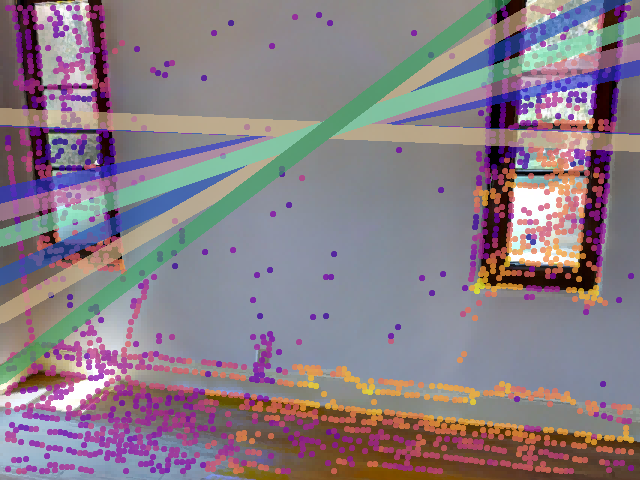}
            \caption{Query}
            \label{scene-q}
        \end{subfigure} & 
        \begin{subfigure}{0.28\linewidth}
            \centering
            \includegraphics[width=\linewidth]{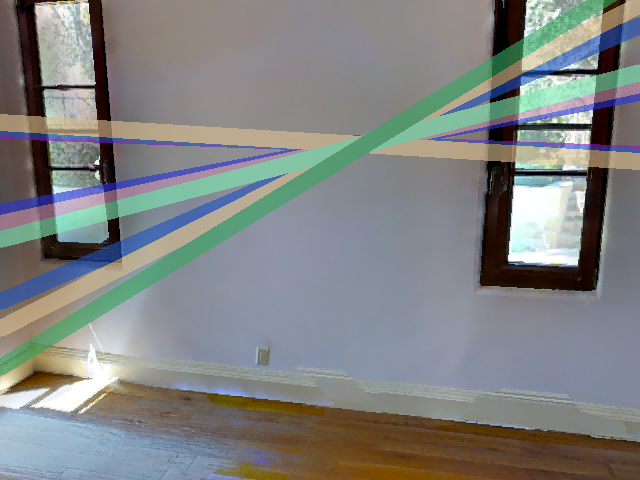}
            \caption{Ground Truth}
            \label{scene-gt}
        \end{subfigure} \\
        \begin{subfigure}{0.28\linewidth}
            \centering
            \includegraphics[width=\linewidth]{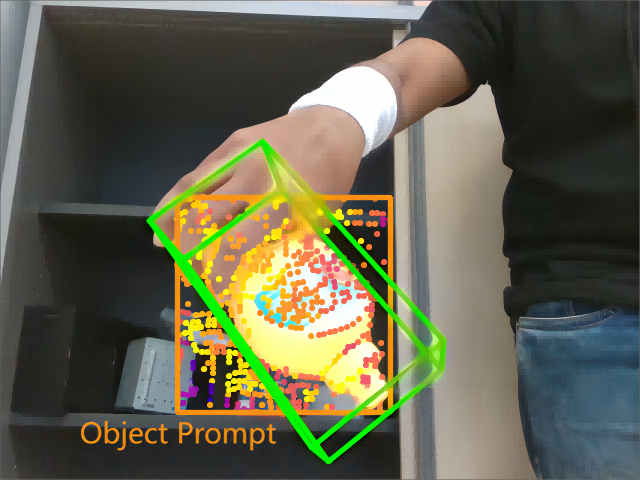}
            \caption{Reference}
            \label{object-r}
        \end{subfigure} & 
        \begin{subfigure}{0.28\linewidth}
            \centering
            \includegraphics[width=\linewidth]{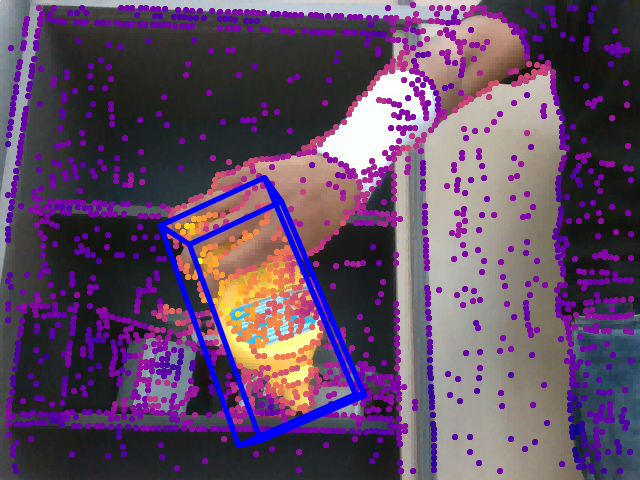}
            \caption{Query}
            \label{object-q}
    	\end{subfigure} & 
        \begin{subfigure}{0.28\linewidth}
            \centering
            \includegraphics[width=\linewidth]{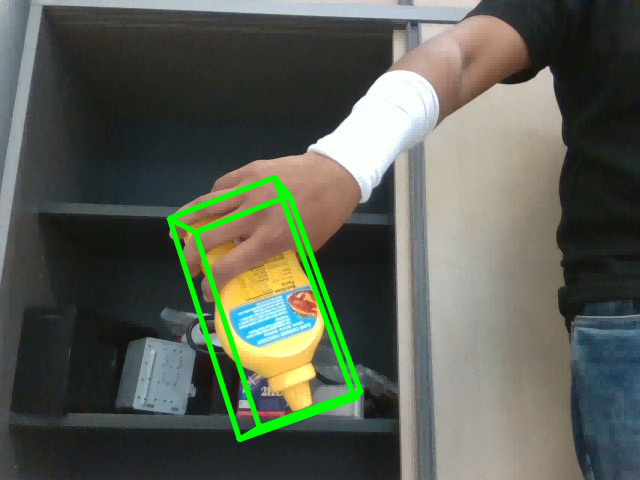}
            \caption{Ground Truth}
            \label{object-gt}
        \end{subfigure}\\
    \end{tabular}
    \vspace{-3.5mm}
    \caption{\textbf{Relative pose estimation by SRPose.} Dots drawn in the figures visualize the cross-attention scores of sparse keypoints across the two images, with brighter dots representing higher attention. \textbf{Camera-to-world}: (a), (b), (c) visualize the epipolar lines, representing the connections between the nine corresponding points across two views. Higher attention is shown to the overlap of the scenes. \textbf{Object-to-Camera}: (d), (e), (f) show the relative 6D pose estimation in the query image with only one accessible object prompt in the reference image. Higher attention is shown to the target object. \textbf{SRPose establishes implicit correspondences.}}
    \label{cover}
    \vspace{-7mm}
\end{figure}

Relative pose estimation between two images plays a crucial role in many 3D vision tasks, including visual odometry and map-free visual relocalization~\cite{arnold2022map}, 6D object pose tracking~\cite{wen2021bundletrack, wen2023bundlesdf}, \etc. These tasks can be categorized into two scenarios: camera-to-world and object-to-camera estimation. In the camera-to-world scenario, such as in map-free visual relocalization, we estimate the pose transformation from the reference image to the query image, predicting the change in camera extrinsics between the two views. In the object-to-camera scenario, taking object pose tracking for example, we aim to track the 6D pose of the target object in the videos by estimating its relative transformation in the camera coordinate system between two adjacent frames. While these two scenarios may seem distinct, they both entail the prediction of the pose matrix from the two RGB images depicting the scene or object, which in turn represents the relative pose transformation from the first image to the second.

Traditional pose estimation approaches typically involve detecting and matching keypoints or local features in the two images to establish point-to-point correspondences, then use a robust estimator to denoise the outliers and recover the relative pose from the essential matrix by solving a two-view geometry problem, \ie, the epipolar constraint equation~\cite{nister2004efficient}. These approaches produce accurate results regardless of image sizes and camera intrinsics, thanks to the intrinsic calibration performed before solving the equation. Given the segmentation of the target object in the two images, these approaches can also yield acceptable predictions of the relative object pose transformation. However, on the downside, robust estimators can be computationally expensive. Indeed, the process of eliminating mismatched correspondences and solving the equations using robust estimators is significantly slower compared to detecting and matching keypoints or features, making it prohibitive in the real-time application. Moreover, real-time object pose tracking requires video object segmentation models to eliminate off-target keypoints \cite{wen2021bundletrack,wen2023bundlesdf}, which can also introduce additional overhead.

The maturity of deep learning offers the advantage of directly regressing the relative pose transformation from two RGB inputs, significantly boosting the runtime performance. However, existing regression methods lack of generalizability or precision to images with varying sizes and camera intrinsics \cite{arnold2022map, rockwell20228, zhou2020learn, wang2023posediffusion, zhang2022relpose, lin2024relpose++}. They typically employ image encoders that only accept fixed-size inputs during batched training. And unlike traditional approaches, most deep learning-based regressors lack awareness of the original camera intrinsic parameters, unable to utilize two-view geometry to achieve higher precision in pose estimation. Furthermore, state-of-the-art deep regressors only apply to camera-to-world pose estimation, despite the common solution shared by both the camera-to-world and object-to-camera tasks.

To address the aforementioned challenges, we propose \textbf{SRPose}: a \textbf{S}parse keypoint-based framework for \textbf{R}elative \textbf{Pose} estimation. SRPose estimates the relative pose matrix based on two-view geometry by implicitly solving the epipolar constraint, as shown in \cref{cover}. First, SRPose employs a point detector to extract sparse keypoints and associated descriptors to form the candidate set of implicit correspondences. Then, we use an intrinsic-calibration (IC) position encoder to modulate the keypoints before it computes the position embeddings, accounting for varying image sizes and camera intrinsics. The keypoint descriptors and the position embeddings are then fused into the prior knowledge-guided attention layers, which leverage the prior knowledge of keypoint similarities to establish implicit cross-view correspondences. With the mechanism of the promptable cross-attention in these layers, SRPose only requires an accessible object prompt in one of the two views to compute the relative 6D pose transformation of the target object, eliminating the need for elusive object segmentations.

The main contribution of this paper can be summarized as follows:
\vspace{-2mm}
\begin{itemize}
    \item We propose a novel framework using sparse keypoints, SRPose, for two-view relative pose estimation. To the best of our knowledge, it is the first attempt to directly regress relative poses from sparse keypoints for this task.
    \item We introduce an intrinsic-calibration position encoder to adapt SRPose to different image sizes and camera intrinsics.
    \item We enable object-to-camera estimation in addition to camera-to-world estimation by utilizing one accessible object prompt.
    \item Our SRPose greatly reduces estimation time by replacing robust estimators with direct regression, achieving state-of-the-art performance in accuracy and speed in camera-to-world and object-to-camera scenarios. 
\end{itemize}

%% file: related.tex
\section{Related works}
\vspace{-3mm}
\subsubsection{Relative Pose Estimation:}
In relative pose estimation, traditional approaches recover the rotation and translation from the essential matrix \cite{longuet1981computer, hartley1997defense}. By using the epipolar constraint~\cite{nister2004efficient}, the essential matrix is estimated based on the cross-view correspondences established through the matching of sparse keypoints or dense features. Using this pipeline, sparse keypoint-based approaches typically start by detecting sparse keypoints and their associated descriptors, including classic hand-crafted detectors~\cite{lowe2004distinctive, bay2006surf, rublee2011orb}, and deep learning-based detectors~\cite{yi2016lift, detone2017toward, detone2018superpoint, dusmanu2019d2, r2d2, bhowmik2020reinforced, tyszkiewicz2020disk, zhao2022alike, Zhao2023ALIKED, liu2019gift, luo2020aslfeat, Gleize_2023_SILK, li2022decoupling}. Recent methods further employ deep learning-based matchers~\cite{sarlin20superglue, lindenberger2023lightglue, chen2021learning, shi2022clustergnn, xue2023imp} to establish keypoint matches, while classical approaches rely on nearest neighbor search. In contrast to using sparse keypoints, dense matchers~\cite{sun2021loftr, tang2022quadtree, chang2023structured, chen2022aspanformer, wang2022matchformer, li2020dual, ni2023pats, xue2023sfd2, yu2023adaptive, edstedt2023dkm, zhu2023pmatch, edstedt2023roma} are detector-free and perform pixel-wise dense feature matching. Once the correspondences are established through matching, a robust estimator such as RANSAC~\cite{fischler1981random} is employed to estimate the essential matrix and recover the relative pose. Overall, while the traditional matcher-based approaches have shown promising results, they still face challenges with their time-consuming robust estimators. Furthermore, dense matchers, while providing exceptional accuracy, may suffer from slow performance due to their resource-intensive design.
\vspace{-5mm}
\subsubsection{Relative Pose Regression:}

Deep learning regressors offer an alternative approach by directly predicting the relative pose using dedicated neural networks, without explicitly establishing correspondences. Two-view relative pose estimation focuses on frame-to-frame pose transformations, aiming at visual odometry and map-free relocalization~\cite{en2018rpnet, melekhov2017relative, winkelbauer2021learning, arnold2022map, abouelnaga2021distillpose, laskar2017camera, balntas2018relocnet}. Leveraging prior knowledge is an important aspect in two-view estimation. For instance, SparsePlanes~\cite{jin2021planar}, PlaneFormer~\cite{cai2021extreme} and NOPE-SAC~\cite{tan2023nope} utilize geometry relations of the planar surfaces to enhance performance. Unfortunately, these approaches rely on image encoders designed for images of fixed sizes and intrinsics, lacking generalizability across different settings. Sparse-view relative pose estimation emphasizes global pose optimization or end-to-end SfM involving multiple views~\cite{sinha2023sparsepose, zhang2022relpose, lin2024relpose++, wang2023posediffusion}. Although these sparse-view methods can be extended to two-view estimation, and may even predict unknown intrinsics, their adaptability to varying camera settings depends on multi-view information. In contrast, SRPose leverages sparse keypoints and two-view geometry to achieve higher precision and generalizability across different image sizes and camera intrinsics.

\vspace{-5mm}
\subsubsection{Object Pose Estimation:}
While most of the aforementioned approaches focus on camera pose estimation in a fixed scene, some tasks, such as 6D object pose tracking, require relative object pose estimation as well. Most frameworks predict the relative object pose using the same pipeline of keypoint correspondences. BundleTrack~\cite{wen2021bundletrack} and BundleSDF~\cite{wen2023bundlesdf} track the object poses in videos by matching keypoints between two frames, and conducting global pose graph optimization. POPE~\cite{fan2023pope} combines advanced foundation models of segmentation~\cite{kirillov2023segment} and matching~\cite{oquab2023dinov2} to enable zero-shot object pose estimation. Robust as these classic methods may be, they require dedicated object masks to focus on the objects, which are provided by other resource-consuming segmentation models. To get rid of object segmentation, OnePose~\cite{sun2022onepose}, OnePose++~\cite{he2022onepose++}, and Gen6D~\cite{liu2022gen6d} rely on multiple views of the object, and ZeroShot~\cite{goodwin2022zero} establishes semantic correspondences to mute the background, and employs depth map to enable pose estimation. Nonetheless, these methods require additional high-quality information, and they restrict the querying view to only accommodate a single object. As a remedy, SRPose employs an accessible object prompt to focus on the target object, enabling effective depth-free and mask-free two-view relative object estimation with superior performance in the object-to-camera scenario. 

%% file: method.tex
\section{Method}
\vspace{-3.5mm}

\renewcommand\floatpagefraction{.9}
\renewcommand\topfraction{.9}
\renewcommand\bottomfraction{.9}
\renewcommand\textfraction{.1}
\setcounter{totalnumber}{50}
\setcounter{topnumber}{50}
\setcounter{bottomnumber}{50}

\begin{figure}[t]
    \centering
    \includegraphics[width=0.95\linewidth]{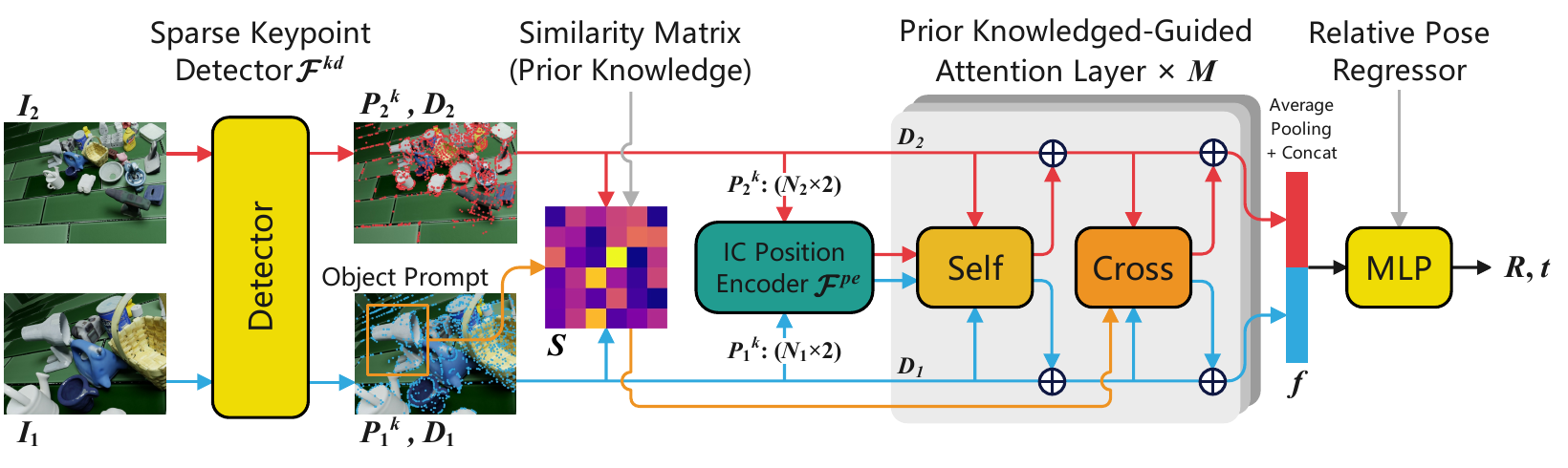}
    \vspace{-3.5mm}
    \caption{\textbf{Overview.} SRPose comprises four main components: \textbf{1)} The sparse keypoint detector detects keypoints associated with descriptors separately from the two images; \textbf{2)} The intrinsic-calibration (IC) position encoder modulates the keypoints' coordinates with camera intrinsics, and encodes their position information; \textbf{3)} Guided by the prior knowledge of keypoint similarities, along with the object prompt, the attention layers establish implicit cross-view correspondences; \textbf{4)} The regressor estimates relative pose \(R, t\) under the constraints of implicit correspondences.}
    \label{fig:overview}
    \vspace{-5mm}
\end{figure}

Given two images \(I_1, I_2\), SRPose estimates the relative pose consisting of a rotation \(R\in\mathcal{SO}(3)\), and a translation \(t\in \mathbb{R}^3\) based on the two-view geometry. Detailed problem definitions can be found in the supplementary materials. Traditional approaches use the epipolar constraint~\cite{nister2004efficient} to estimate the following essential matrix \(E\), from which \(R, t\) are recovered:
\begin{gather}
    (K_1^{-1}q_1)^\top E K_2^{-1}q_2 = 0, \quad\forall(q_1,q_2), \label{epi-con} \\
    E = [t]_{\times} R \label{ess},
\end{gather}
where \(q_1, q_2\) are two arbitrary corresponding points in \(I_1, I_2\), and \(K_1, K_2\) are the camera intrinsics of the two images. SRPose builds and solves the constraint equation~\cref{epi-con} with neural networks implicitly. With SRPose denoted as \(\mathcal{SR}\), the function of our framework is represented as:
\begin{equation}
    [R|t] = \mathcal{SR}(I_1, I_2). \label{lp}
\end{equation}
Figure~\ref{fig:overview} summarizes SRPose, involving four components: a sparse keypoint detector, an intrinsic-calibration position encoder, the promptable prior knowledge-guided attention layers, and an MLP regressor. 

\vspace{-4mm}
\subsection{Sparse Keypoint Detector}
\vspace{-2mm}
Instead of employing an image encoder to extract image features, we use a detector to extract sparse keypoints and associated descriptors from the two images. To this end, we can use one of the existing techniques developed over the years, which include classic methods such as SIFT~\cite{lowe2004distinctive}, and deep learning-based methods such as SuperPoint~\cite{detone2018superpoint}, ALIKE~\cite{zhao2022alike}, DISK~\cite{tyszkiewicz2020disk}, etc. Given an image \(I\), our detector, denoted as \(\mathcal{F}^{kd}\), detects keypoints consisting of \(N\) coordinates \(P^k \subset \mathbb{R}^{2} \) in image space, and associated descriptors \(D \subset \mathbb{R}^{d}\), describing the local features in the image, where \(d\) denotes the dimension of the keypoint descriptors in the network. To summarize, our detector is defined as the following function:
\begin{equation}
    \{P^k, D\} = \mathcal{F}^{kd}(I). \label{keypoint-detector}
\end{equation}
In practice, we resize all images to a standardized size to enable parallel batched detection, then rescale the keypoints' coordinates \(P^k\) to their original positions. The sparse keypoint detector yields two sets of keypoints \(P_1^k, P_2^k\) from \(I_1, I_2\), from which a set of correspondences \(\{(q_1, q_2)\} \subset P_1^k \times P_2^k\) will be established implicitly later for solving the constraint \cref{epi-con}. 

\vspace{-4mm}
\subsection{Intrinsic-Calibration (IC) Position Encoder}
\vspace{-2mm}
According to \cref{epi-con}, for the keypoints detected from an image \(I\), their coordinates should be first calibrated or normalized by the camera intrinsics \(K\):
\begin{equation}
    P^c = K^{-1}P^k. \label{in} \\
\end{equation}
Enforcing the intrinsic calibration (IC) offers a crucial improvement of both the accuracy and robustness \cite{nister2004efficient}. IC allows SRPose to adapt images captured by different cameras with varying intrinsic parameters. Previous regressors~\cite{arnold2022map, rockwell20228} extract implicit image features with image encoders trained on fixed-size images, assuming the camera intrinsics of all inputs are the same by default. While the sparse keypoints in SRPose are modulated by intrinsic calibration, which provides the accurate positions, \ie \(P^c\), of all keypoints in a unified camera coordinate system across different intrinsics. 

After the intrinsic calibration, the keypoints' coordinates are encoded to address the keypoints based on their positions. Due to the possible position distortion caused by relative position encoding \cite{lindenberger2023lightglue, sun2021loftr}, we employ absolute position encoding that maintains the precise coordinate values, which is crucial to the two-view geometry problems. With a single-layer fully connected network, denoted as the following function \(\mathcal{F}^{pe}\), the position encoder maps 2D calibrated coordinates to high-dimension position embeddings:
\begin{equation}
    P^{e} = \mathcal{F}^{pe}(K^{-1}P^k). \label{pos-enc}
\end{equation}
The resulting encoded position embeddings, denoted as \(P^{e}\), are then incorporated with the keypoint descriptors and descriptor embeddings, denoted as \(D \oplus P^{e}\), which provides calibrated and normalized position information for solving the constraint~\cref{epi-con} using a set of attention layers.

\vspace{-4mm}
\subsection{Promptable Prior-Knowledge-Guided Attention}
\begin{figure}[t]
    \centering
    \includegraphics[width=0.95\linewidth]{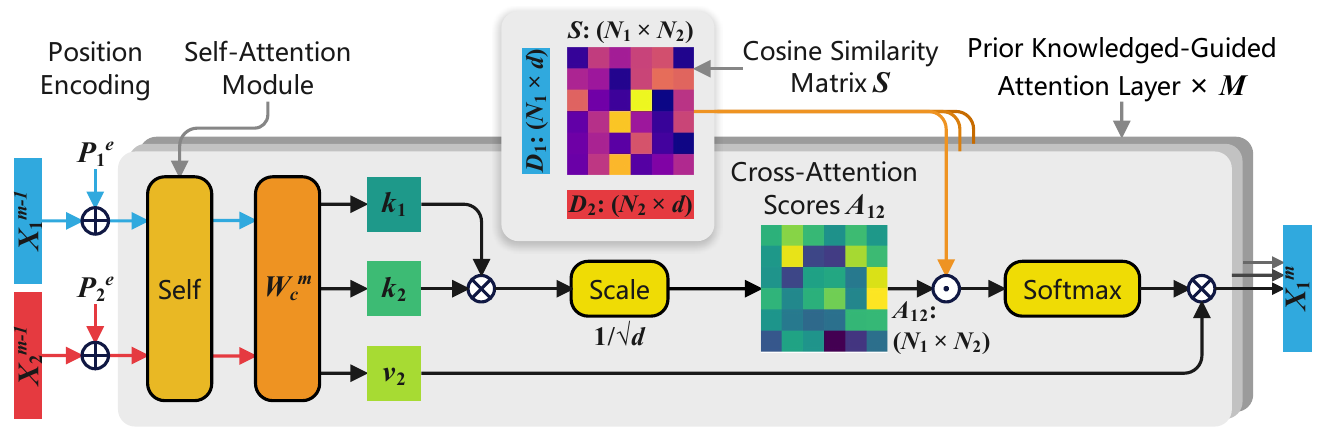}
    \vspace{-2mm}
    \caption{\textbf{Overview of the prior knowledge-guided attention layers.} Each layer contains a self-attention and a cross-attention module. A similarity matrix \(S\) of the keypoint descriptors is utilized as the prior knowledge to readjust the cross-attention scores, guiding more attention to cross-view keypoint pairs with implicit correspondences. The object prompt and the residual connections are omitted in the figure.}
    \label{fig:pkga}
    \vspace{-6mm}
\end{figure}

\vspace{-2mm}
Given the position embeddings and keypoint descriptors, we use a multi-layer attention network to exploit semantic information and establish implicit correspondences. Our network consists of \(M\) consecutive attention layers, with each layer comprising a self-attention module and a promptable prior knowledge-guided cross-attention module. We denote \(X^m\subset \mathbb{R}^{d} \) as the keypoint descriptor embeddings being processed in the \(m^{th}\) layer, in which \(m \in M\), with \(X^m\) being either \(X_1^m\) or \(X_2^m\) from the image \(I_1\) or \(I_2\). Each layer takes \(X^{m-1}\) as inputs, then computes the deeper embeddings \(X^m\). In summary, the entire multi-layer network inputs keypoint descriptors \(D_1\, D_2\), and outputs \(X_1^M, X_2^M\) after \(M\) layers, i.e., we define \(D\) as the initial state of embeddings \(X^0\). Prior to each attention layer, \(X_1^{m-1}\), \(X_2^{m-1}\) are incorporated with the associated position embeddings \(P_1^{e}\), \(P_2^{e}\). Following \cite{he2016deep}, residual connections are applied to every self- and cross-attention module. The architecture of the attention layers is outlined in~\cref{fig:pkga}.

\vspace{-5mm}
\subsubsection{Self-Attention:}
The self-attention module captures the comprehensive semantic information of all keypoints within an entire image. In this module, \(X^{m-1} \oplus P^{e} \) of both the images is first mapped to three vectors by a linear transformation \(W_s^m\), denoted as the query \(q\), the key \(k\), and value \(v\) respectively, where \(q, k, v \subset \mathbb{R}^{d} \). Then \(k\) and \(q\) are used to compute the self-attention scores \(A_s\), which are then scaled by the dimension of descriptors and embeddings \(d\):
\begin{gather}
    [q | k | v] = W_s^M (X^{m-1} \oplus P^{e}), \label{sa-qkv} \\
    A_s = \frac{q k^\top}{\sqrt{d}}. \label{sa-a}
\end{gather}
After being normalized by the \(\mathrm{Softmax}\) function, the self-attention scores \(A_s\) are then used to extract the highly attended information from \(v\):
\begin{equation}
    X^{m-1}_s = \mathrm{Softmax}(A_s)v. \label{sa-softmax}
\end{equation}
In the self-attention module, keypoint position embeddings and descriptor embeddings are computed together to summarize the semantic information within one entire image before deeper cross-view correspondences are established.

\vspace{-5mm}
\subsubsection{Prior Knowledge-guided Cross-Attention:}
Guided by the prior knowledge of keypoint similarities, the cross-attention module exploits the mutual information across the two images to establish implicit correspondences. Similar to the self-attention, cross-attention maps the input embeddings \(X_s^{m-1}\) to two keys \(k_1, k_2\) and values \(v_1, v_2\) of the two images respectively through another linear transformation \(W_c^m\) layer. Following~\cite{wang2022bidirectional}, we compute the bidirectional attention scores across the two images from the resulting embeddings of the previous self-attention module, using the cross-view keys \(k_1, k_2\):
\begin{gather}
    [k_1 | v_1] = W_c^m X_{s, 1}^{m-1}, \label{ca-qkv1} \\
    [k_2 | v_2] = W_c^m X_{s, 2}^{m-1}, \label{ca-qkv2} \\
    A_{21}^\top = A_{12} = \frac{k_1 k_2^\top}{\sqrt{d}}. \label{ca-a} 
\end{gather}
The module recognizes the overlapping areas of the scene or object between the two views by identifying the mutuality between two sets of keypoint embeddings from the computed cross-attention scores. To facilitate the process, prior knowledge is used to guide and enhance the attention on highly correlated keypoints, by adjusting the cross-attention scores with a keypoint similarity matrix \(S\in \mathbb{R}^{N_1 \times N_2} \). Each entry \(s_{ij}\) of the matrix is computed as the cosine similarity between the descriptors of every cross-view keypoint pair:
\begin{gather}
    s_{ij} = (\frac{d_i \cdot d_j }{\| d_i \| \| d_j \|} + 1) / 2, \quad \{(d_i, d_j)\} \subseteq D_1 \times D_2. \label{sim} 
\end{gather}
Keypoints with similar local features tend to possess similar descriptors, describing correlated semantic information of similar views. Thus, the cosine similarity captures the semantic similarities between two keypoints, and guides the framework to recognize the overlapping parts between the images.

We normalize the matrix \(S\) to \([0,1]\) and element-wise multiply it with the cross-attention scores. 
Finally, the scores are used to extract and exchange crucial cross-view information from each others' values \(v_1, v_2\):
\begin{gather}
    A_{21}'^\top = A_{12}' = A_{12} \cdot S, \label{sim-guide} \\
    X^{m-1}_1 = \mathrm{Softmax}(A_{12}')v_2, \label{ca_softmax1} \\
    X^{m-1}_2 = \mathrm{Softmax}(A_{21}')v_1. \label{ca_softmax2}
\end{gather}
By capturing and incorporating relevant information from each image into the other, the implicit correspondences are established across the two images. As shown in Figs.~\ref{scene-r} and \ref{scene-q}, with brighter dots representing higher cross-attention scores, SRPose establishes implicitly corresponding points densely in the overlapping areas of the two images, which is utilized to construct and solve an implicit version of the epipolar constraint~\cref{epi-con}.

\vspace{-5mm}
\subsubsection{Object Prompt:}
In the object-to-camera estimation, we employ an accessible user-provided object prompt in only one image to identify the target object. In contrast, traditional matching approaches require high-quality object segmentations in both images. An object prompt \(b\) is a bounding box consisting of the two coordinates of top-left and bottom-right points, which bounds the object \(o\) in one of the images \(I_1\), making \(I_1\) as the reference view. Once the keypoints are extracted from \(I_1\), the object prompt removes the keypoints outside of the bounding box in the reference view, eliminating most of the irrelevant keypoints lying on the background or other objects. Thus, the descriptors and positions of keypoints within bounds, focusing on the target object \(o\), will be later processed in the Similarity Matrix and other following modules. With the guided cross-attention modules, SRPose learns to identify highly correlated keypoints lying on the same object \(o\) from the entire query view \(I_2\), excluding irrelevant background keypoints, as shown in Figs.~\ref{object-r} and~\ref{object-q}. In practical implementation, to facilitate parallel computing, all sets of keypoints within a batch are padded to the same dimension after various numbers of keypoints are removed from different images. With the object prompt, SRPose searches the implicit corresponding points on the target object and computes the relative 6D object pose transformation in the same process as the camera-to-world scenario.

\vspace{-4mm}
\subsection{Relative Pose Regressor}
\vspace{-2mm}
After the embeddings \(X_1^M, X_2^M\) are computed via the previous modules, a Multi-Layer Perceptron (MLP) is employed to regress the relative pose as our last step. The outputs of the previous attention layers \(X_1^M, X_2^M\) first undergo 2D average pooling, and then are concatenated together, resulting in the joint embedding \(f \in \mathbb{R}^{2d} \), which carries the deep features and implicit correspondences across the two images. Finally, embedding \(f\) is input to the MLP, and the regressor solves the epipolar constraint~\cref{epi-con} implicitly and regresses the relative pose, consisting of a rotation \(R\), and a translation \(t\). Following~\cite{zhou2019continuity}, we set the rotation output of the regressor to be a 6d vector \(r \in \mathbb{R}^6 \), which is then transformed into a rotation matrix \(R \in \mathbb{R}^{3\times3} \) through a partial Gram-Schmidt process. This technique, as argued in~\cite{zhou2019continuity}, enables a continuous representation of rotations.

\vspace{-4mm}
\subsection{Loss Function}
\vspace{-2mm}
SRPose is trained in a supervised manner. With the ground-truth rotation \(R_{gt}\), we compute the Huber Loss~\cite{huber1964robust} of the error in the rotation angle, which minimizes the angular error between the predicted and ground-truth rotations:
\begin{equation}
    \mathcal{L}_R(I_1, I_2) = \mathcal{H} \left( \arccos{ \left( \frac{\mathrm{Tr}(R^\top R_{gt} - 1)}{2} \right) } \right), \label{rot-loss}
\end{equation}
where \(\mathcal{H}\) denotes the Huber Loss function. Similarly, given the ground-truth translation \(t_{gt}\), the error is computed in both normalized and unnormalized forms. To further enhance accuracy, we also incorporate the angular error of translation, resulting in the following three loss terms:
\begin{gather}
    \mathcal{L}_t(I_1, I_2) = \mathcal{H}(t - t_{gt}), \label{t-loss} \\
    \mathcal{L}_{t_n}(I_1, I_2) = \mathcal{H} \left( \frac{t}{\|t\|} - \frac{t_{gt}}{\| t_{gt} \|} \right), \label{tn-loss} \\
    \mathcal{L}_{t_a}(I_1, I_2) = \mathcal{H} \left( \arccos{ \left( \frac{t \cdot t_{gt}}{\| t \| \|t_{gt} \|} \right) } \right). \label{ta_loss}
\end{gather}
The above three loss functions supervise both the scale and direction of translations. 
We compute the final loss by the weighted sum of the three functions:
\begin{equation}
    \mathcal{L}(I_1, I_2) = \mathcal{L}_R + \lambda_t \mathcal{L}_t + \lambda_{t_n} \mathcal{L}_{t_n} + \lambda_{t_a} \mathcal{L}_{t_a}, \label{final-loss}
\end{equation}
where \(\lambda_t, \lambda_{t_n}, \lambda_{t_a} \) are scalars used to balance the weights of four different losses.

\vspace{-4mm}
\subsection{Implementation Details}
\vspace{-2mm}
SRPose is first trained on ScanNet~\cite{dai2017scannet} and then fine-tuned on other datasets to achieve better performance.  For all datasets, we resize the images to 640\(\times\)480, and extract 1,024 sparse keypoints using SuperPoint~\cite{detone2018superpoint} for training and 2,048 for evaluation. Other sparse keypoint detectors can also be used for this step. We utilize the pre-trained SuperPoint provided by the official source and freeze its weights in the training stage. The remaining learnable modules are initialized with random weights. In practice, SRPose consists of 6 guided attention layers, with 4 attention heads in each attention module. The dimension \(d\) of keypoint descriptors and embeddings is set to 256. The rotation and translation are estimated using two separate 3-layer MLPs. All three balancing scalars \(\lambda_t, \lambda_{t_n}, \lambda_{t_a}\) are set to 1. On ScanNet, our framework is trained using AdamW optimizer~\cite{loshchilov2018decoupled} and follows 1cycle learning rate policy~\cite{smith2019super} with a maximum learning rate of \(1 \times 10^{-4}\) for 500 epochs. We fix the batch size to 32 for training. The pre-training requires 120 hours utilizing 8 RTX 3090 GPUs. More implementation details can be found in the supplementary materials.

%% file: experiment.tex
\section{Experiments}
\vspace{-3mm}

We evaluate the performance of SRPose for two-view relative pose estimation in both the camera-to-world and object-to-camera scenarios.
Additionally, we assess its effectiveness in the application of map-free relocalization.
Finally, we conduct an ablation study to analyze the roles of the components we proposed.

\vspace{-4mm}
\subsection{Camera-to-World Pose Estimation}
\vspace{-2mm}
\subsubsection{Setup:}
In the camera-to-world scenario, we evaluate our framework's performance on Matterport~\cite{chang2017matterport3d} and ScanNet~\cite{dai2017scannet}. 
Following \cite{jin2021planar}, for Matterport, we report the median and mean of the translation and rotation error, and the translation/rotation accuracy at the thresholds of 1m/\ang{30}, respectively.
For ScanNet, we adopt the metrics from \cite{sarlin20superglue} to compute the AUC of the pose error at \ang{5}, \ang{10}, and \ang{20}. The pose error here is the maximum angular error in rotation and translation. 
We also report an analysis of the time consumption on the ScanNet-1500 test set.
For all matcher-based approaches, we use the implementation of RANSAC~\cite{fischler1981random} from OpenCV~\cite{opencv_library} to recover relative poses.
All methods are compared on a device with RTX 4090, i9-13900K, and 128GB memory.

\vspace{-5mm}
\subsubsection{Baselines:}
\label{sec:scene-baselines}
We choose three categories of approaches as our baselines, including sparse keypoint-based matchers, dense matchers, and deep learning-based relative pose regressors.
Results of all the baselines are reported according to the official papers or implementations.
All sparse matchers and SRPose employ SuperPoint~\cite{detone2018superpoint} as their sparse keypoint detector.
All regressors are trained on the specific dataset, \ie Matterport or ScanNet, before evaluation.
LightGlue~\cite{lindenberger2023lightglue} is trained on MegaDepth~\cite{MDLi18}, while all other matchers are trained on ScanNet.

\begin{table}[t]
    \caption{\textbf{Relative Pose Estimation on Matterport~\cite{chang2017matterport3d}.} 
    Without depth, matcher-based approaches are incapable of scaled translation estimation, while SRPose achieves high accuracy in translation scales using only RGB inputs.}
    \vspace{-3mm}
    \label{tab:matterport3d}
    \centering
    \scriptsize
    \tabcolsep=4pt
    \begin{tabular}{llcccccc}
    \toprule
        \multirow{2}{*}{Category} & \multirow{2}{*}{Method} & \multicolumn{3}{c}{Rot. (\(^\circ\))}&  \multicolumn{3}{c}{Trans. (m)}\\
         &  &  Med.\(\downarrow\)&  Avg.\(\downarrow\)&  \(\leq\ang{30}\uparrow\)&  Med.\(\downarrow\)&  Avg.\(\downarrow\)& \(\leq1\)m\(\uparrow\)\\
    \midrule
        \multirow{3}{*}{Sparse} &  SuperGlue~\cite{sarlin20superglue}&  3.88&  24.17&  77.8&  n/a&  n/a& n/a\\
         &  SGMNet~\cite{chen2021learning}&  \textbf{1.06}&  \textbf{21.32}&  \textbf{80.9}&  n/a&  n/a& n/a\\
         &  LightGlue~\cite{lindenberger2023lightglue}&  1.32&  22.45&  80.0&  n/a&  n/a& n/a\\
    \midrule
        \multirow{3}{*}{Dense} &  LoFTR~\cite{sun2021loftr}&  0.71&  \textbf{11.11}&  \textbf{90.5}&  n/a&  n/a& n/a\\
         &  ASpanFormer~\cite{chen2022aspanformer}&  3.73&  31.45&  75.7&  n/a&  n/a& n/a\\
         &  DKM~\cite{edstedt2023dkm}&  \textbf{0.46}&  12.89&  89.0&  n/a&  n/a& n/a\\
    \midrule
        \multirow{4}{*}{Regressor} &  SparsePlanes~\cite{jin2021planar}&  7.33&  22.78&  83.4&  0.63&  1.25& 66.6\\
         &  PlaneFormers~\cite{agarwala2022planeformers}&  5.96&  22.20&  83.8&  0.66&  1.19& 66.8\\
         &  8point~\cite{rockwell20228}&  8.01&  19.13&  85.4&  0.64&  1.01& 67.4\\
         &  NOPE-SAC~\cite{tan2023nope}&  2.77&  14.37&  89.0&  0.52&  0.94& 73.2\\
         &  SRPose&  \textbf{2.65}&  \textbf{11.12}&  \textbf{91.6}&  \textbf{0.27}&  \textbf{0.61}& \textbf{83.7}\\
    \bottomrule
    \end{tabular}
    \vspace{-1.5mm}
\end{table}

\begin{table}[t]
    \caption{\textbf{Relative Pose Estimation on ScanNet~\cite{dai2017scannet}.} 
    SRPose outperforms state-of-the-art regressors and sparse matchers in terms of various metrics and reduces computational time greatly due to the riddance of robust estimators.}
    \vspace{-3mm}
    \label{tab:scannet}
    \centering
    \scriptsize
    \tabcolsep=8pt
    \begin{tabular}{llccccc}
    \toprule
        \multirow{2}{*}{Category} & \multirow{2}{*}{Method} &  \multicolumn{3}{c}{Pose estimation AUC}& \multicolumn{2}{c}{Time (ms)}\\
         &  &  @\ang{5}&  @\ang{10}&  @\ang{20}&  Match&  Total\\
    \midrule
        \multirow{3}{*}{Sparse} &  SuperGlue~\cite{sarlin20superglue}&  16.2&  \textbf{33.8}&  \textbf{51.8}&  102.5 &  361.9 \\
         &  SGMNet~\cite{chen2021learning}&  15.4&  32.1&  48.3&  111.7&  439.7 \\
         &  LightGlue~\cite{lindenberger2023lightglue}&  \textbf{16.4}&  33.6&  50.2&  31.5&  269.7  \\
    \midrule
        \multirow{3}{*}{Dense} &  LoFTR~\cite{sun2021loftr}&  22.0&  40.8&  57.6&  44.4&  292.4 \\
         &  ASpanFormer~\cite{chen2022aspanformer}&  25.6&  46.0&  63.3&  59.6 &  370.9 \\
         &  DKM~\cite{edstedt2023dkm}&  \textbf{29.4}&  \textbf{50.7}&  \textbf{68.3}&  235.0&  491.6  \\
    \midrule
        \multirow{2}{*}{Regressor} &  Map-free~\cite{arnold2022map}&  2.70&  11.5&  29.0&  n/a&  12.7\\
         &  SRPose&  \textbf{13.3}&  \textbf{34.3}&  \textbf{56.8}&  n/a&  28.5 \\
    \bottomrule
    \end{tabular}
    \vspace{-1.5mm}
\end{table}

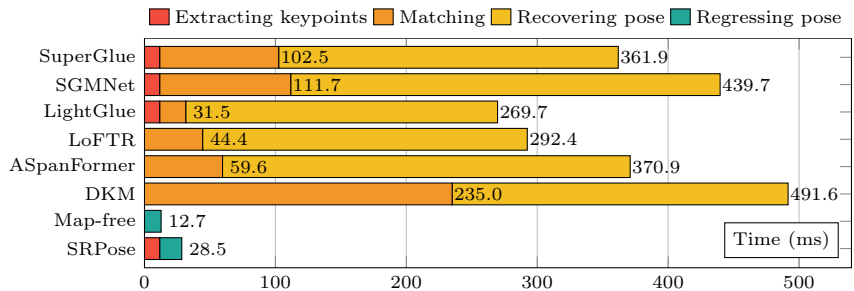
\begin{figure}
    \definecolor{ext}{RGB}{242,76,61}
    \definecolor{mat}{RGB}{242,151,39}
    \definecolor{rec}{RGB}{242,190,34}
    \definecolor{reg}{RGB}{34,166,153}
    
    \pgfplotstableread[row sep=\\,col sep=&]{
        Method & Extracting & Matching & Recovering & Regressing \\
        SuperGlue & 11.6 & 90.9 & 259.4 & 0 \\
        SGMNet & 11.6 & 100.1 & 328 & 0 \\
        LightGlue & 11.6 & 19.9 & 238.2 & 0 \\
        LoFTR & 0 & 44.4 & 248 & 0 \\
        ASpanFormer & 0 & 59.6 & 311.4 & 0 \\
        DKM & 0 & 235 & 256.6 & 0 \\
        Map-free & 0 & 0 & 0 & 12.7 \\
        SRPose & 11.6 & 0 & 0 & 16.9 \\
    }\timedata
    
    \centering
    \begin{tikzpicture}
        \tikzstyle{every node}=[font=\scriptsize]
        \begin{axis}[
            width=0.9\linewidth,
            height=0.38\linewidth,
            bar width=2.9mm,
            xbar stacked, 
            ytick=data,
            xmin=0,
            xmax=540,
            symbolic y coords={SRPose, Map-free, DKM, ASpanFormer, LoFTR, LightGlue, SGMNet, SuperGlue},
            xmajorgrids=true,
            legend columns=4,
            x label style={
                fill=white,
                draw=black,
                at={(0.82, 0.12)},
                anchor=west,
            },
            legend style={
                at={(0.5,1.0)},
                anchor=south,
                draw=none,
            },
            xlabel=Time (ms)
        ]
            \node [coordinate,pin=center:{102.5}] at (axis cs:122.5,SuperGlue) {};
            \node [coordinate,pin=center:{111.7}] at (axis cs:131.7,SGMNet) {};
            \node [coordinate,pin=center:{31.5}] at (axis cs:51.5,LightGlue) {};
            
            \node [coordinate,pin=center:{44.4}] at (axis cs:64.4,LoFTR) {};
            \node [coordinate,pin=center:{59.6}] at (axis cs:79.6,ASpanFormer) {};
            \node [coordinate,pin=center:{235.0}] at (axis cs:255.0,DKM) {};

            \node [coordinate,pin=center:{361.9}] at (axis cs:381.9,SuperGlue) {};
            \node [coordinate,pin=center:{439.7}] at (axis cs:459.7,SGMNet) {};
            \node [coordinate,pin=center:{269.7}] at (axis cs:289.7,LightGlue) {};
            
            \node [coordinate,pin=center:{292.4}] at (axis cs:312.4,LoFTR) {};
            \node [coordinate,pin=center:{370.9}] at (axis cs:390.9,ASpanFormer) {};
            \node [coordinate,pin=center:{491.6}] at (axis cs:511.6,DKM) {};
            
            \node [coordinate,pin=center:{12.7}] at (axis cs:32.7,Map-free) {};
            \node [coordinate,pin=center:{28.5}] at (axis cs:48.5,SRPose) {};
                
            \addplot+ [fill=ext, draw=black] table[x=Extracting, y=Method] {\timedata};
            \addplot+ [fill=mat, draw=black] table[x=Matching, y=Method] {\timedata};
            \addplot+ [fill=rec, draw=black] table[x=Recovering, y=Method] {\timedata};
            \addplot+ [fill=reg, draw=black] table[x=Regressing, y=Method] {\timedata};
            
            \legend{Extracting keypoints, Matching, Recovering pose, Regressing pose}
        \end{axis}
    \end{tikzpicture}
    \vspace{-3mm}
    \caption{\textbf{Time consumption comparison on ScanNet~\cite{dai2017scannet}.} Regressors, including SRPose, achieve much higher computational efficacy than all matchers.}
    \label{fig:time}
    \vspace{-4mm}
\end{figure}

\vspace{-5mm}
\subsubsection{Results:}
Table~\ref{tab:matterport3d} demonstrates that SRPose greatly outperforms the existing deep learning-based regressors across all metrics on Matterport. Our framework achieves the most accurate scaled translation estimation, which is unattainable for matcher-based approaches lacking depth information.
Table~\ref{tab:scannet} shows SRPose achieves competitive or superior performance compared to all sparse matchers and the regressor and on ScanNet.
Although dense matchers outperform our framework in accuracy, SRPose offers a significant advantage in computational speed compared to all the matcher-based approaches by replacing robust estimators with direct regression. As shown in \cref{fig:time}, matcher-based approaches spend a substantial proportion of time on recovering pose using robust estimators after matching.
In contrast, SRPose consumes even less time than the matching stage of all matchers, saving at least 200ms in the total time.

\vspace{-4mm}
\subsection{Object-to-Camera Pose Estimation}
\vspace{-2mm}
\subsubsection{Setup:}
In the object-to-camera scenario, we evaluate SRPose on HO3D~\cite{hampali2020honnotate}.
HO3D contains videos of hand-holding objects of 21 categories from the YCB dataset~\cite{calli2015ycb}. 
The depth captured by the stereo depth camera and the object segmentation of each frame in the videos are also provided.
We randomly select image pairs with small object rotation transformations for training and evaluation.
More details can be found in the supplementary materials.
Following \cite{hinterstoisser2013model, xiang2018posecnn}, we adopt the widely-used Average Distance (ADD, ADD-S) metric for evaluation.
To measure the smaller scales in object pose transformation, we set the threshold of metrics to 10cm, which differs from the 1m threshold used in the camera-to-world scenario.
Specifically, we calculate the AUC using both ADD and ADD-S with the threshold set to 10cm. We also report the median, mean error of translation/rotation, and their accuracy at 10cm/\ang{30}.

\vspace{-4mm}
\subsubsection{Baselines:}
We only choose matchers listed in \cref{sec:scene-baselines} as our baselines, for no deep learning-based regressors are available in this two-view relative object pose estimation task.
We utilize the stereo depth offered in the dataset to enable matcher-based approaches to estimate scaled relative poses via Orthogonal Procrustes~\cite{eggert1997estimating}, while SRPose only requires RGB images as inputs. The segmentations in the dataset are adopted for matcher-based approaches to identify the target object in both images, while SRPose utilizes the bounding box of the segmentation in the first image as the object prompt.

\begin{table}[t]
    \caption{\textbf{Relative object pose estimation on HO3D~\cite{hampali2020honnotate}.} SRPose achieves state-of-the-art performance in object-to-camera estimation without additional depth information, enabling scale pose estimation from only RGB inputs.}
    \centering
    \scriptsize
    \tabcolsep=4pt
    \vspace{-2.5mm}
    \begin{tabular}{lcccccccc}
    \toprule
        \multirow{2}{*}{Method} &  \multicolumn{3}{c}{Rot. (\(^\circ\))} &  \multicolumn{3}{c}{Trans. (cm)} & \multirow{2}{*}{ADD\(\uparrow\)} & \multirow{2}{*}{ADD-S\(\uparrow\)}\\
         &  Med.\(\downarrow\) &  Avg.\(\downarrow\) & \(\leq\ang{30}\)\(\uparrow\) & Med.\(\downarrow\) & Avg.\(\downarrow\)& \(\leq10\)cm\(\uparrow\) &  & \\
    \midrule
        SGMNet~\cite{chen2021learning} &  24.1 &  29.0 &  62.7 &  13.9 &  17.5 &  36.8 &  16.8 & 31.3 \\
        LightGlue~\cite{lindenberger2023lightglue} &  24.4 &  28.7 &  61.7 &  13.5 &  16.3 &  38.2 &  17.2 & 31.2 \\
        LoFTR~\cite{sun2021loftr} &  29.2 &  57.9 &  50.9 &  13.9 &  16.4 &  35.6 &  16.7 & 30.9 \\
        DKM~\cite{edstedt2023dkm} &  23.3 &  25.6 &  64.8 &  13.6&  15.4 &  39.8 &  17.8 & 33.3 \\
    \midrule
        SRPose &  \textbf{8.9} &  \textbf{11.4} & \textbf{95.4} & \textbf{5.9} & \textbf{8.0} &  \textbf{73.9} & \textbf{36.4} & \textbf{56.8} \\
    \bottomrule
    \end{tabular}
    \vspace{-4.5mm}
    \label{tab:ho3d}
\end{table}

\vspace{-4mm}
\subsubsection{Results:}
As shown in Table~\ref{tab:ho3d}, SRPose outperforms state-of-the-art matcher-based approaches significantly.
SRPose exhibits superior accuracy in relative 6D object pose estimation between two images using the object prompts in the reference view. 
The low quality of stereo depth leads to limited performance in scaled pose estimation of the matcher-based approaches, while SRPose can learn scaled information to achieve lower estimation errors.

\vspace{-3mm}
\subsection{Map-Free Visual Relocalization}
\vspace{-2mm}
We evaluate the application of SRPose on Niantic~\cite{arnold2022map}, a dataset for the map-free visual relocalization task.
For each scene in the validation and test sets, a reference image and a sequence of query images are provided to relocalize the query locations relative to the reference.
Following \cite{arnold2022map}, we adopt the Virtual Correspondence Reprojection Error (VCRE) and the median error of translation and rotation as the metrics.
Specifically, we compute the precision of VCRE at 90 pixels, and the precision of relocalization at 0.25m and \ang{5}.

\vspace{-4mm}
\subsubsection{Baselines:}
We still select the three categories of approaches as our baselines, including sparse matchers, SIFT~\cite{lowe2004distinctive}, SuperGlue~\cite{sarlin20superglue}; dense matchers, LoFTR~\cite{sun2021loftr}, RoMa~\cite{edstedt2023roma}; and the regressor Map-free~\cite{arnold2022map}.
For matcher-based approaches, we estimate the pose by recovering from the essential matrix, which typically yields slightly higher precision than using Orthogonal Procrustes in this task according to \cite{arnold2022map}.
Then we use the depth estimated by DPT~\cite{Ranftl2020}, a depth prediction model, to provide the matchers with the scale information.

\begin{table}
    \vspace{-2mm}
    \caption{\textbf{Map-free visual relocalization results on Niantic~\cite{arnold2022map}.} SRPose achieves competitive performance compared to state-of-the-art approaches without relying on an additional depth prediction model to assist with scaled estimation.}
    \vspace{-3mm}
    \centering
    \scriptsize
    \tabcolsep=4pt
    \begin{tabular}{llccc}
    \toprule
        \multirow{2}{*}{Category} & \multirow{2}{*}{Method} &  VCRE&  Med. (m / \(^{\circ}\))& \multirow{2}{*}{Prec.\(\uparrow\)}\\
         &  &  Prec.\(\uparrow\) / Med.\(\downarrow\)&  Trans.\(\downarrow\) / Rot.\(\downarrow\)&  \\
    \midrule
        \multirow{2}{*}{\makecell[c]{Sparse\\+DPT~\cite{Ranftl2020}}} &  SIFT~\cite{lowe2004distinctive} & 25.0 / 222.8 & 2.93 / 61.4 & 10.3 \\
         & SuperGlue~\cite{sarlin20superglue} & 36.1 / 160.3 & 1.88 / 25.4 & 16.8 \\
    \midrule
        \multirow{2}{*}{\makecell[c]{Dense\\+DPT~\cite{Ranftl2020}}} &  LoFTR~\cite{sun2021loftr} & 34.7 / 167.6 & 1.98 / 30.5 & 15.4 \\
         &  RoMa~\cite{edstedt2023roma} & 45.6 / 128.8 & \textbf{1.23} / \textbf{11.1} & \textbf{22.8} \\
    \midrule
        \multirow{2}{*}{Regressor} &  Map-free~\cite{arnold2022map} & 40.2 / 147.1 & 1.68 / 22.9 & 6.0\\
         &  SRPose& \textbf{46.4} / \textbf{127.7} & 1.37 / 17.2 & 16.9\\
    \bottomrule
    \end{tabular}
    \label{tab:niantic}
    \vspace{-2mm}
\end{table}

\vspace{-4mm}
\subsubsection{Results:}
Table~\ref{tab:niantic} shows SRPose outperforms other sparse matchers and the regression method in map-free visual relocalization. 
Although the state-of-the-art dense matcher RoMa achieves higher precision in relocalization using scales estimated by an additional depth prediction model, SRPose still outperforms in VCRE.
Our framework exhibits competitive performance compared to state-of-the-art approaches in the map-free visual relocalization task.

\subsection{Insights}
\vspace{-2mm}
\begin{table}[t]
    \caption{\textbf{Ablation study on Matterport~\cite{chang2017matterport3d}.} The prior knowledge-guided attention layers assist in establishing implicit cross-view correspondences.}
    \vspace{-3mm}
    \scriptsize
    \centering
    \tabcolsep=4pt
    \begin{tabular}{lcccc}
    \toprule
        \multirow{2}{*}{Method} &  \multicolumn{2}{c}{Trans. (m)}&  \multicolumn{2}{c}{Rot. (\(^{\circ}\))}\\
         &  Avg.\(\downarrow\)&  \(\leq1\)m\(\uparrow\) & Avg.\(\downarrow\) & \(\leq\ang{30}\)\(\uparrow\) \\
    \midrule
        8point~\cite{rockwell20228} & 1.01 & 67.4 & 19.13 & 85.4\\
    \midrule
        SRPose (full) & \textbf{0.84} & \textbf{74.2} & \textbf{14.32} & \textbf{88.9} \\
        1) no prior knowledge guidance & 0.97 & 69.6 & 16.91 & 86.8 \\
        2) replace all cross-attn. with self-attn. & 1.90 & 26.1 & 48.0 & 42.0 \\
        3) no position encoding & 1.17 & 61.0 & 21.39 & 81.3\\
    \bottomrule
    \end{tabular}
    \vspace{-5.5mm}
    \label{tab:ablation-matterport}
\end{table}

\begin{wraptable}{r}{0.5\linewidth}
    \vspace{-8mm}
    \caption{\textbf{Ablation study on MegaDepth \cite{MDLi18}.} Intrinsic-calibration position encoder enables adaption on varying image sizes and camera intrinsics.}
    \scriptsize
    \centering
    \tabcolsep=3pt
    \begin{tabular}{lccc}
    \toprule
        \multirow{2}{*}{Method} &  \multicolumn{3}{c}{Pose est. AUC}\\
         &  @\ang{5}&  @\ang{10}& @\ang{20}\\
    \midrule
        Map-free~\cite{arnold2022map} & 2.6 & 9.3 & 22.9 \\
    \midrule
        SRPose (full) & \textbf{16.6} & \textbf{36.0} & \textbf{58.0} \\
        1) no intrinsic calibration & 1.5 & 8.5 &  24.3\\
        2) no position encoding & 1.0 & 6.2 &  18.9\\
    \bottomrule
    \end{tabular}
    \vspace{-7mm}
    \label{tab:ablation-megadepth}
\end{wraptable}

\subsubsection{Ablation study:}
We evaluate the effectiveness of the different components in SRPose by Matterport~\cite{chang2017matterport3d} and MegaDepth~\cite{MDLi18} datasets.
Specifically, we train four different variants initialized with random weights on Matterport by removing the three different designs we proposed to validate their effectiveness.
As shown in Table~\ref{tab:ablation-matterport}:
\textbf{1)} Removing the prior knowledge guidance of keypoint similarities in the cross-attention modules results in a moderate decrease in accuracy, as it assists SRPose in seeking the cross-view correspondences.
\textbf{2)} Replacing all cross-attention modules with self-attention modules leads to a significant accuracy drop, as the capability to establish implicit cross-view correspondences through cross-attention is eliminated.
\textbf{3)} Removing the entire position encoder leads to a moderately lower accuracy due to the absence of keypoint position information.
The experiment highlights the crucial roles of the three designs of SRPose in accurate relative pose estimation.

Moreover, we train three different variants initialized with random weights on MegaDepth, a dataset that consists of images with different sizes and camera intrinsics, to further validate the effectiveness of the intrinsic-calibration position encoder, with results shown in Table~\ref{tab:ablation-megadepth}.
\textbf{1)} Without enforcing intrinsic calibration on keypoint coordinates, SRPose only achieves comparable performance to Map-free, both inadaptable to varying image sizes and camera intrinsics in MegaDepth.
\textbf{2)} Removing the entire IC Position Encoder results in a significant drop in accuracy as expected owing to the elimination of critical position information. 
By comparing the full framework with the variants lacking IC and the entire position encoder, we demonstrate the essential roles of the two designs in adapting different image sizes and camera intrinsics.

\vspace{-5mm}
\subsubsection{Visualizing attention:}
\begin{figure}[htbp]
    \vspace{-4mm}
	\centering
    \begin{tabular}{ccccc}
        \scriptsize
        \rotatebox[origin=l]{90}{\hspace{6.5mm}Full} & 
    	\begin{subfigure}{0.22\linewidth}
            \centering
            \includegraphics[width=\linewidth]{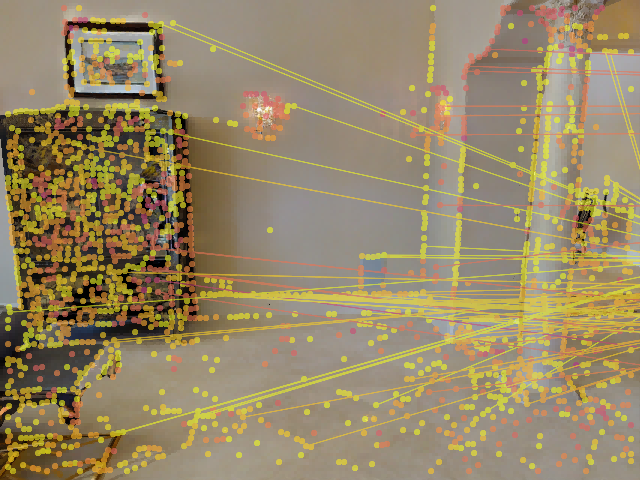}
    	\end{subfigure} &
        \begin{subfigure}{0.22\linewidth}
            \centering
            \includegraphics[width=\linewidth]{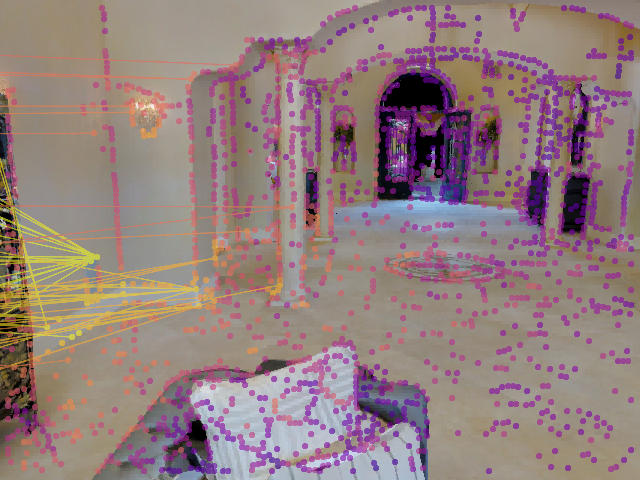}
    	\end{subfigure} &
        \begin{subfigure}{0.22\linewidth}
            \centering
            \includegraphics[width=\linewidth]{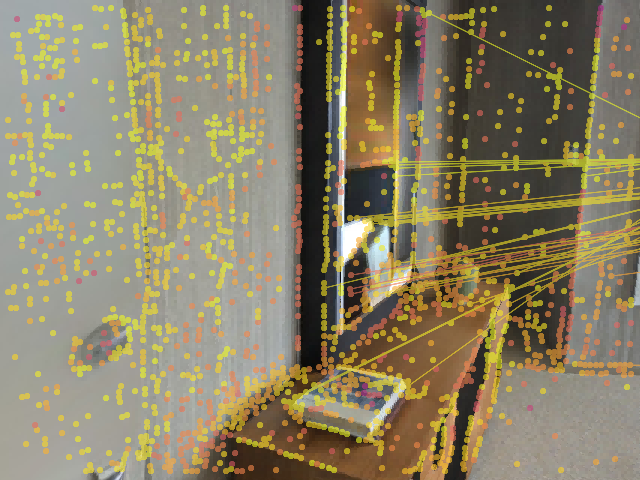}
    	\end{subfigure} &
        \begin{subfigure}{0.22\linewidth}
            \centering
            \includegraphics[width=\linewidth]{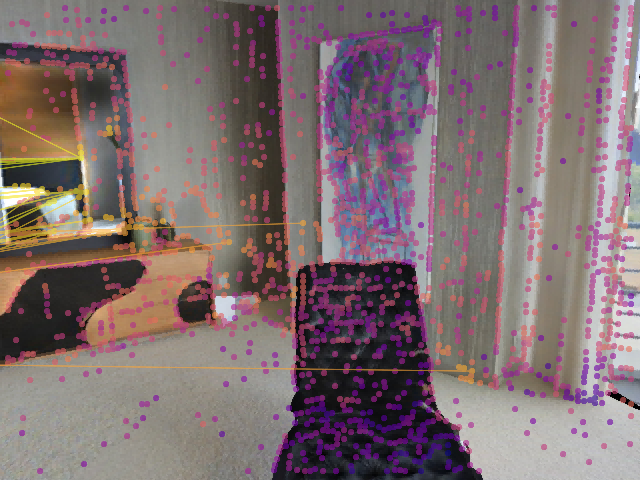}
        \end{subfigure} \\
        \scriptsize
        \rotatebox[origin=l]{90}{\hspace{1mm} No guidance} & 
        \begin{subfigure}{0.22\linewidth}
            \centering
            \includegraphics[width=\linewidth]{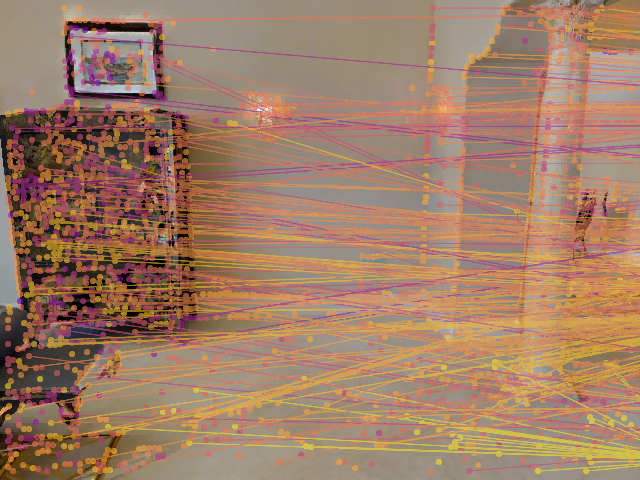}
    	\end{subfigure} &
    	\begin{subfigure}{0.22\linewidth}
            \centering
            \includegraphics[width=\linewidth]{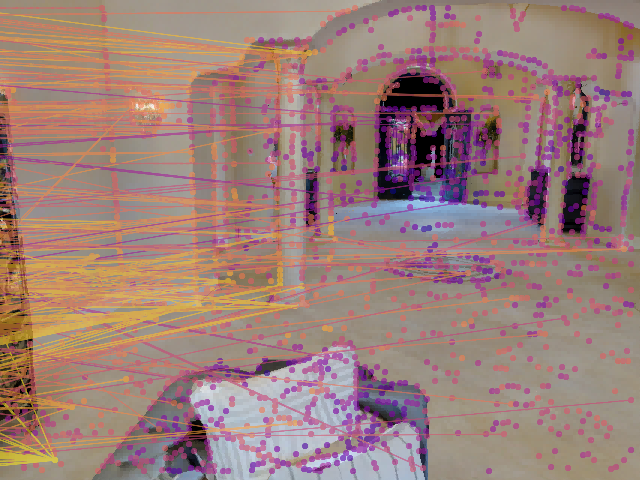}
    	\end{subfigure} &
        \begin{subfigure}{0.22\linewidth}
            \centering
            \includegraphics[width=\linewidth]{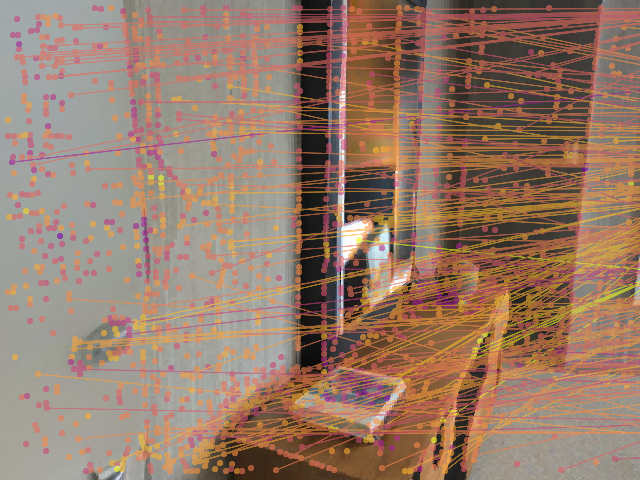}
    	\end{subfigure} &
    	\begin{subfigure}{0.22\linewidth}
            \centering
            \includegraphics[width=\linewidth]{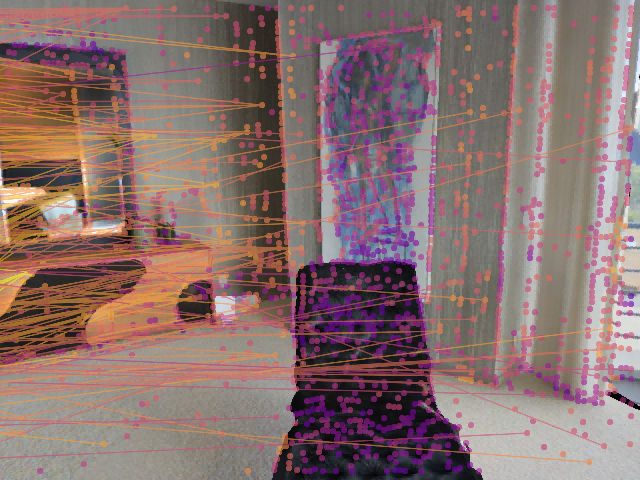}
    	\end{subfigure} \\
        \scriptsize
        \rotatebox[origin=l]{90}{\hspace{6.5mm} Full} & 
        \begin{subfigure}{0.22\linewidth}
            \centering
            \includegraphics[width=\linewidth]{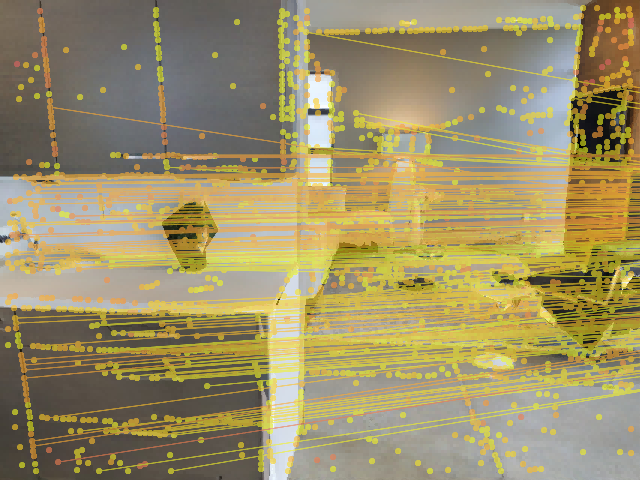}
    	\end{subfigure} &
    	\begin{subfigure}{0.22\linewidth}
            \centering
            \includegraphics[width=\linewidth]{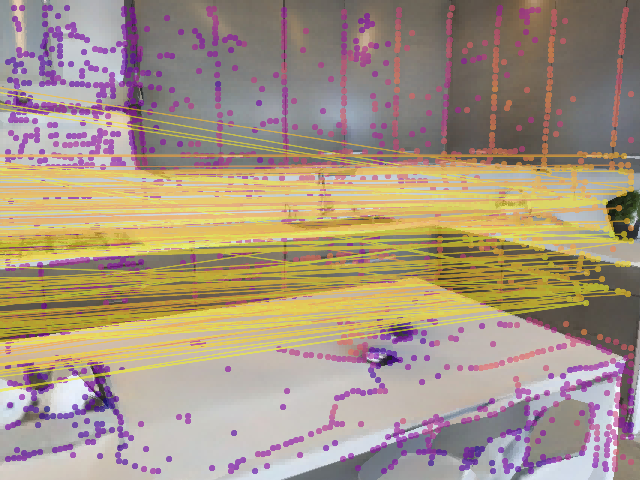}
    	\end{subfigure} &
        \begin{subfigure}{0.22\linewidth}
            \centering
            \includegraphics[width=\linewidth]{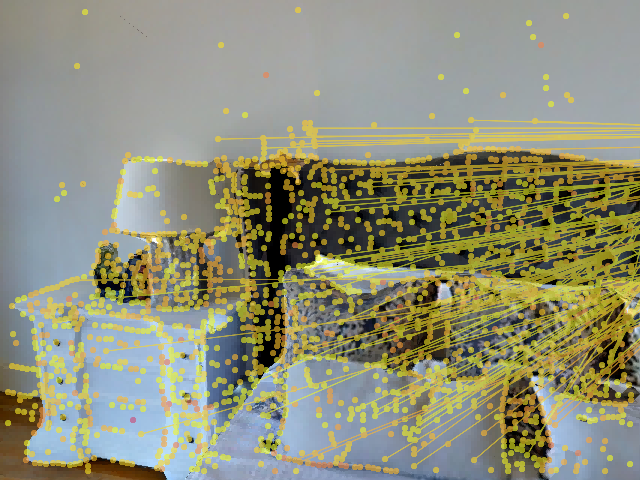}
    	\end{subfigure} &
    	\begin{subfigure}{0.22\linewidth}
            \centering
            \includegraphics[width=\linewidth]{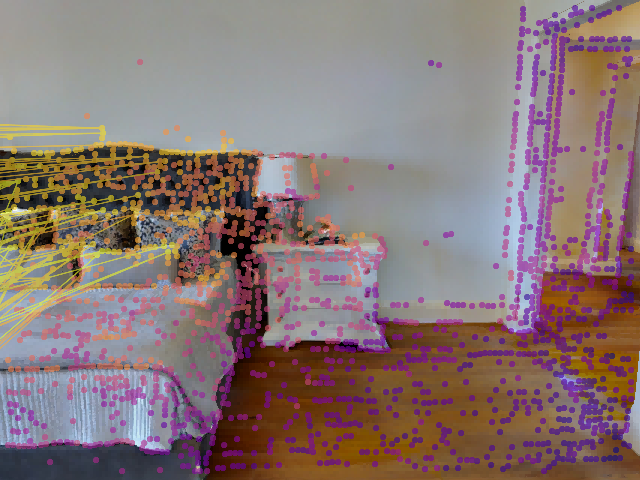}
    	\end{subfigure} \\
        \scriptsize
        \rotatebox[origin=l]{90}{\hspace{1mm} No guidance} & 
        \begin{subfigure}{0.22\linewidth}
            \centering
            \includegraphics[width=\linewidth]{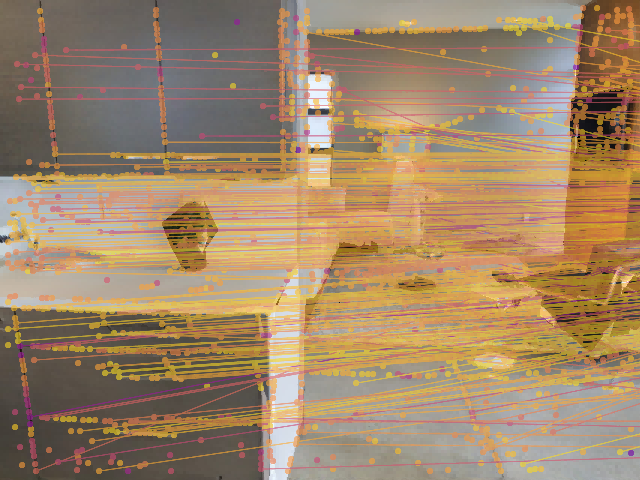}
    	\end{subfigure} &
    	\begin{subfigure}{0.22\linewidth}
            \centering
            \includegraphics[width=\linewidth]{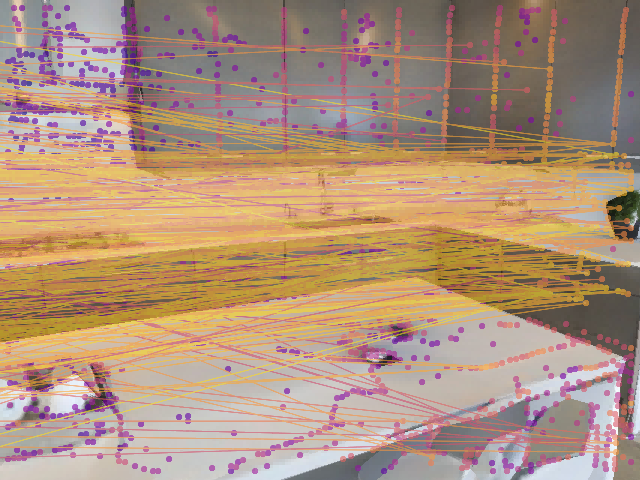}
    	\end{subfigure} &
        \begin{subfigure}{0.22\linewidth}
            \centering
            \includegraphics[width=\linewidth]{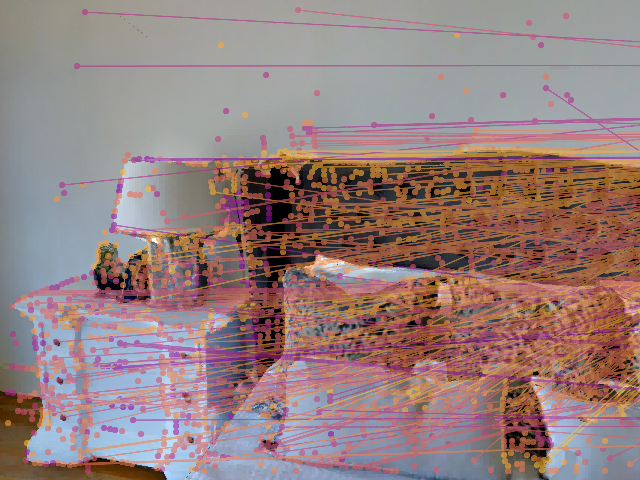}
    	\end{subfigure} &
    	\begin{subfigure}{0.22\linewidth}
            \centering
            \includegraphics[width=\linewidth]{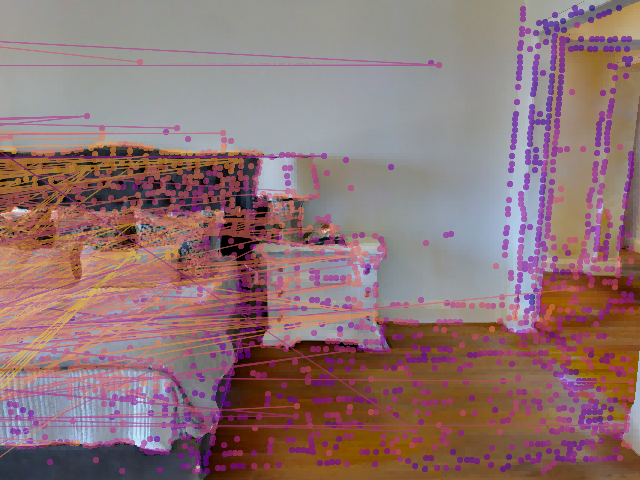}
    	\end{subfigure} \\
    \end{tabular}
    \vspace{-3mm}
    \caption{\textbf{Visualization of implicit cross-view correspondences.} Dots and lines drawn with brighter colors represent higher cross-view attention scores. The prior knowledge guidance enhances attention to the overlapping areas, removing the irrelevant cross-view connections, and assisting in establishing implicit correspondences.}
    \label{fig:ablation}
    \vspace{-8mm}
\end{figure}

In \cref{fig:ablation}, we visualize the cross-attention scores \(A_{12}'\), and compare SRPose without prior knowledge guidance to its full designs.
We also draw the connections of highly-attended cross-view keypoint pairs, which represent the implicit correspondences established by SRPose.
\renewcommand\floatpagefraction{.9}
\renewcommand\topfraction{.9}
\renewcommand\bottomfraction{.9}
\renewcommand\textfraction{.1}
\setcounter{totalnumber}{50}
\setcounter{topnumber}{50}
\setcounter{bottomnumber}{50}

%% file: conclusion.tex
\section{Conclusion}
\vspace{-3mm}
This paper presents SRPose, a novel deep learning-based regressor utilizing sparse keypoints for two-view relative pose estimation in camera-to-world and object-to-camera scenarios. As our key innovations, SRPose extracts keypoints with sparse keypoint detectors and employs an intrinsic-calibration position encoder to enable adaptability to different image sizes and camera intrinsics. Further, our proposed promptable prior knowledge-guided attention layers establish implicit correspondences to estimate the relative pose under the epipolar constraint in both scenarios. By directly regressing the rotation and translation, SRPose achieves a significant decrease in computational time, with a minimum time reduction of 200ms. Extensive experiments demonstrate that SRPose achieves state-of-the-art performance in terms of accuracy and speed in the two scenarios, as well as in the map-free visual relocalization task.

%% file: appendix.tex
\vspace{-3mm}

\section{Overview}
\vspace{-2mm}
This is the supplementary material \textit{SRPose: Two-view Relative Pose Estimation with Sparse Keypoints}. In \cref{sec:problem}, we provide the detailed definitions of the problems addressed in both scenarios. In \cref{sec:ic}, we further elaborate on why and how we enforce intrinsic calibration in SRPose. \cref{sec:imp} contains more implementation details in our experiments. \cref{sec:qua} and \cref{sec:vis} provide more visualizations of the results and mechanisms of our framework. In \cref{sec:limit}, we discuss the limitations of SRPose and propose several directions for future research. The code for SRPose can be accessed at \href{https://github.com/frickyinn/SRPose/tree/main}{https://github.com/frickyinn/SRPose/tree/main}.

\vspace{-3mm}
\input{problem.tex}

\vspace{-3mm}
\section{Intrinsic Calibration}
\label{sec:ic}
\vspace{-2mm}
\begin{figure}[htbp]
    \centering
    \includegraphics[width=0.88\linewidth]{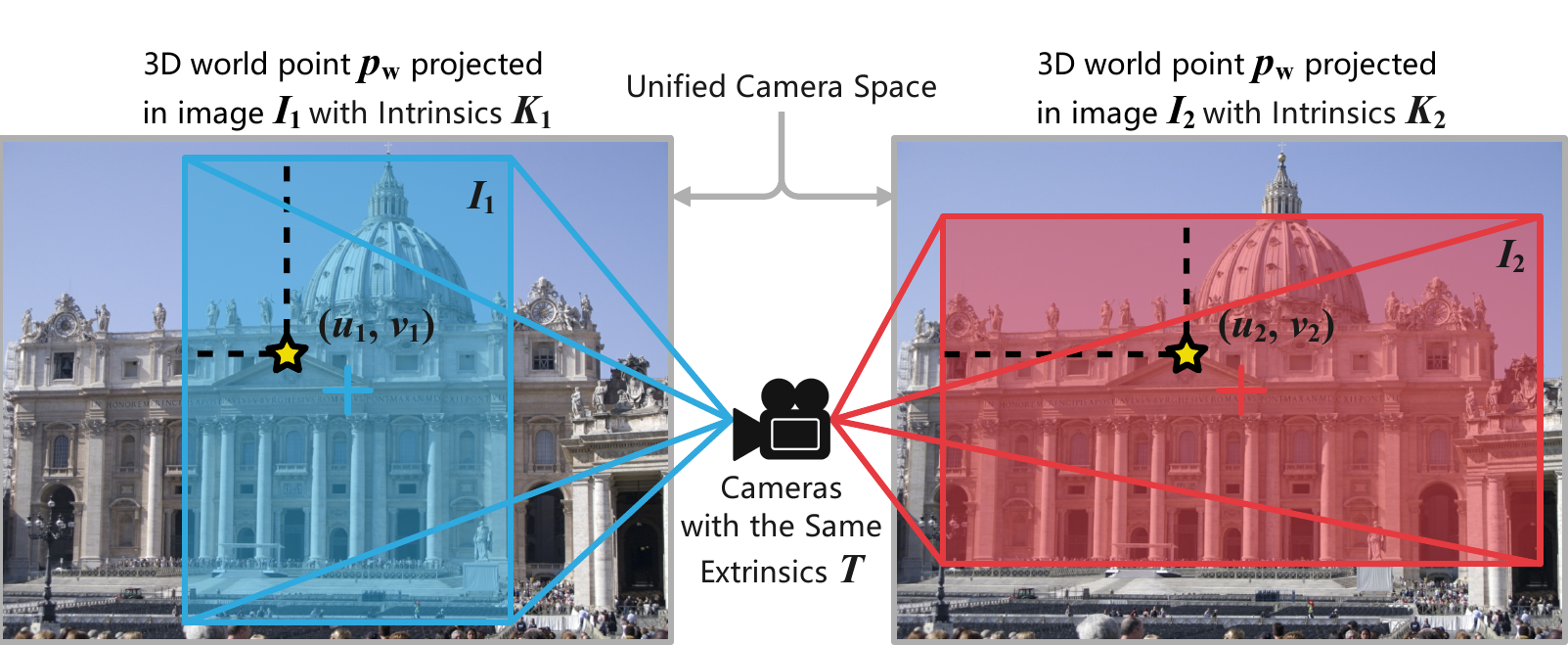}
    \vspace{-1mm}
    \caption{The 3D world point \(p_\mathrm{w} \in P_\mathrm{w}\) captured by two cameras with the same extrinsics \(T\) but different intrinsics \(K_1, K_2\), leading to distinct coordinates in the two images \(I_1, I_2\). Intrinsic calibration is performed to undistort the coordinates of the keypoints, normalizing them into a unified camera space.}\label{fig:ic}
    \vspace{-4mm}
\end{figure}
\cref{fig:ic} illustrates the projection of the same 3D world point \(p_\mathrm{w}\) in the scene onto two camera spaces of \(I_1, I_2\) by two cameras with different intrinsics. Although the two cameras have the same extrinsics \(T\), \ie, the same camera location and orientation, the resulting coordinates are different:
\begin{gather}
    P_1 = K_1 T P_{\mathrm{w}}, \label{diff-K1} \\
    P_2 = K_2 T P_{\mathrm{w}}. \label{diff-K2}
\end{gather}
The difference arises from the variations in image sizes and camera intrinsics.
Previous regressors resize all inputs to a fixed size, which alters the pixel or point coordinates in the images, leading to inaccurate position information.
While SRPose also first resizes all images to the same size for batched keypoint extraction, it then rescales the detected keypoints' coordinates to their original positions, so that the impacts of varying image sizes are eliminated. 
Image encoders employed by existing regressors process images solely with positions in the camera coordinate system, \ie, \(P_1, P_2\) without considering different intrinsics.
However, different values in the camera intrinsic matrices \(K_1, K_2\) resulted from two different pieces of equipment, \eg, a cellphone and a digital single-lens camera, project \(P_w\) to different positions in the camera space.
This difference impedes the establishment of implicit correspondences due to the resulting inaccurate position relations.
To this end, SRPose enforces intrinsic calibration according to ~\cite{nister2004efficient}.
Undistorting keypoints' coordinates with intrinsics offers the precise position information for correspondence establishment:
\begin{gather}
    P_1^c = K_1^{-1} P_1, \label{int1} \\
    P_2^c = K_2^{-1} P_2, \label{int2}
\end{gather}
Intrinsic calibration normalizes the points from different camera spaces to a unified camera coordinate system, which ensures a camera-invariant, generalizable, and accurate performance in relative pose estimation.

\vspace{-3mm}
\section{Implementation Details}
\label{sec:imp}

\vspace{-2mm}
\subsection{Fine-tuning}
\vspace{-2mm}
To achieve better performance, SRPose is first trained on ScanNet~\cite{dai2017scannet} for 500 epochs as the pre-training stage. Then we fine-tune the framework on other datasets, including Matterport~\cite{chang2017matterport3d}, Linemod~\cite{hinterstoisser2013model}, HO3D~\cite{hampali2020honnotate}, Niantic~\cite{arnold2022map}, and MegaDepth~\cite{MDLi18}.
Since only Matterport and Niantic have the validation sets, we select the model checkpoints and the hyper-parameters with the best performance on their validation sets.
For other datasets, we select \(2\times 10^{-5}\) as the maximum learning rate (LR) of 1cycle policy~\cite{smith2019super} for fine-tuning.
As ScanNet and MegaDepth contain more than 1 million image pairs in the training sets, we set their training epochs to 500 to learn the information fully, while others to 200.
Due to the memory limitation, we extract 1,024 keypoints on all datasets in the training stage, except on Linemod. As the target objects in Linemod only occupy a small fraction of pixels, we extract 1,200 keypoints to facilitate the training.
Notably, we train our framework initialized with random weights in the ablation study and the comparison experiments of different sparse keypoint detectors.
Table \ref{tab:hyper} lists the hyper-parameter selections on different datasets.

\begin{table}
    \vspace{-4mm}
    \caption{\textbf{Hyper-parameter selections on different datasets.}}
    \vspace{-3mm}
    \centering
    \footnotesize
    \tabcolsep=6pt
    \begin{tabular}{lccc}
    \toprule
        Dataset & Max. LR & Epochs\\
    \midrule
        Matterport~\cite{chang2017matterport3d} & \(5\times 10^{-5}\) & 200 &\\
        ScanNet~\cite{dai2017scannet} & \(2\times 10^{-4}\) & 500\\
        MegaDepth~\cite{MDLi18} & \(2\times 10^{-5}\) & 500\\
        HO3D~\cite{hampali2020honnotate} & \(2\times 10^{-5}\) & 200\\
        Linemod~\cite{hinterstoisser2013model} & \(2\times 10^{-5}\) & 200\\
        Niantic~\cite{arnold2022map} & \(2\times 10^{-5}\) & 200\\
    \bottomrule
    \end{tabular}
    \vspace{-8mm}
    \label{tab:hyper}
\end{table}

\vspace{-2mm}
\subsection{Experimental Settings}
\vspace{-2mm}
In this section, we introduce more details about the datasets and the experimental settings we used in the main text and the supplementary materials.
Then, we explain some of the complicated metrics we used in the experiments.

\vspace{-3mm}
\subsubsection{Camera-to-world pose estimation:}
In the camera-to-world scenario, we evaluation our SRPose on Matterport~\cite{chang2017matterport3d}, ScanNet~\cite{dai2017scannet} and MegaDepth~\cite{MDLi18}.
Matterport is a re-rendering dataset from real scenes using the Habitat~\cite{savva2019habitat} system. We adopt the preprocessed dataset following \cite{jin2021planar} for training and evaluation. The dataset consists of 31,932/4,707/7,996 images in the train/validation/test sets, respectively. 
For the test set, the average rotation angle is \ang{53.5}, and the average translation length is 2.31m. ScanNet is a dataset that consists of real scenes, containing 1613 monocular sequences. Following the guidelines in \cite{sarlin20superglue}, we sample 230M image pairs for training and use the ScanNet-1500 test set for evaluation. The ScanNet-1500 test set has an average rotation angle of \ang{29.6} and an average translation length of 0.88m. MegaDepth is a dataset that consists of 1M internet images of 196 different outdoor scenes. We use the MegaDepth-1500 test set following \cite{tyszkiewicz2020disk, sun2021loftr}, which consists of two scenes excluded from the train set: Sacre Coeur, and St. Peter's Square. The test set has an average rotation angle of \ang{12.7} and an average translation length of 2.46m.

\vspace{-3mm}
\subsubsection{Object-to-camera pose estimation:}
In the object-to-camera scenario, we evaluate SpaRelPose on HO3D~\cite{hampali2020honnotate} and Linemod~\cite{hinterstoisser2013model}.
HO3D contains videos of hand-holding objects of 21 categories from the YCB dataset~\cite{calli2015ycb}. Each video portrays a single object being held by a human hand transforming its pose, with fixed cameras and backgrounds.
Linemod consists of 15 categories of objects, with the entire training set being synthetic data, and the test set being real-scene data. Each image in Linemod contains multiple objects arbitrarily scattered in various scenes.
Linemod offers depth maps of each view in the dataset, which are captured by LiDAR depth camera with high precision.
The bounding boxes and object segmentation of the target object are also provided.
For both datasets, We randomly select image pairs from the training sets with relative rotation angles less than \ang{45} for training, and the pairs with rotation angles less than \ang{30} for evaluation.
For HO3D, we select 3000 frame pairs from the five videos excluded from the training set as the new test set, with an average rotation angle and average translation length of \ang{17.2} and 0.12m.
The test set contains three categories of texture-less objects that appear in the training set, and two categories of regular-shape, unseen, and textureful objects.
For Linemod, we randomly pick 1500 image pairs from the real-scene data as the new test set, with an average rotation angle and translation length of \ang{20.7} and 0.88m.

For matcher-based baselines, we first crop the target object out of two views.
The resulting two cropped images are then resized such that their larger dimension is 640 pixels.
We mask out the matching features on the background with the object segmentation, and finally recover the relative poses from the essential matrices just as in the camera-to-world scenario.

\vspace{-3mm}
\subsubsection{Map-free visual relocalization:}
We evaluate SRPose in the task of map-free relocalization on the Niantic~\cite{arnold2022map} dataset. Niantic is a dataset built specifically for the task of map-free visual localization, consisting of 460/65/130 scenes in the training/validation/test set, respectively.
For validation and test sets, a single reference image and a sequence of query images are provided. The goal is to relocalize the querying locations according to the reference.
The ground truth of the test set is not publicly available, the evaluation of this dataset is performed by submitting the relative poses estimated by our SRPose on the project page. The evaluation scores of different metrics will be measured by the server.

\vspace{-3mm}
\subsubsection{Metrics:}
We adopt the average distance metrics, \ie, ADD and ADD-S, to evaluate the performance in the object-to-camera scenario, following \cite{hinterstoisser2013model}. Given the ground-truth rotation and translation \(R_{gt}, t_{gt}\), and the estimated \(R, t\), ADD computes the mean of the pairwise distances between the 3D object model points transformed according to the ground truth and the estimation:
\begin{equation}
    ADD = \frac{1}{m} \sum_{x\in \mathcal{M}} \| (R_{gt}x + t_{gt}) - (Rx + t)\|, \label{add}
\end{equation}
in which \(\mathcal{M}\) is the set of 3D model points, \ie, the object point cloud, and the \(m\) is the number of the points.
To evaluate symmetric objects, such as Mug and Banana, ADD-S is also computed using the closest point distance:
\begin{equation}
    ADD = \frac{1}{m} \sum_{x\in \mathcal{M}} \| (R_{gt}x + t_{gt}) - (Rx + t)\|. \label{add-s}
\end{equation} 
Following \cite{xiang2018posecnn}, we compute and report the area under the curve (AUC) of ADD and ADD-S with the threshold set to 10cm.

We evaluate SRPose in the map-free visual relocalization with the metric of Virtual Correspondence Reprojection Error (VCRE) following \cite{arnold2022map}. The ground truth and estimated relative pose transformations are used to project virtual 3D points, located in the query camera's local coordinate system. VCRE is the average Euclidean distance of the reprojection errors:
\begin{equation}
    VCRE = \frac{1}{|\mathcal{V}|} \sum_{v\in \mathcal{V}} \| \pi(v) - \pi(TT_{gt}^{-1} v) \|_2, \quad T = [R|t],
\end{equation}
where \(\pi\) is the image projection function, and \(\mathcal{V}\) is a set of 3D points in camera space representing virtual objects. This metric provides an intuitive measure of AR content misalignment in the map-free relocalization task.

\vspace{-3mm}
\section{Additional Results}
\label{sec:add}
\vspace{-2mm}
In this section, we present more experimental results to further assess the advantages and limitations of our framework. The section includes two additional evaluation results on Linemod~\cite{hinterstoisser2013model} and MegaDepth~\cite{MDLi18} datasets that are excluded from the main text, the further reports on HO3D~\cite{hampali2020honnotate}, and the comparison of SRPose variants using different sparse keypoint detectors.

\vspace{-2mm}
\subsection{Object-to-Camera Pose Estimation}

\vspace{-3mm}
\subsubsection{Results on Linemod:}
As Table \ref{tab:linemod} shows, SRPose underperforms on Linemod in ADD and ADD-S compared to other matcher-based approaches.
Although our framework exhibits competitive results in terms of rotation estimation, the failure in translation hinders higher accuracy in the overall performance.
First, the highly precise LiDAR depths offer more information about the 6D pose transformation to baseline approaches.
Second, target objects in Linemod images only occupy a small fraction of pixels among multiple objects in the scenes.
SRPose can only extract about 100 keypoints from the object prompt in each image, making the task difficult.
While matcher-based approaches require object segmentation in both images, and they resize the segmented object to larger sizes to obtain higher accuracy.
Nonetheless, SRPose achieves competitive performance in rotation with only one object prompt in one of the views.

\begin{table}[t]
    \caption{\textbf{Relative object pose estimation on Linemod~\cite{hinterstoisser2013model}.} SRPose fails in the translation estimation while achieving competitive performance in rotation compared to matcher-based approaches plus LiDAR depth maps and object segmentation.}
    \centering
    \scriptsize
    \tabcolsep=3pt
    \vspace{-2mm}
    \begin{tabular}{lcccccccc}
    \toprule
        \multirow{2}{*}{Method} &  \multicolumn{3}{c}{Trans. (cm)}&  \multicolumn{3}{c}{Rot. (\(^\circ\))}& \multirow{2}{*}{ADD\(\uparrow\)} & \multirow{2}{*}{ADD-S\(\uparrow\)}\\
         &  Med.\(\downarrow\)&  Avg.\(\downarrow\)& \(\leq10\)cm\(\uparrow\) & Med.\(\downarrow\) & Avg.\(\downarrow\) & \(\leq\ang{30}\)\(\uparrow\) &  & \\
    \midrule
        SGMNet~\cite{chen2021learning} & n/a&  n/a&  n/a& 24.16&  40.1 &  60.1&   n/a &  n/a \\
        LightGlue~\cite{lindenberger2023lightglue} &  n/a &  n/a&   n/a&  22.24&  40.46&  62.7&   n/a &  n/a \\
        LoFTR~\cite{sun2021loftr} &  n/a &  n/a&  n/a&  21.17&  41.38&  64.1&   n/a &  n/a \\
        DKM~\cite{edstedt2023dkm} &   n/a &   n/a&   n/a&   \textbf{12.14}& \textbf{ 23.24}& \textbf{ 78.3}&   n/a &  n/a \\
    \midrule
        SGMNet~\cite{chen2021learning} + Depth &  11.8&  24.4&  45.2&  9.7&  17.2&  86.0&  20.3 & 35.1 \\
        LightGlue~\cite{lindenberger2023lightglue} + Depth &  13.0 &  24.3&  43.3&  9.7&  20.4&  83.5&  23.6 & 38.1 \\
        LoFTR~\cite{sun2021loftr} + Depth &  10.1 &  19.2&  49.9&  7.8&  22.6&  85.6&  25.8 & 41.7 \\
        DKM~\cite{edstedt2023dkm} + Depth &  \textbf{6.0} &  \textbf{10.2}&  \textbf{67.7}&  \textbf{4.6}&  \textbf{7.7}&  96.5&  \textbf{37.8} & \textbf{56.5} \\
    \midrule
        SRPose &  13.8&  16.1& 29.7& 9.1& 10.5&  \textbf{98.7}& 10.2 & 27.8 \\
    \bottomrule
    \end{tabular}
    \vspace{-2mm}
    \label{tab:linemod}
\end{table}

\vspace{-3mm}
\subsubsection{Additional results on HO3D:}
Table \ref{tab:cate} shows the results of different categories in HO3D. Typically, all methods achieve higher accuracy in objects with rich textures, for it helps establish correspondences.
In Table \ref{tab:linemod-depth}, we also compare the baselines using Orthogonal Procrustes~\cite{eggert1997estimating} to themselves without depth maps using essential matrices.
We only evaluate the performance in rotation because traditional approaches are incapable of scaled translation estimation without depth.
Although stereo depth has lower precision compared to LiDAR depth, it still contributes to the relative 6D object pose estimation for the matchers.

\begin{table}
    \vspace{-4mm}
    \centering
    \scriptsize
    \tabcolsep=4pt
    \caption{\textbf{Categorized relative pose estimation results on HO3D~\cite{hampali2020honnotate}.} 003: Cracker box; 006: Mustard bottle; 011: Banana; 025: Mug; 037: Scissors.}
    \vspace{-2mm}
    \begin{tabular}{lccccc}
    \toprule
        \multirow{2}{*}{Method} &  \multicolumn{5}{c}{ADD / ADD-S}\\
         &  003&  006&  011&  025&  037\\
    \midrule
        SGMNet~\cite{chen2021learning} & 16.5 / 36.7 & 12.4 / 24.1 & 17.6 / 30.1 & 17.1 / 32.3 & 22.7 / 37.5 \\
        LightGlue~\cite{lindenberger2023lightglue} & 16.6 / 36.9 & 12.6 / 23.9 & 17.1 / 29.7 & 17.9 / 31.8 & 22.4 / 34.6 \\
        LoFTR~\cite{sun2021loftr} & 16.8 / 36.9 & 12.2 / 23.9 & 18.2 / 30.8 & 15.2 / 29.6 & 20.6 / 35.6 \\
        DKM~\cite{edstedt2023dkm} & 16.6 / 36.7 & 12.8 / 24.9 & 18.6 / 32.0 & 18.0 / 34.4 & 24.0 / 38.7 \\
    \midrule
        SRPose & \textbf{44.9} / \textbf{68.2} & \textbf{55.6} / \textbf{75.1} & \textbf{21.7} / \textbf{35.8} & \textbf{36.2} / \textbf{60.1} & \textbf{22.5} / \textbf{44.9} \\
    \bottomrule
    \end{tabular}
    \vspace{-8mm}
    \label{tab:cate}
\end{table}

\begin{table}
    \vspace{-4mm}
    \caption{\textbf{Comparison with and without depth on HO3D~\cite{hampali2020honnotate}.} Depth maps assist matcher-based approaches in estimating relative object pose transformation.}
    \vspace{-2mm}
    \scriptsize
    \tabcolsep=4pt
    \centering
    \begin{tabular}{lcccc}
    \toprule
        \multirow{2}{*}{Method} &  \multicolumn{4}{c}{Rotation (\(^\circ\))}\\
         & Med.\(\downarrow\) & Avg.\(\downarrow\) & \(\leq\ang{30}\)\(\uparrow\) & \(\leq\ang{15}\)\(\uparrow\)\\
    \midrule
        SGMNet~\cite{chen2021learning} &  27.7 &  36.0 &  55.1 & 23.4 \\
        SGMNet~\cite{chen2021learning} + Depth &  24.1 &  29.0 &  62.7 & 29.1 \\
    \midrule
        LightGlue~\cite{lindenberger2023lightglue} & 28.0  &  38.2 &  54.1 & 23.9 \\
        LightGlue~\cite{lindenberger2023lightglue} + Depth &  24.4 &  28.7 &  61.7 & 29.1 \\
    \midrule
        LoFTR~\cite{sun2021loftr} & 33.5  &  66.8 &  45.7 & 20.5 \\
        LoFTR~\cite{sun2021loftr} + Depth &  24.4 &  28.7 &  61.7 & 29.1 \\
    \midrule
        DKM~\cite{edstedt2023dkm} & 27.9  &  34.9 &  54.1 & 24.0 \\
        DKM~\cite{edstedt2023dkm} + Depth &  23.3 &  25.6 &  64.8 & 29.9 \\
    \midrule
        SRPose &  \textbf{8.9} &  \textbf{11.4} & \textbf{95.4} & \textbf{73.3} \\
    \bottomrule
    \end{tabular}
    \vspace{-8mm}
    \label{tab:linemod-depth}
\end{table}

\vspace{-2mm}
\subsection{Object-to-World Pose Estimation}

\vspace{-3mm}
\subsubsection{Results on MegaDepth:}
\label{sec:megadepth}

We further evaluate outdoor relative pose estimation on MegaDepth~\cite{MDLi18}. 
Table \ref{tab:megadepth} shows SRPose underperforms on MegaDepth. We discuss the limitations of our framework regarding to this performance in \cref{sec:limit}.

\begin{table}[t]
    \caption{\textbf{Relative pose estimation on MegaDepth~\cite{MDLi18}.} SRPose is not adept at estimating small relative pose transformation.}
    \vspace{-2mm}
    \centering
    \scriptsize
    \tabcolsep=8pt
    \begin{tabular}{llccc}
    \toprule
        \multirow{2}{*}{Category} & \multirow{2}{*}{Method} &  \multicolumn{3}{c}{Pose estimation AUC}\\
         &  &  @\ang{5}&  @\ang{10}&  @\ang{20}\\
    \midrule
        \multirow{3}{*}{Sparse} &  SuperGlue~\cite{sarlin20superglue}&  \textbf{49.7}&  \textbf{67.1}&  \textbf{80.6}\\
         &  SGMNet~\cite{chen2021learning}&  43.2&  61.6&  75.6\\
         &  LightGlue~\cite{lindenberger2023lightglue}&  49.4&  67.0&  80.1\\
    \midrule
        \multirow{3}{*}{Dense} &  LoFTR~\cite{sun2021loftr}& 52.8& 69.2& 81.2\\
         &  ASpanFormer~\cite{chen2022aspanformer}&  55.3&  71.5&  83.1\\
         &  DKM~\cite{edstedt2023dkm}&  \textbf{60.4}&  \textbf{74.9}&  \textbf{85.1}\\
    \midrule
        \multirow{2}{*}{Regressor} &  Map-free~\cite{arnold2022map}&  2.6&  9.3&  22.9\\
         &  SRPose&  \textbf{20.5}&  \textbf{43.1}&  \textbf{65.1}\\
    \bottomrule
    \end{tabular}
    \vspace{-2mm}
    \label{tab:megadepth}
\end{table}

\vspace{-2mm}
\subsection{Different Sparse Keypoint Detectors}
\vspace{-2mm}
SRPose can employ different kinds of methods as its sparse keypoint detector, including the classic method SIFT~\cite{lowe2004distinctive}, and the deep learning-based detectors, such as DISK~\cite{tyszkiewicz2020disk}, ALIKED~\cite{Zhao2023ALIKED}, SuperPoint~\cite{detone2018superpoint}, \etc.
Table \ref{tab:detectors} shows the performance of SRPose on Matterport using different methods as the sparse keypoint detectors. As a result, SuperPoint outperforms other detectors, which is the default detector we choose to evaluate our framework on other datasets.
\begin{table}
    \caption{\textbf{Relative pose estimation on Matterport~\cite{chang2017matterport3d} by SRPose with different sparse keypoint detectors.} SuperPoint~\cite{detone2018superpoint} outperforms other detectors.}
    \vspace{-2mm}
    \scriptsize
    \centering
    \tabcolsep=4pt
    \begin{tabular}{lcccccc}
    \toprule
        \multirow{2}{*}{Method} & \multicolumn{3}{c}{Rot. (\(^\circ\))}&  \multicolumn{3}{c}{Trans. (m)}\\
         & Med.\(\downarrow\)&  Avg.\(\downarrow\)&  \(\leq\ang{30}\uparrow\)&  Med.\(\downarrow\)&  Avg.\(\downarrow\)& \(\leq1\)m\(\uparrow\)\\
    \midrule
        8point~\cite{rockwell20228} & 8.01 & 19.13 & 85.4 & 0.64 & 1.01 & 67.4\\
    \midrule
        SRPose + SIFT~\cite{lowe2004distinctive} & 7.04 & 18.19 & 84.8 & 0.57 & 1.00 & 68.9\\
        SRPose + DISK~\cite{tyszkiewicz2020disk} & 7.73 & 19.96 & 84.0 & 0.60 & 1.03 & 66.6\\
        SRPose + ALIKED~\cite{zhao2022alike} & 5.65 & 15.92 & 88.1 & 0.50 & 0.89 & 73.3\\
        SRPose + SuperPoint~\cite{detone2018superpoint} & \textbf{5.57} & \textbf{14.32} & \textbf{88.9} & \textbf{0.47} & \textbf{0.84} & \textbf{74.2}\\
    \bottomrule
    \end{tabular}
    \vspace{-4mm}
    \label{tab:detectors}
\end{table}

\vspace{-3mm}
\section{Qualitative Results}
\label{sec:qua}
\begin{figure}[htbp]
    \scriptsize
	\centering
    \begin{tabular}{cccccc}
        Reference & Query & Ground Truth & Reference & Query & Ground Truth \\
        
        \includegraphics[width=0.15\linewidth]{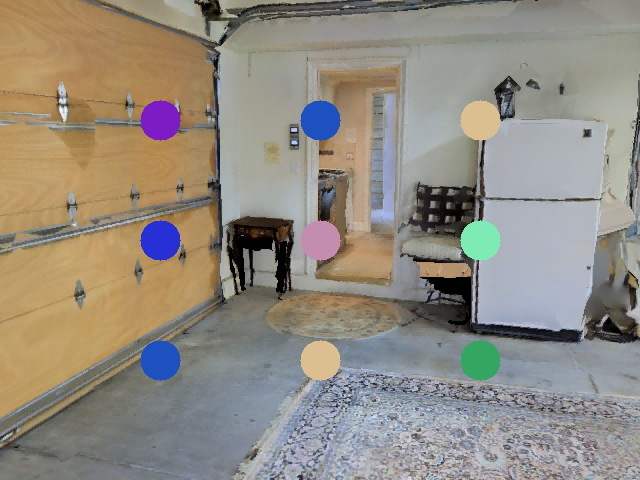} &
        \includegraphics[width=0.15\linewidth]{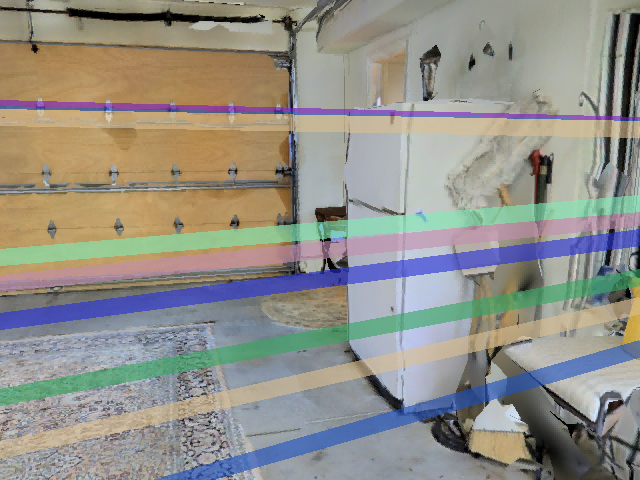} &
        \includegraphics[width=0.15\linewidth]{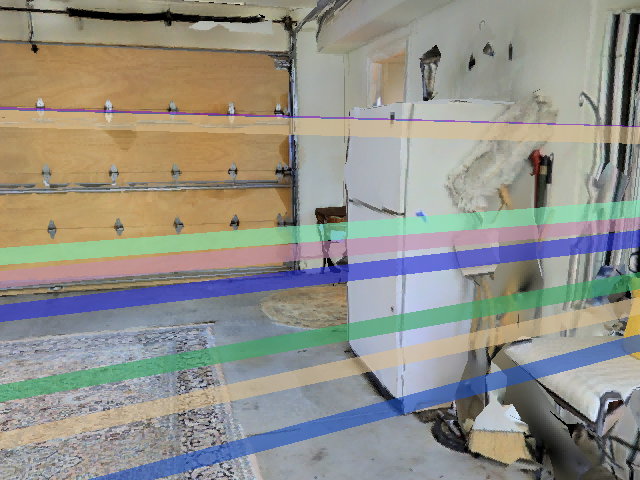} &
        \includegraphics[width=0.15\linewidth]{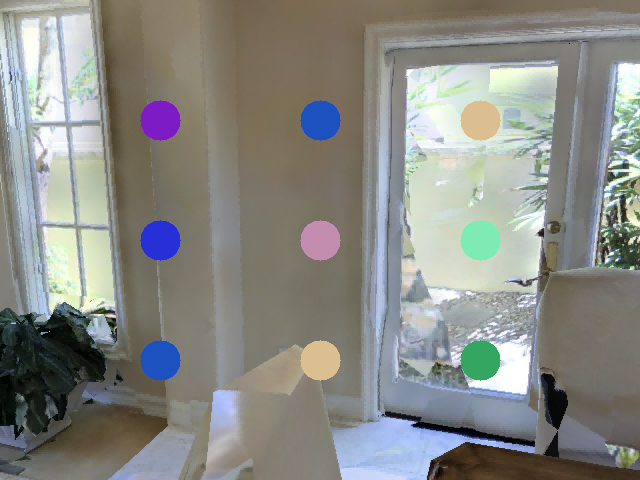} &
        \includegraphics[width=0.15\linewidth]{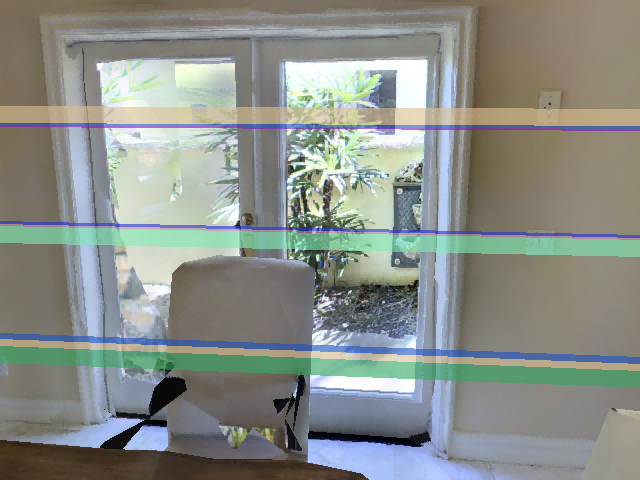} &
        \includegraphics[width=0.15\linewidth]{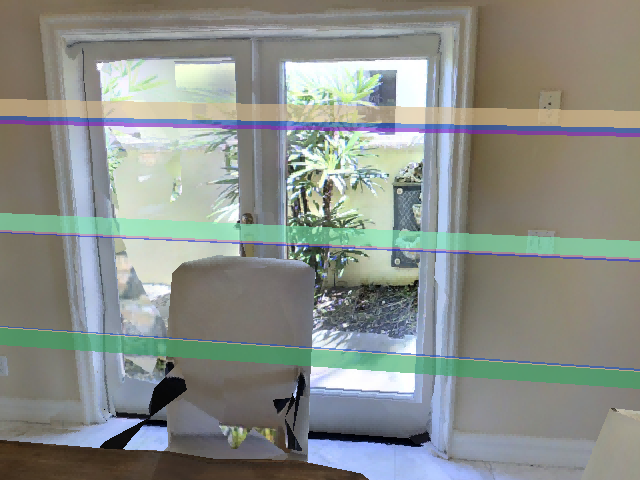} \\

        \includegraphics[width=0.15\linewidth]{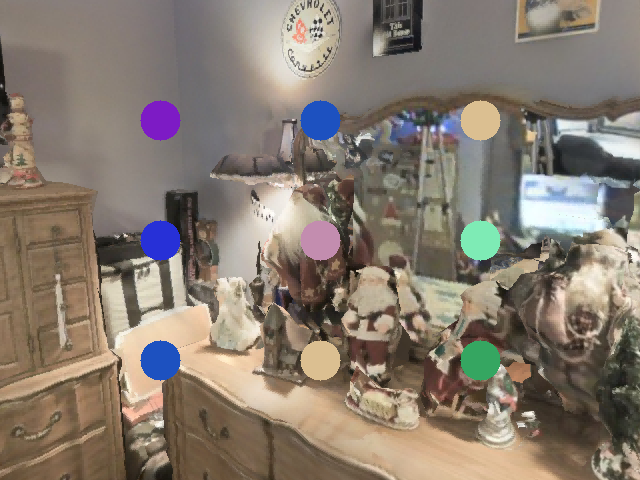} &
        \includegraphics[width=0.15\linewidth]{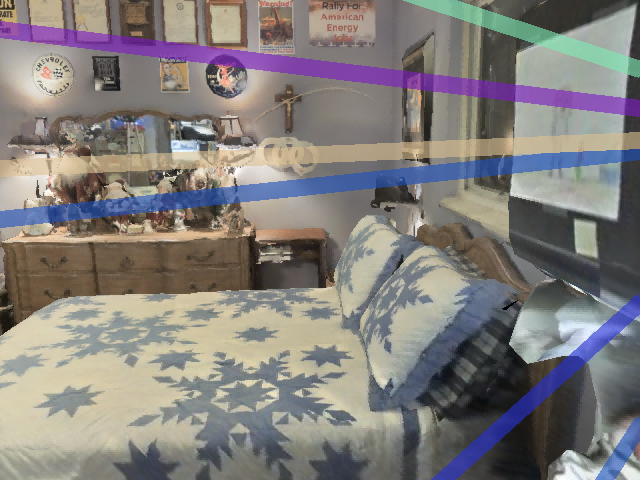} &
        \includegraphics[width=0.15\linewidth]{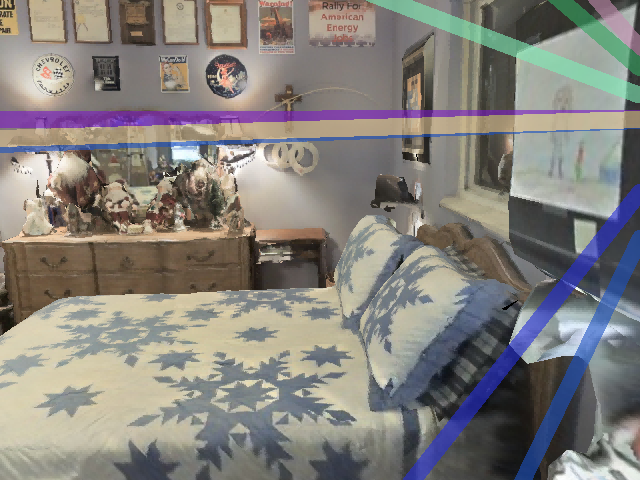} &
        \includegraphics[width=0.15\linewidth]{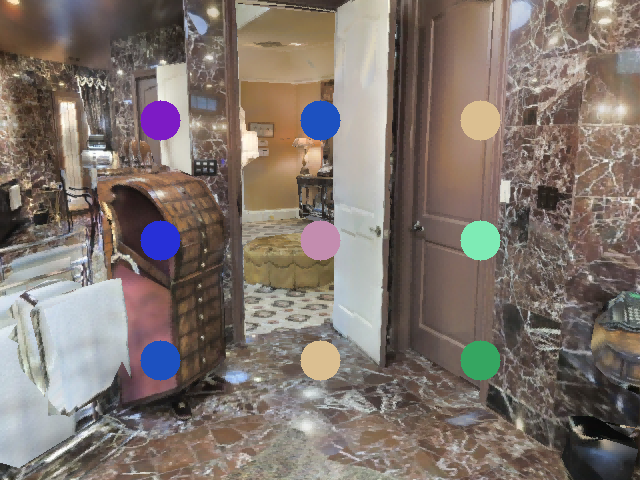} &
        \includegraphics[width=0.15\linewidth]{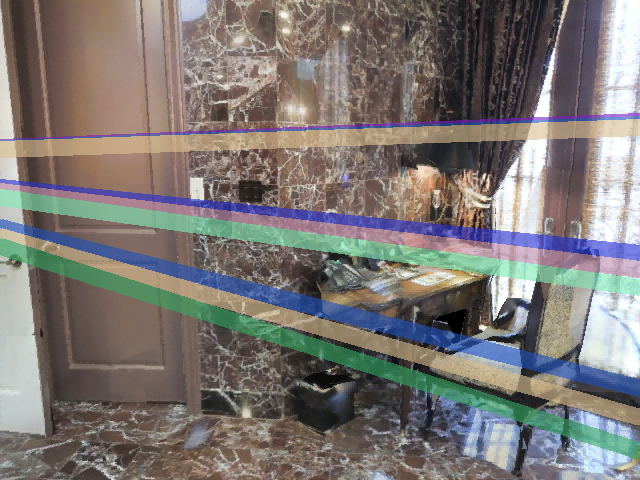} &
        \includegraphics[width=0.15\linewidth]{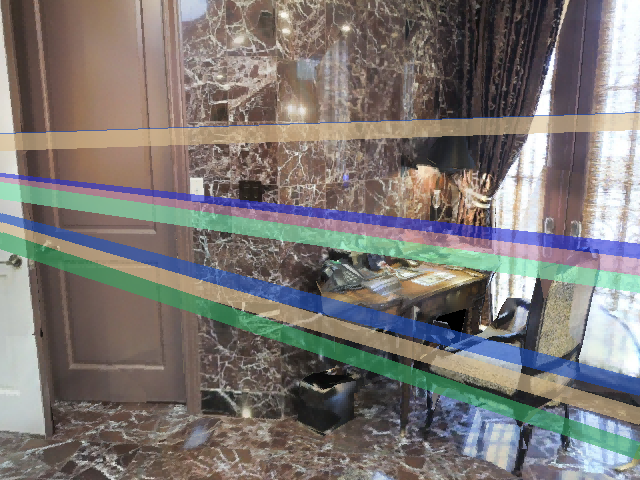} \\

        \includegraphics[width=0.15\linewidth]{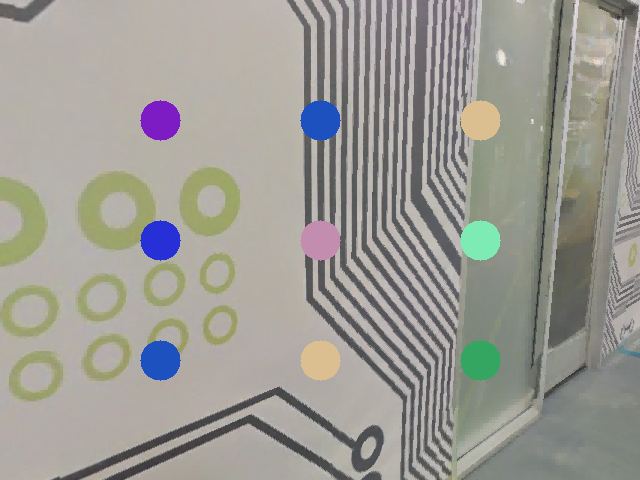} &
        \includegraphics[width=0.15\linewidth]{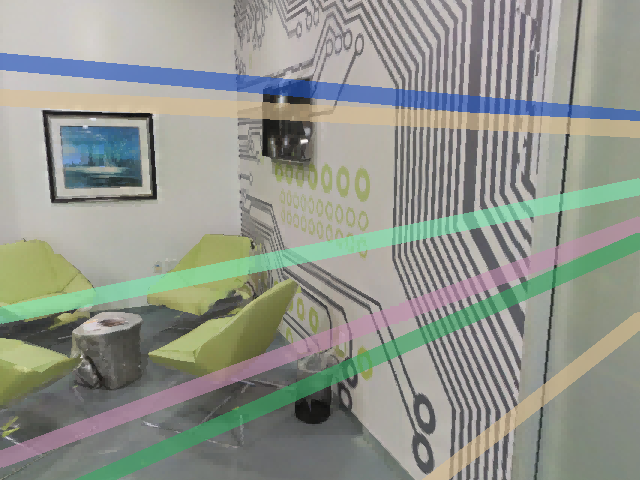} &
        \includegraphics[width=0.15\linewidth]{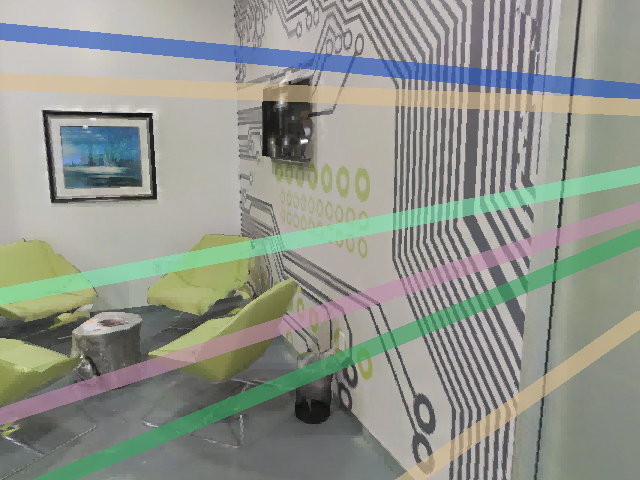} &
        \includegraphics[width=0.15\linewidth]{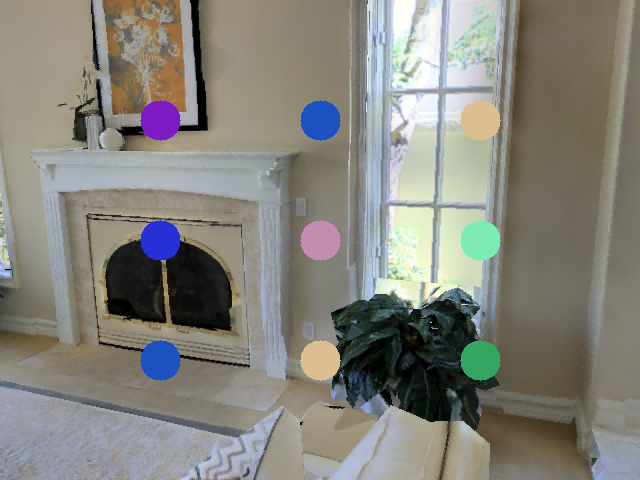} &
        \includegraphics[width=0.15\linewidth]{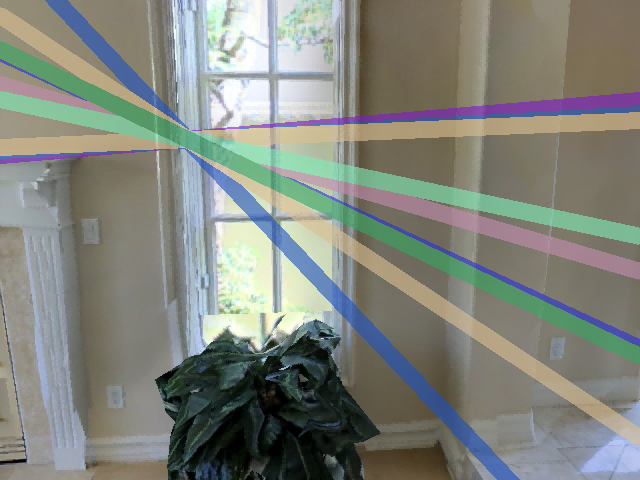} &
        \includegraphics[width=0.15\linewidth]{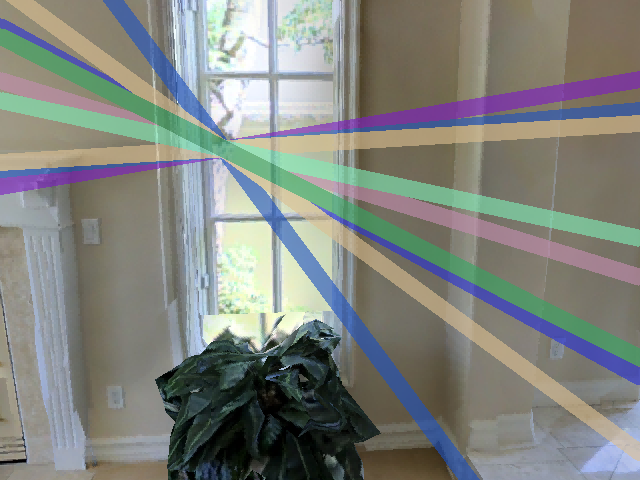} \\

        \includegraphics[width=0.15\linewidth]{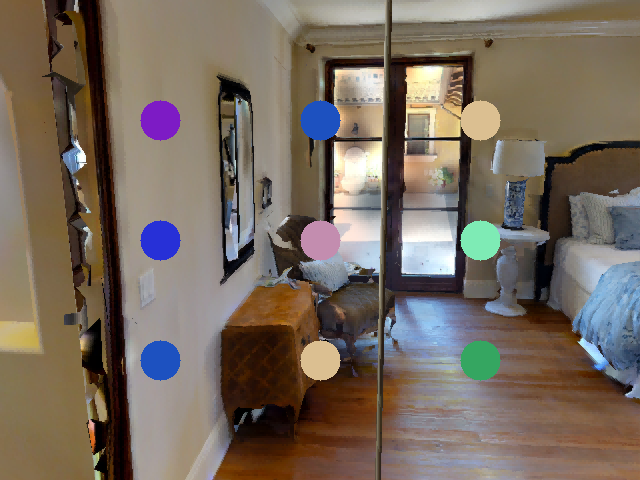} &
        \includegraphics[width=0.15\linewidth]{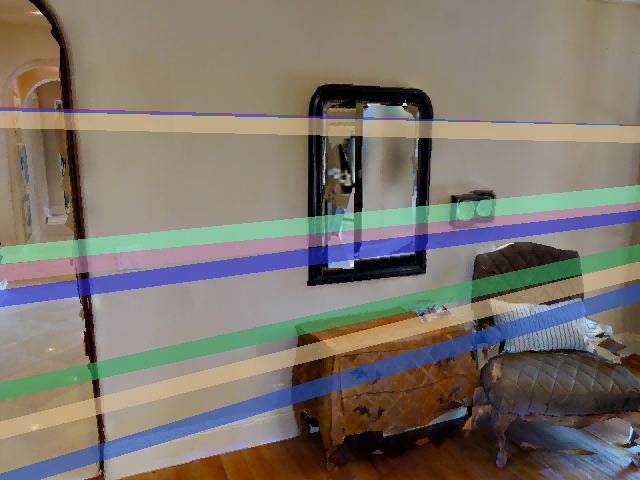} &
        \includegraphics[width=0.15\linewidth]{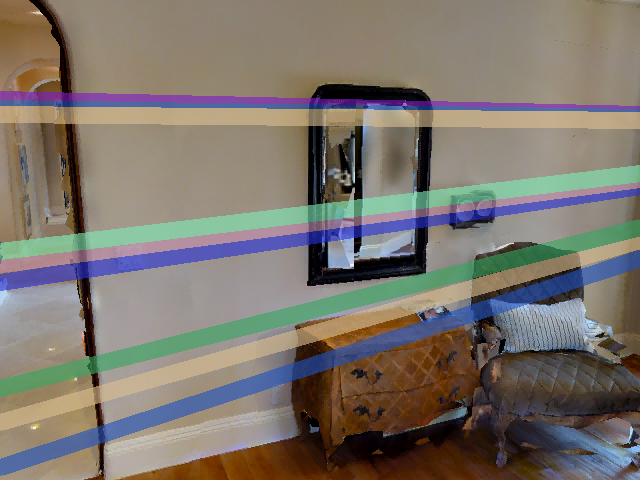} &
        \includegraphics[width=0.15\linewidth]{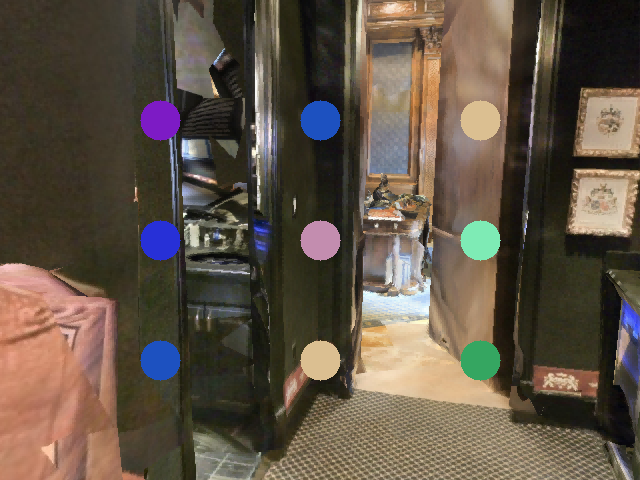} &
        \includegraphics[width=0.15\linewidth]{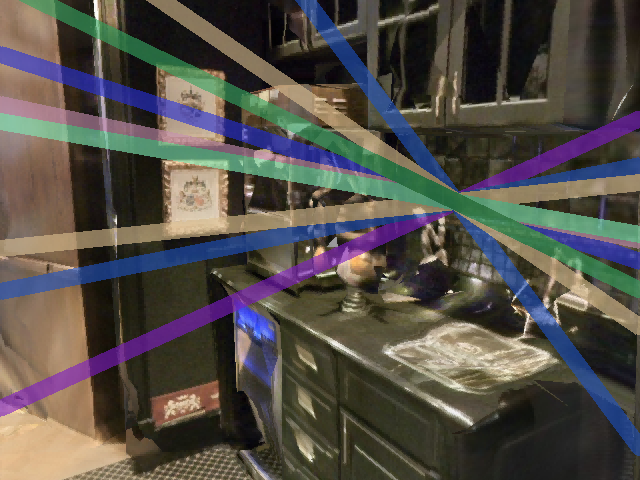} &
        \includegraphics[width=0.15\linewidth]{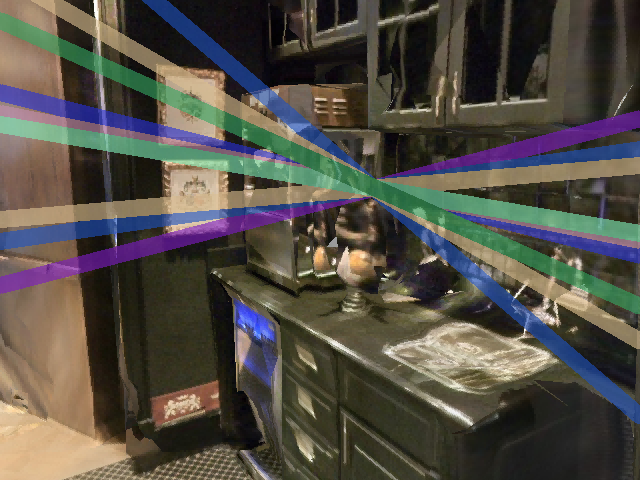} \\

        \includegraphics[width=0.15\linewidth]{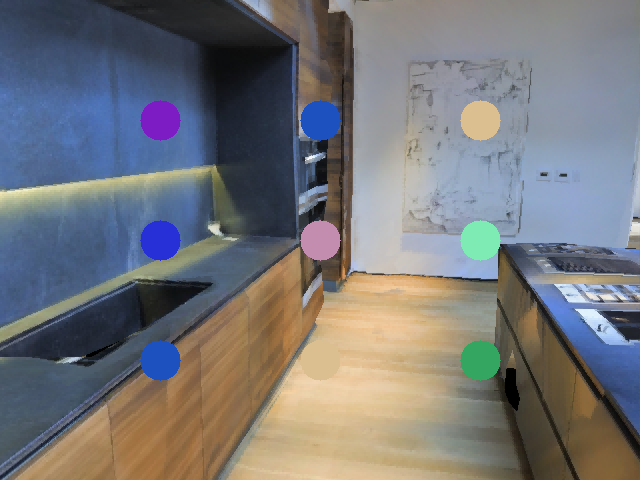} &
        \includegraphics[width=0.15\linewidth]{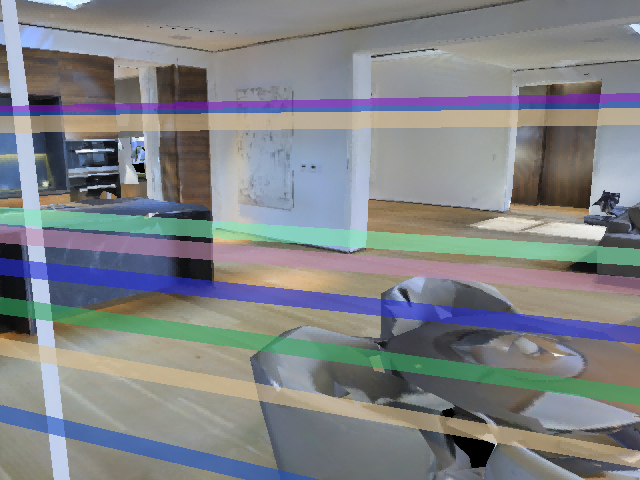} &
        \includegraphics[width=0.15\linewidth]{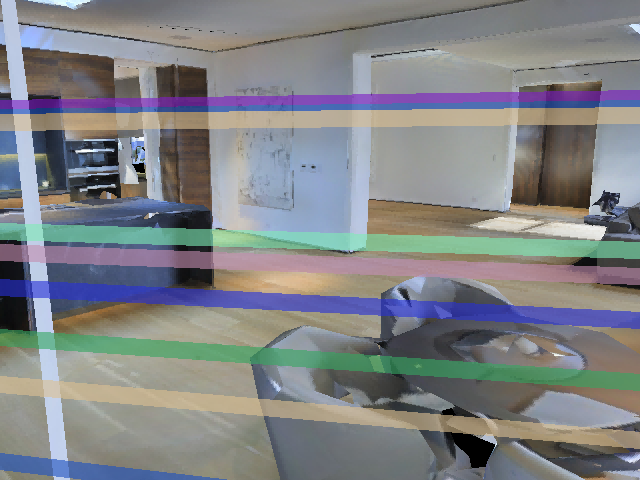} &
        \includegraphics[width=0.15\linewidth]{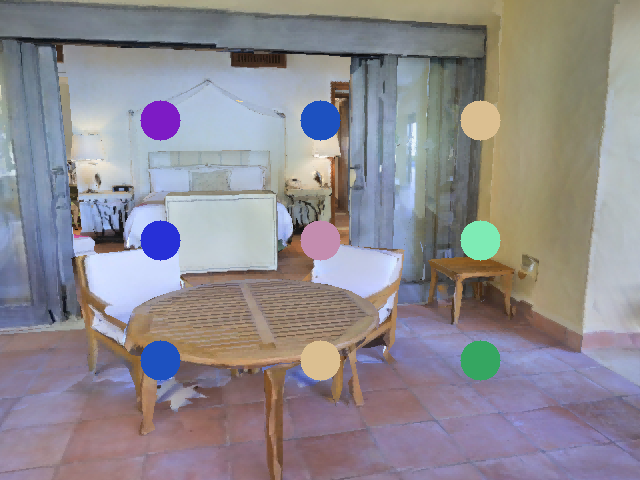} &
        \includegraphics[width=0.15\linewidth]{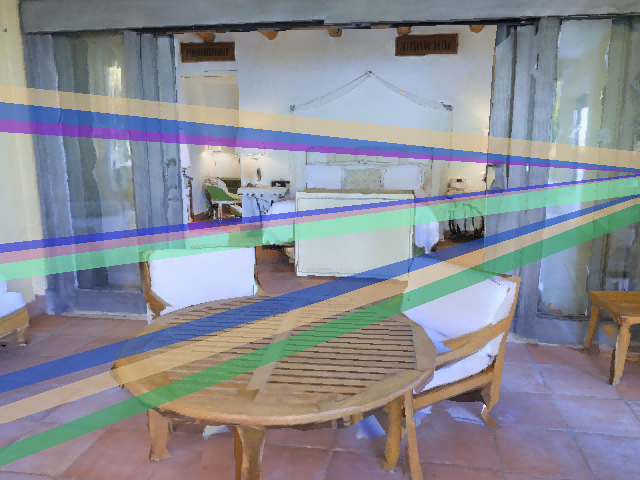} &
        \includegraphics[width=0.15\linewidth]{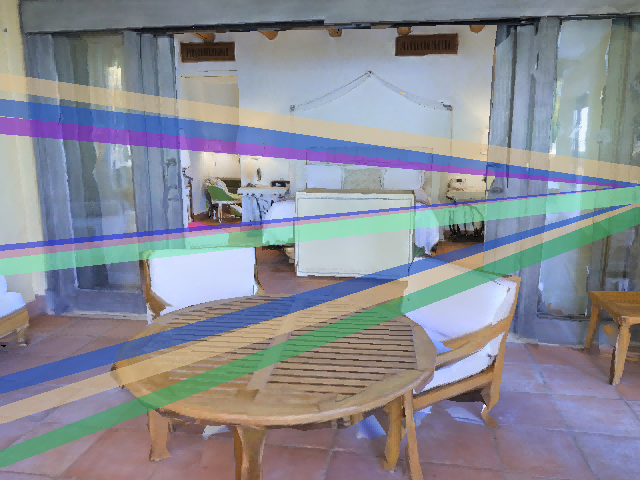} \\

        \includegraphics[width=0.15\linewidth]{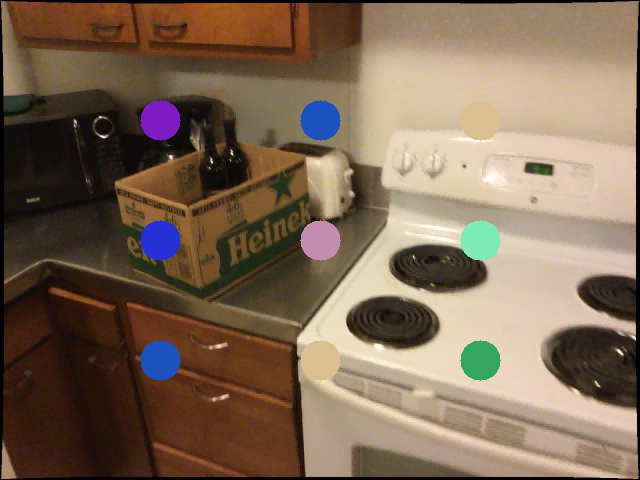} &
        \includegraphics[width=0.15\linewidth]{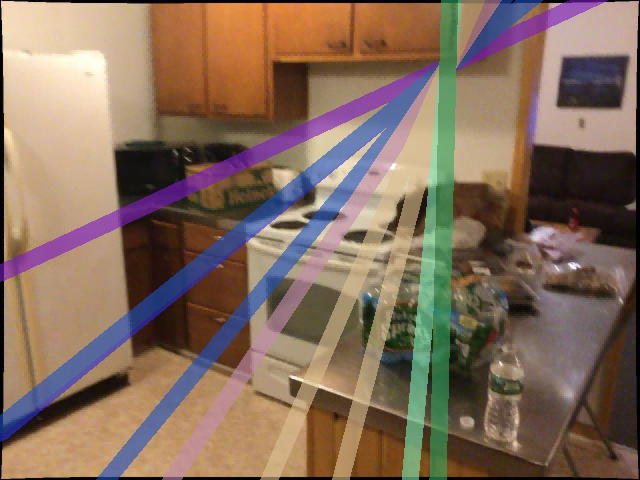} &
        \includegraphics[width=0.15\linewidth]{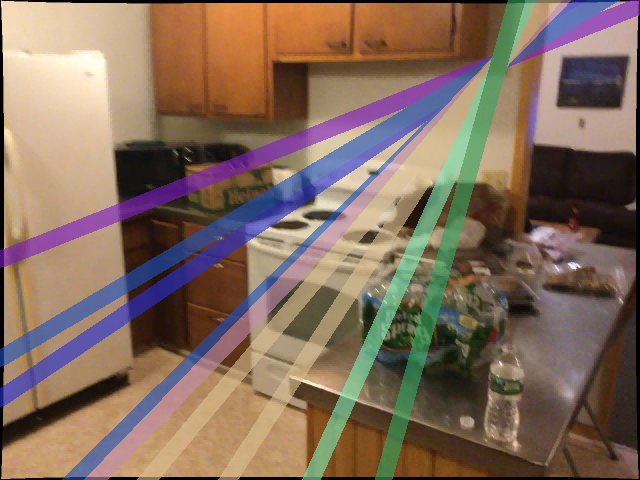} &
        \includegraphics[width=0.15\linewidth]{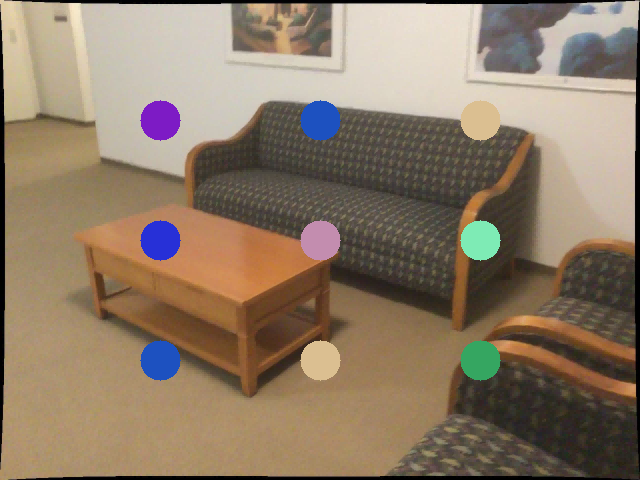} &
        \includegraphics[width=0.15\linewidth]{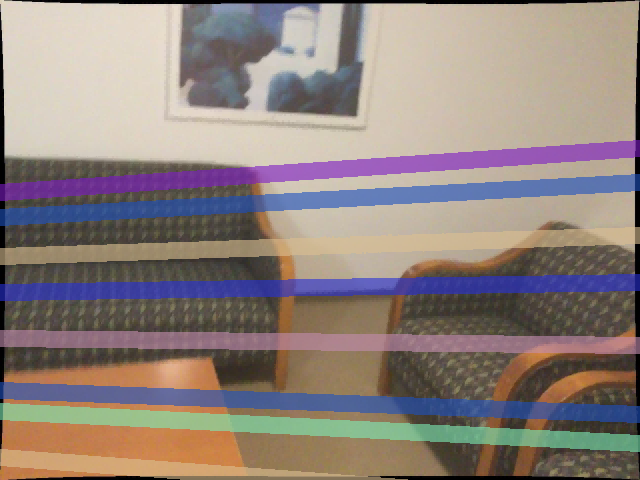} &
        \includegraphics[width=0.15\linewidth]{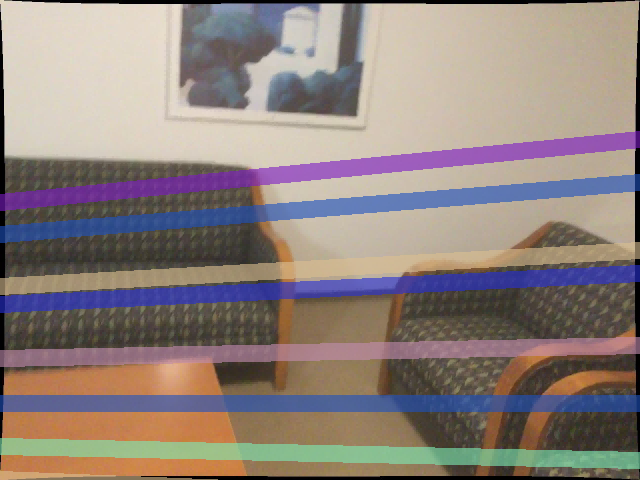} \\

        \includegraphics[width=0.15\linewidth]{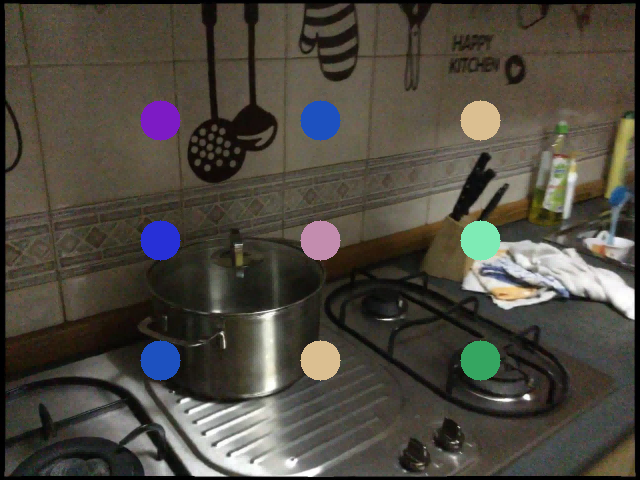} &
        \includegraphics[width=0.15\linewidth]{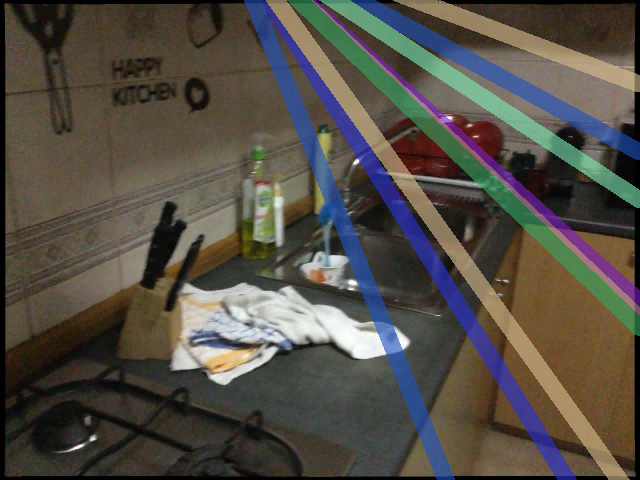} &
        \includegraphics[width=0.15\linewidth]{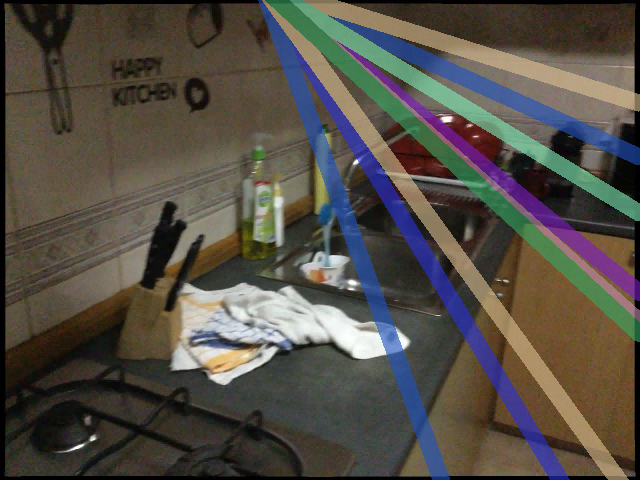} &
        \includegraphics[width=0.15\linewidth]{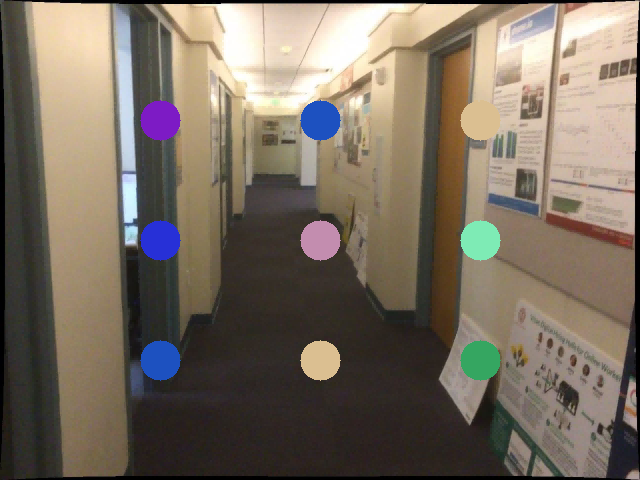} &
        \includegraphics[width=0.15\linewidth]{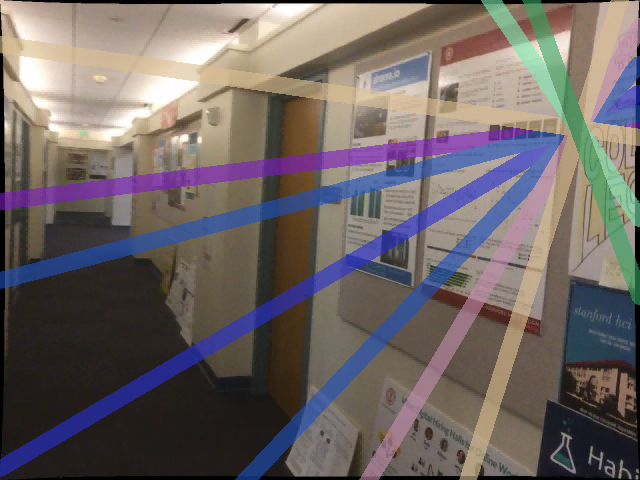} &
        \includegraphics[width=0.15\linewidth]{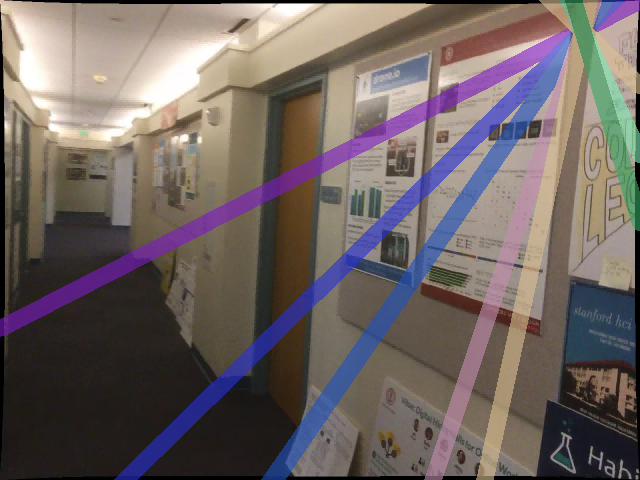} \\

        \includegraphics[width=0.15\linewidth]{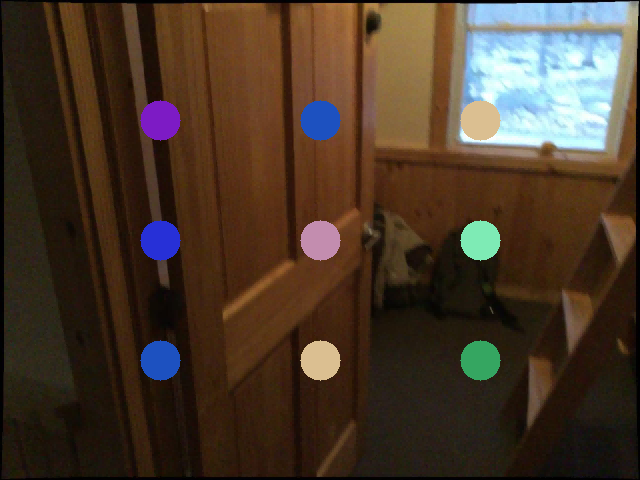} &
        \includegraphics[width=0.15\linewidth]{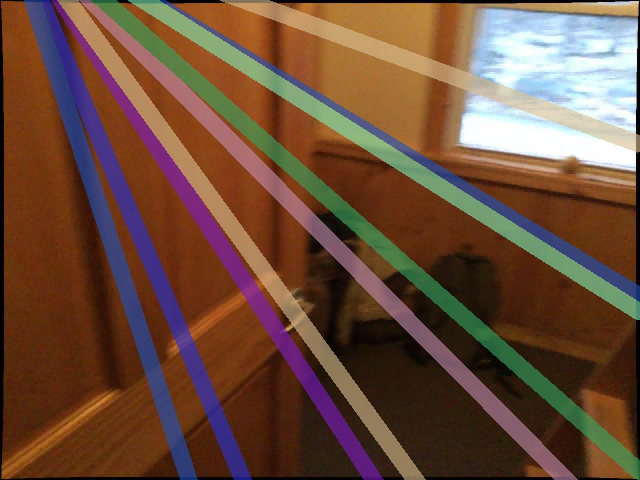} &
        \includegraphics[width=0.15\linewidth]{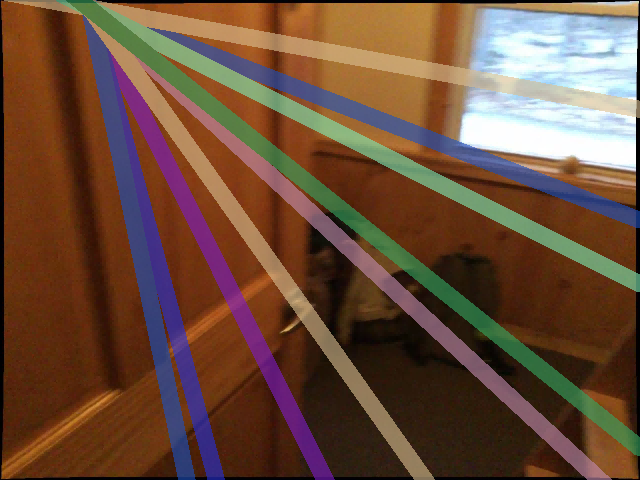} &
        \includegraphics[width=0.15\linewidth]{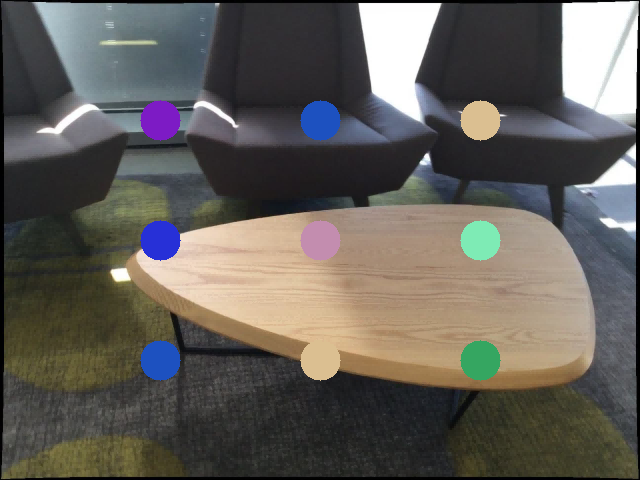} &
        \includegraphics[width=0.15\linewidth]{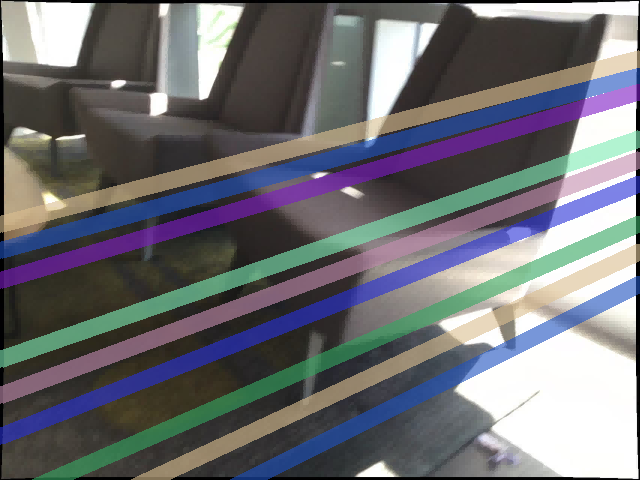} &
        \includegraphics[width=0.15\linewidth]{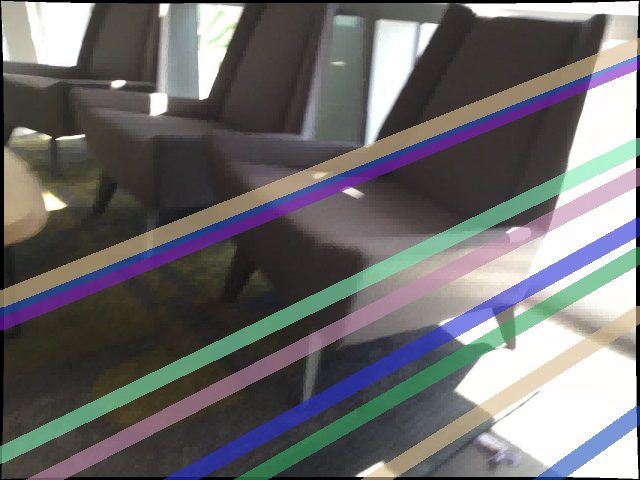} \\

        \includegraphics[width=0.15\linewidth]{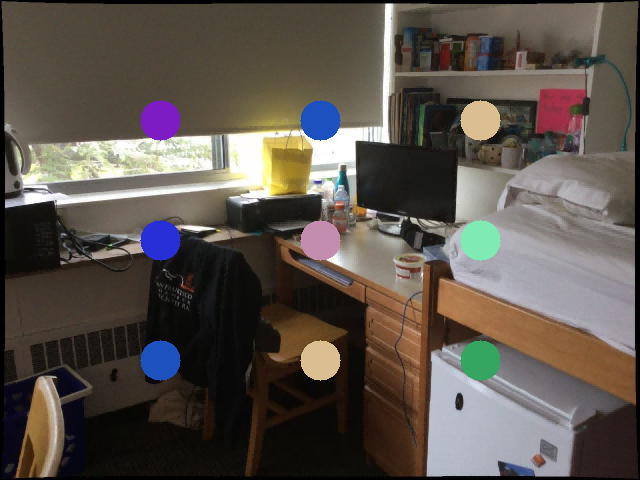} &
        \includegraphics[width=0.15\linewidth]{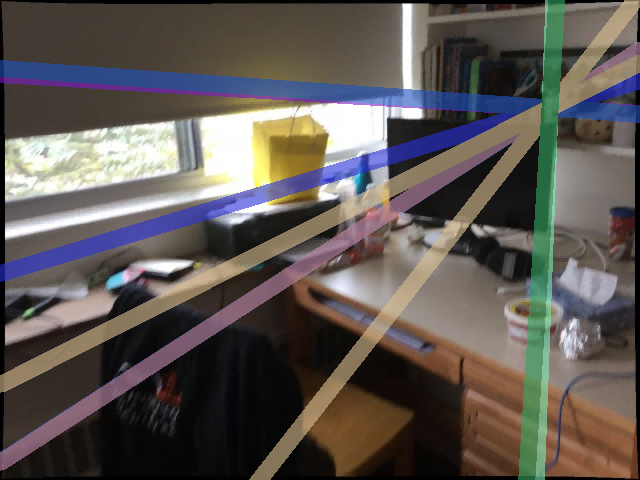} &
        \includegraphics[width=0.15\linewidth]{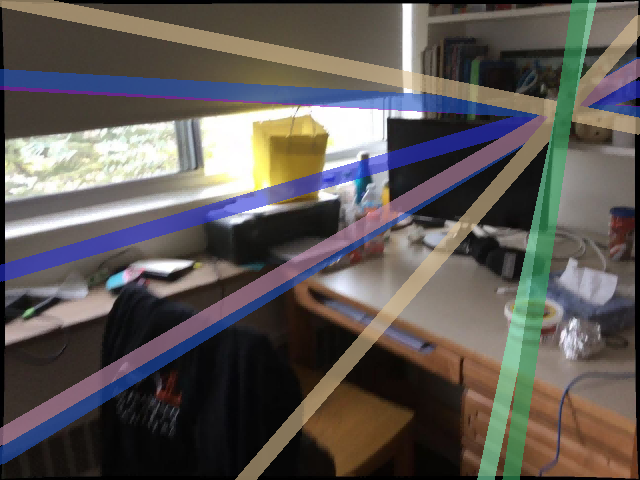} &
        \includegraphics[width=0.15\linewidth]{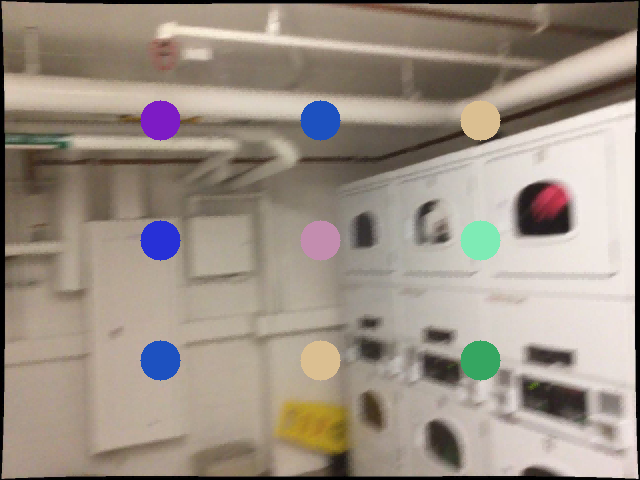} &
        \includegraphics[width=0.15\linewidth]{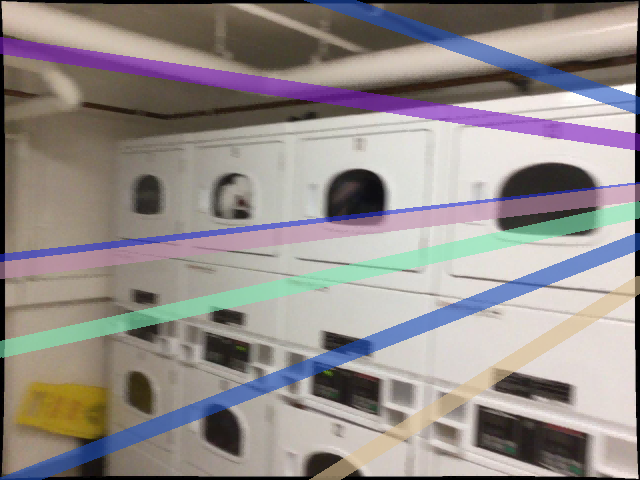} &
        \includegraphics[width=0.15\linewidth]{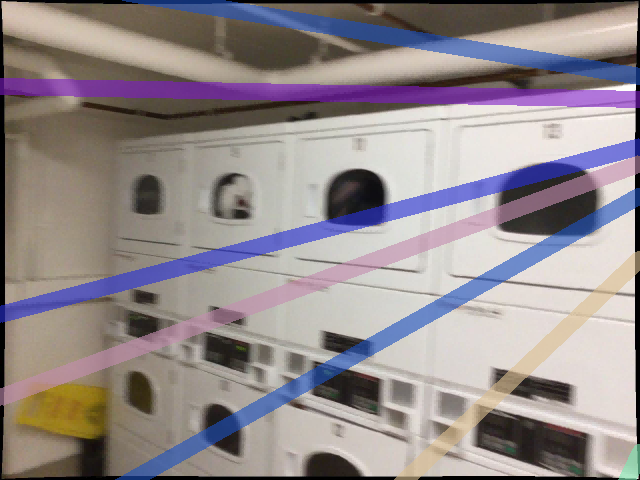} \\

        \includegraphics[width=0.15\linewidth]{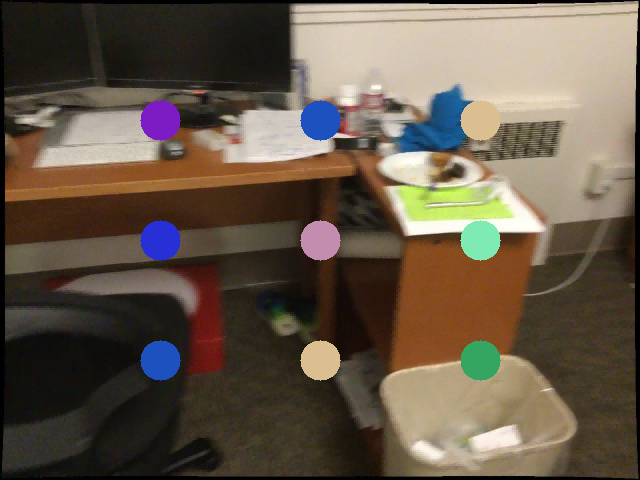} &
        \includegraphics[width=0.15\linewidth]{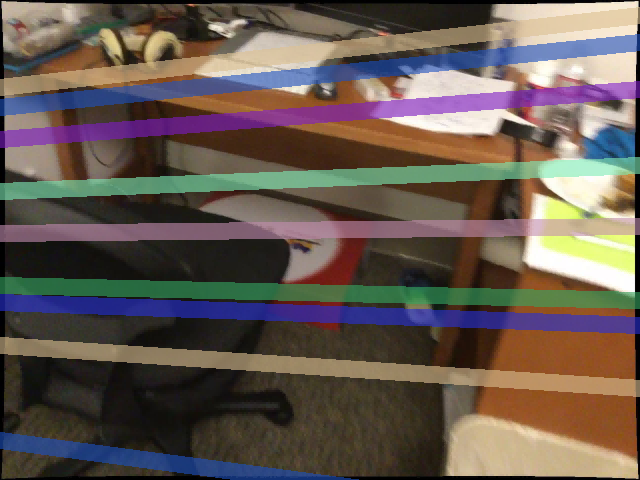} &
        \includegraphics[width=0.15\linewidth]{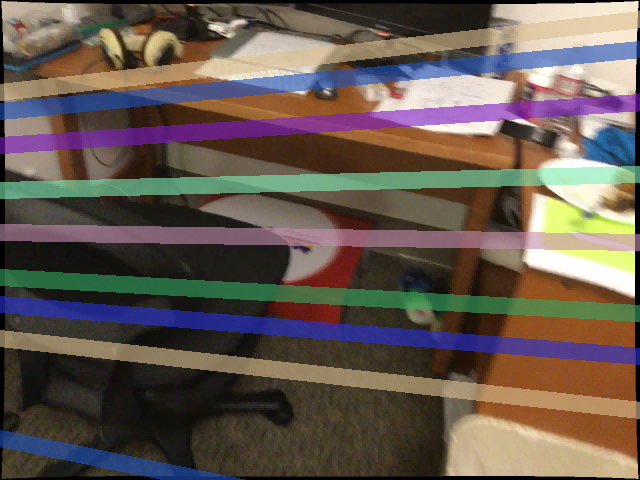} &
        \includegraphics[width=0.15\linewidth]{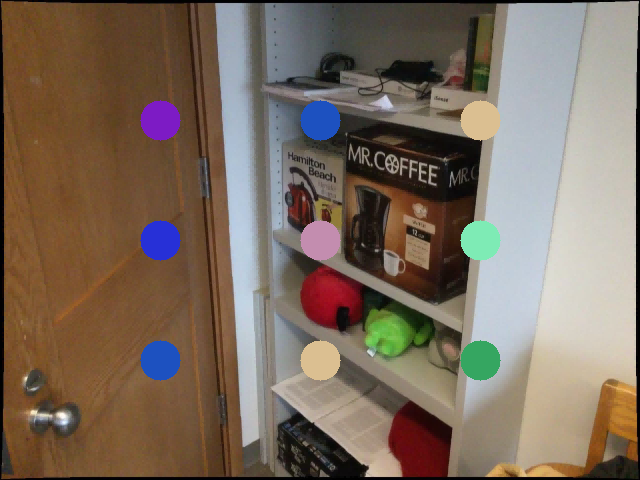} &
        \includegraphics[width=0.15\linewidth]{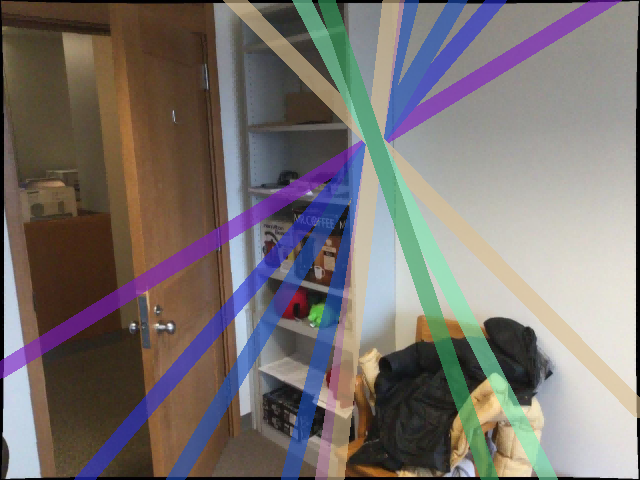} &
        \includegraphics[width=0.15\linewidth]{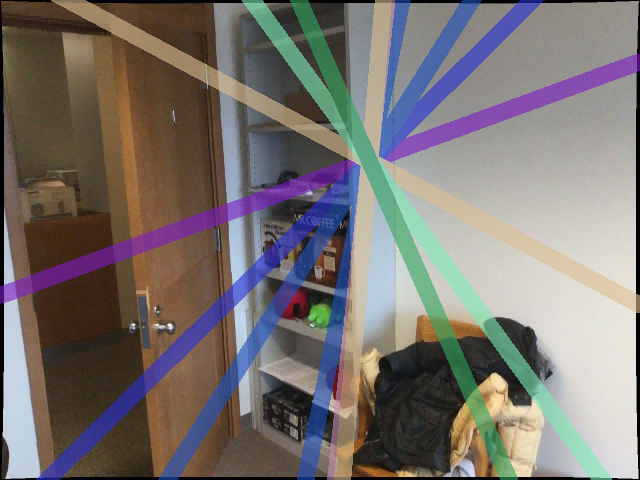} \\
        
    \end{tabular}
    \caption{\textbf{Relative pose estimation on Matterport~\cite{chang2017matterport3d} and ScanNet~\cite{dai2017scannet}}. The epipolar lines represent the connections of the nine points from the reference view to the query view, visualizing the predicted relative pose transformations.}
    \label{fig:qualitative-scene}
\end{figure}

\begin{figure}[htbp]
    \scriptsize
	\centering
    \begin{tabular}{cccc}
        Reference & Query & Reference & Query \\
        
        \includegraphics[width=0.22\linewidth]{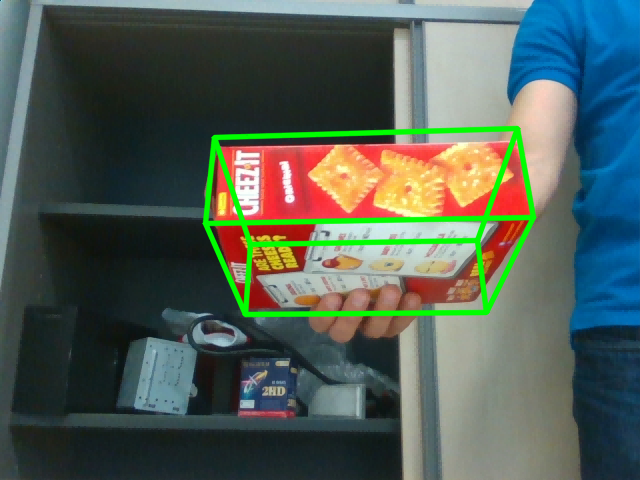} &
        \includegraphics[width=0.22\linewidth]{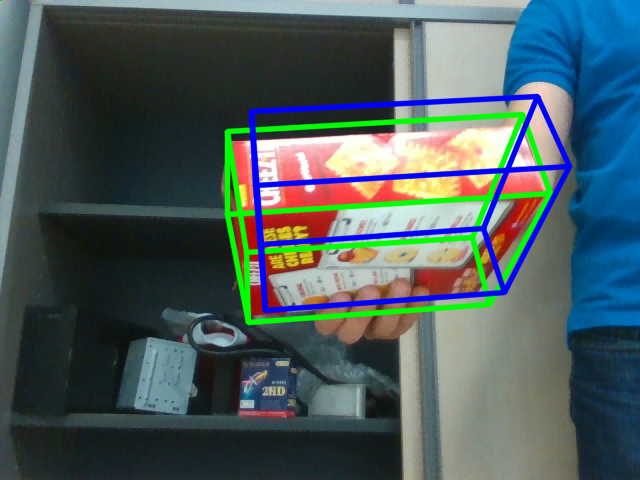} &
        \includegraphics[width=0.22\linewidth]{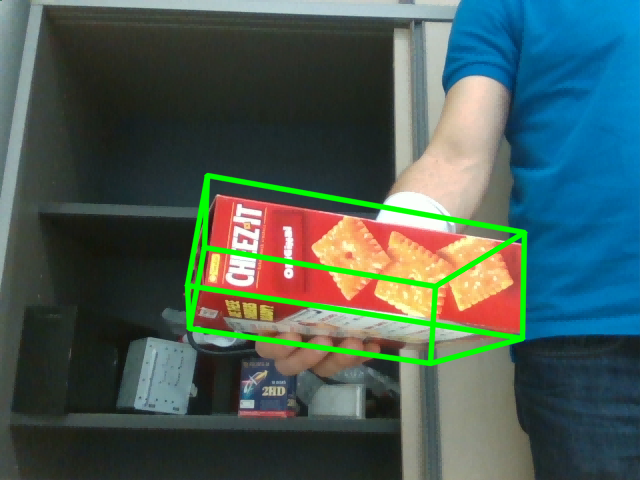} &
        \includegraphics[width=0.22\linewidth]{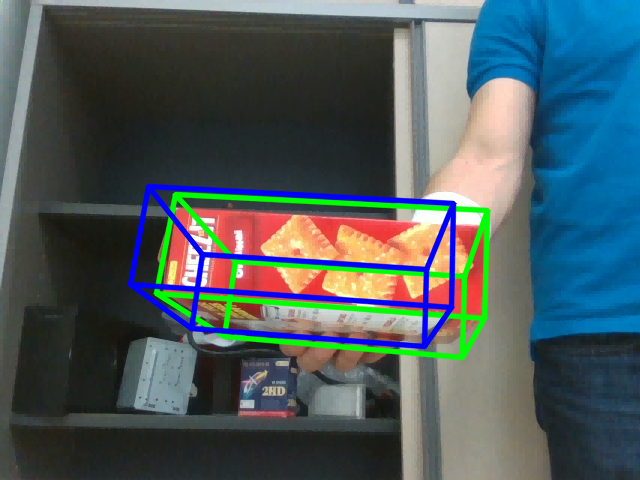} \\

        \includegraphics[width=0.22\linewidth]{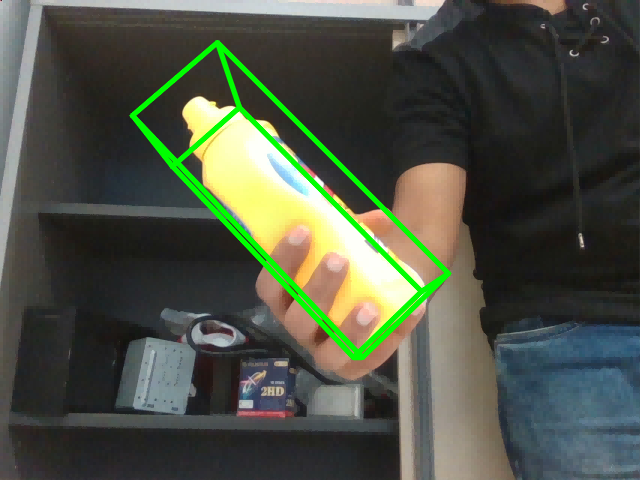} &
        \includegraphics[width=0.22\linewidth]{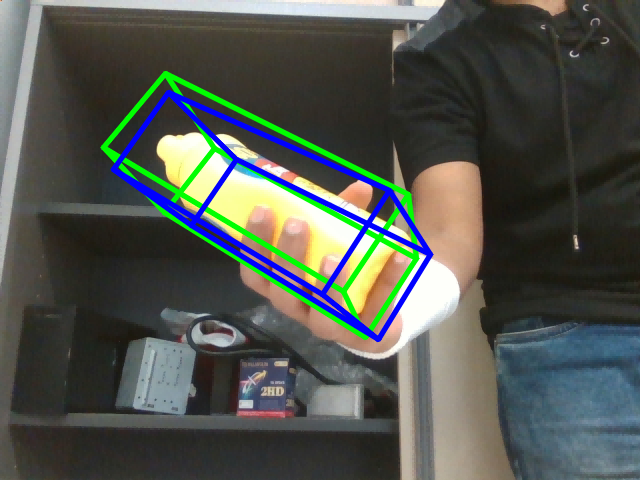} &
        \includegraphics[width=0.22\linewidth]{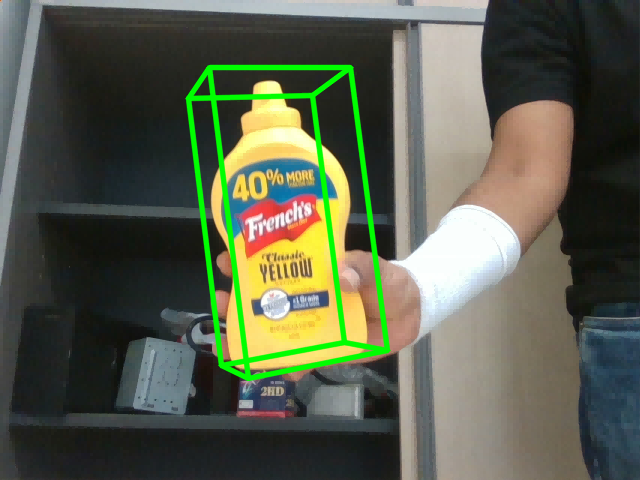} &
        \includegraphics[width=0.22\linewidth]{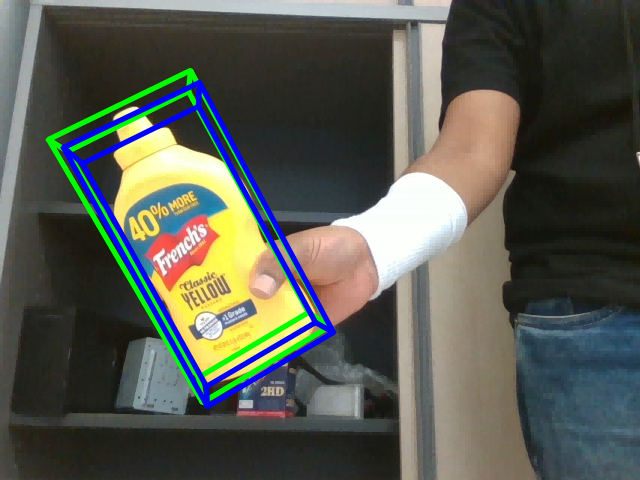} \\

        \includegraphics[width=0.22\linewidth]{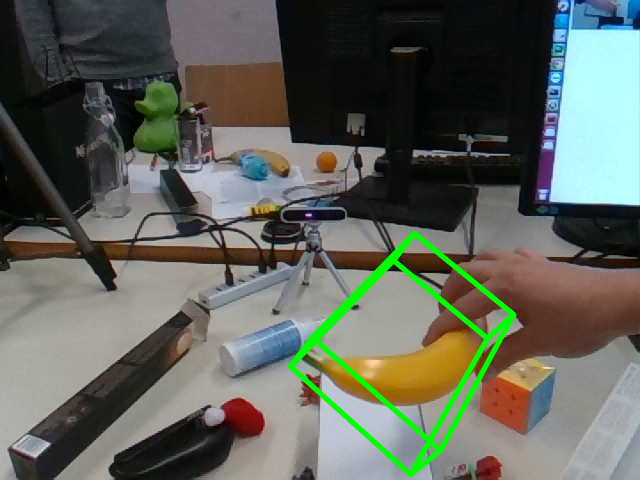} &
        \includegraphics[width=0.22\linewidth]{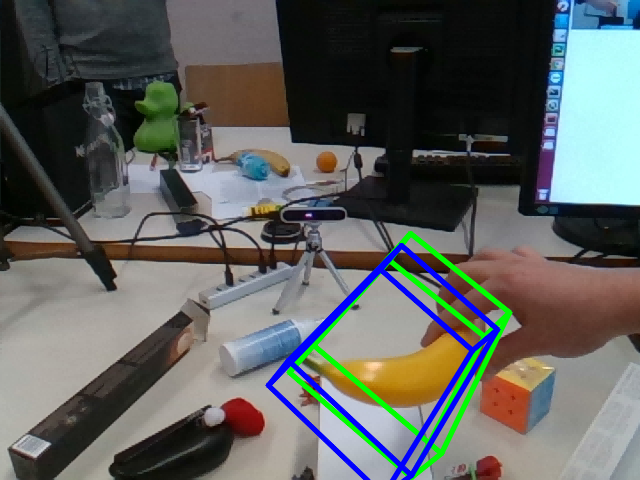} &
        \includegraphics[width=0.22\linewidth]{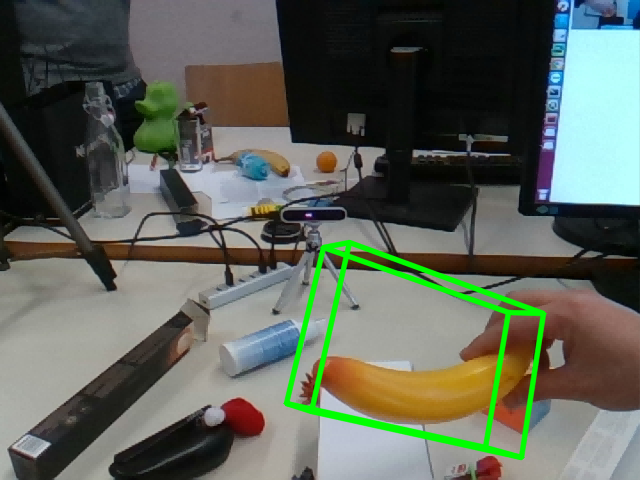} &
        \includegraphics[width=0.22\linewidth]{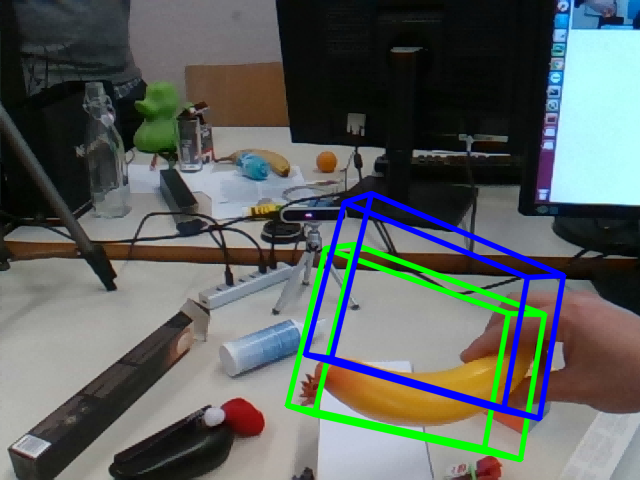} \\

        \includegraphics[width=0.22\linewidth]{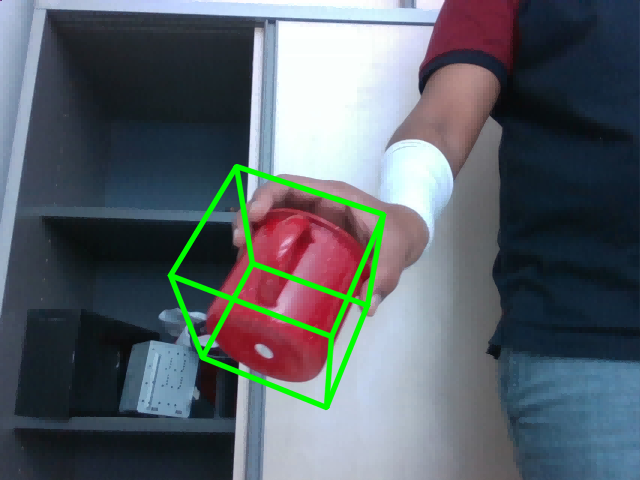} &
        \includegraphics[width=0.22\linewidth]{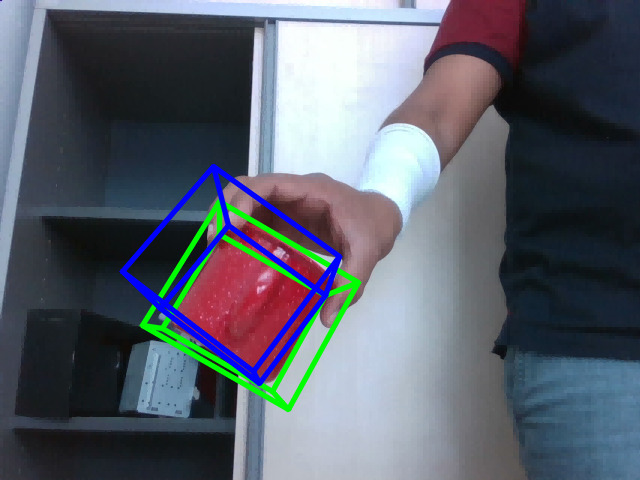} &
        \includegraphics[width=0.22\linewidth]{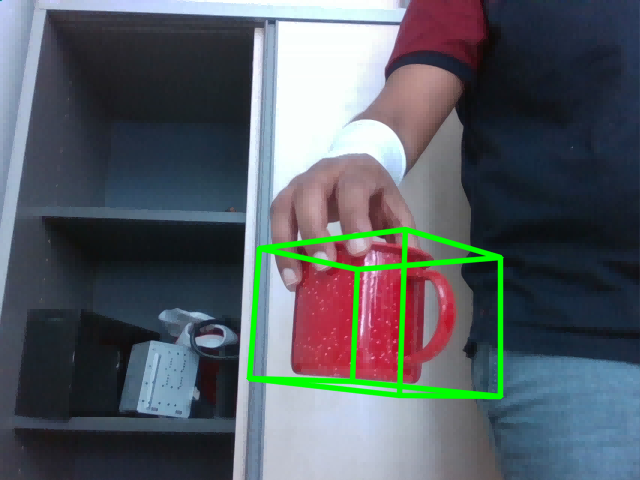} &
        \includegraphics[width=0.22\linewidth]{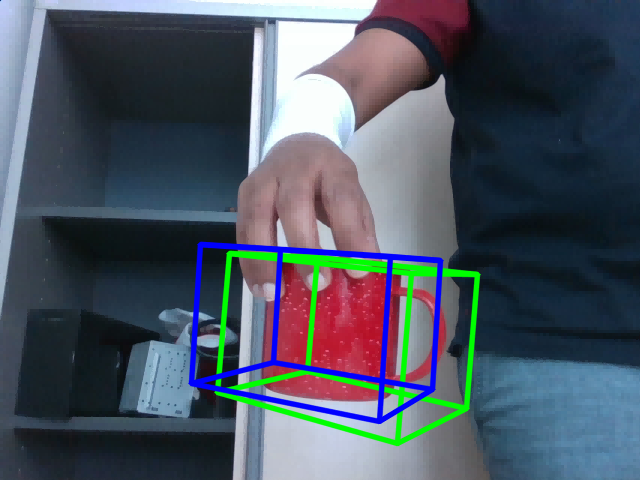} \\

        \includegraphics[width=0.22\linewidth]{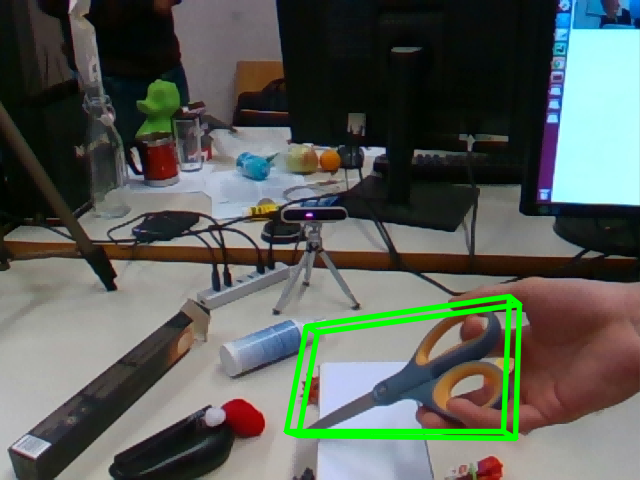} &
        \includegraphics[width=0.22\linewidth]{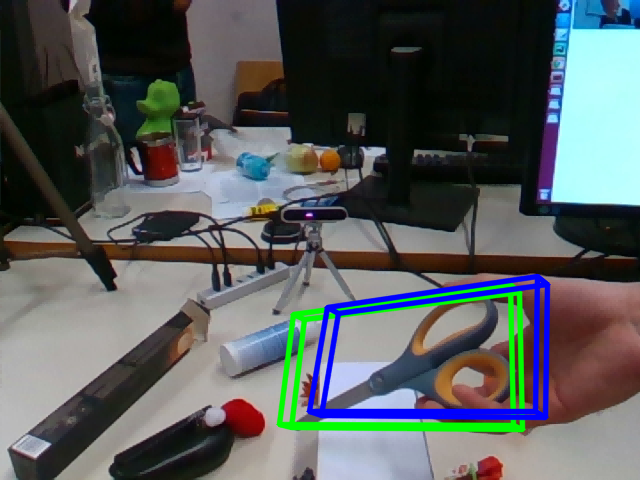} &
        \includegraphics[width=0.22\linewidth]{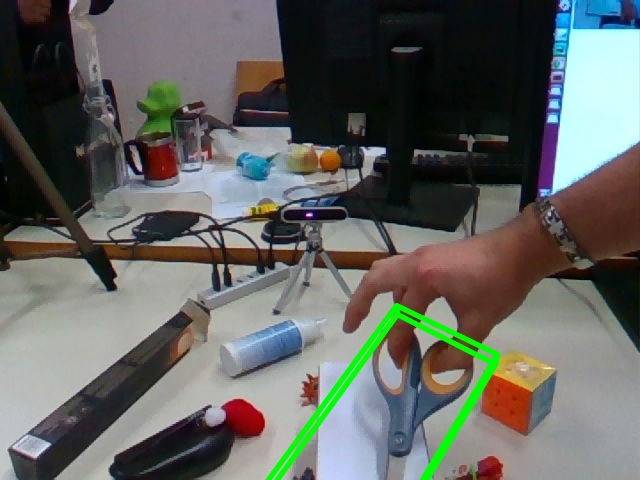} &
        \includegraphics[width=0.22\linewidth]{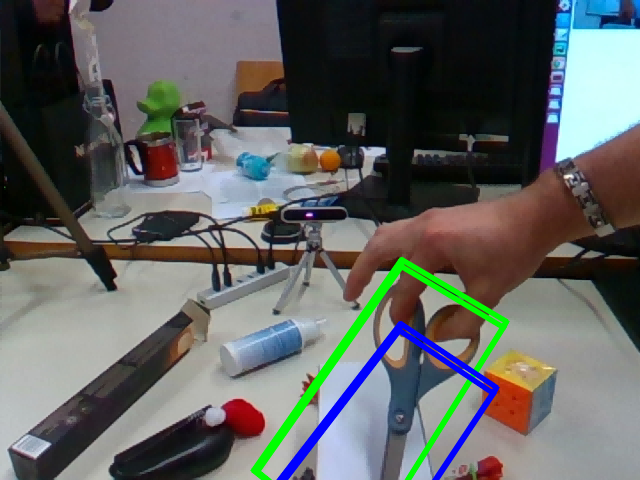} \\

        \includegraphics[width=0.22\linewidth]{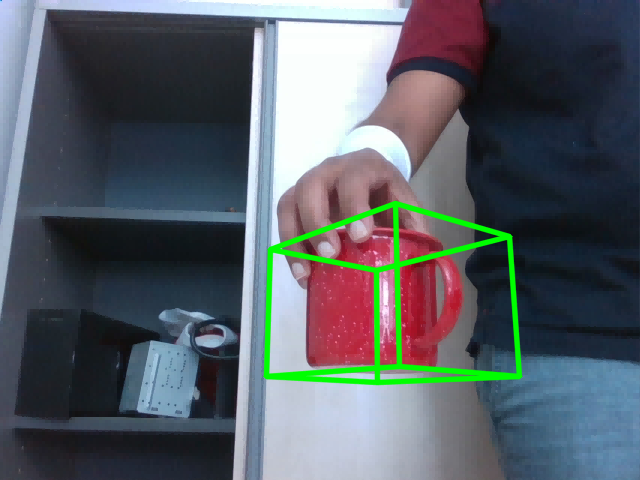} &
        \includegraphics[width=0.22\linewidth]{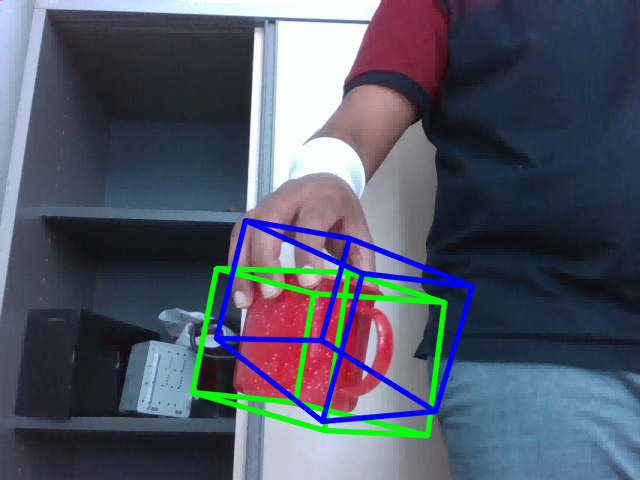} &
        \includegraphics[width=0.22\linewidth]{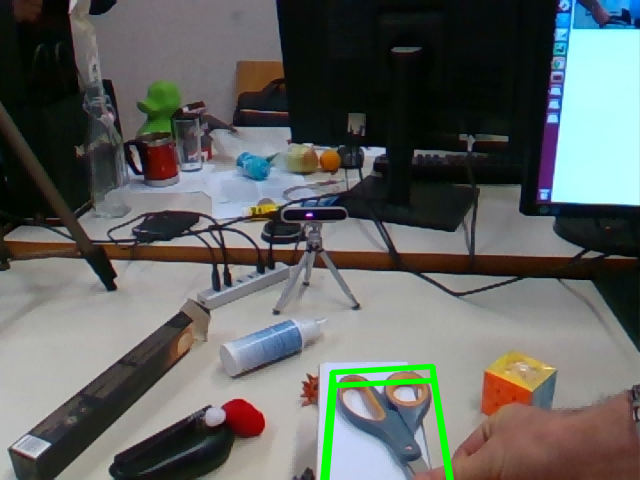} &
        \includegraphics[width=0.22\linewidth]{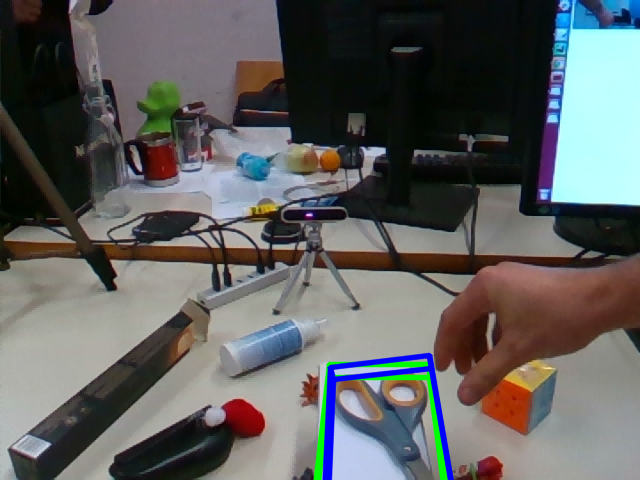} \\

        \includegraphics[width=0.22\linewidth]{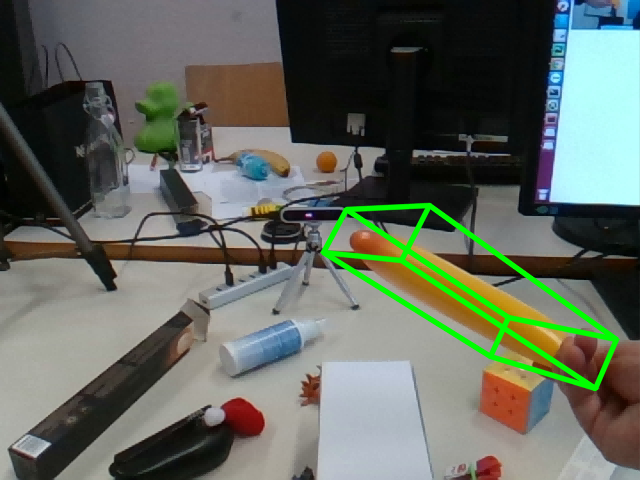} &
        \includegraphics[width=0.22\linewidth]{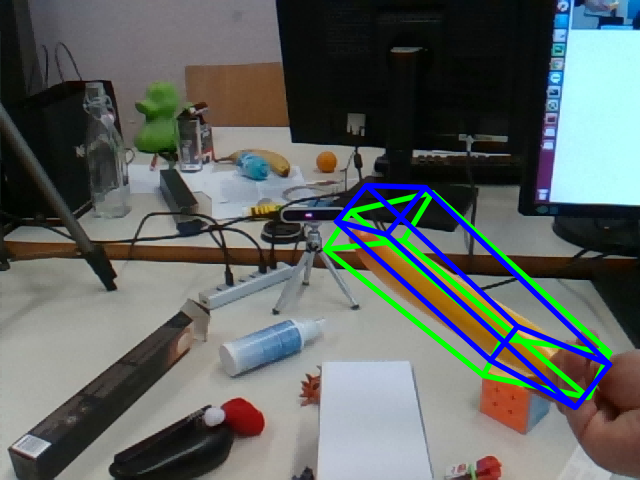} &
        \includegraphics[width=0.22\linewidth]{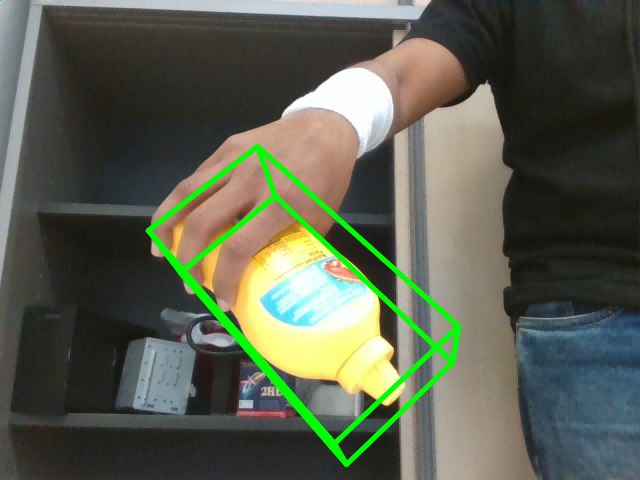} &
        \includegraphics[width=0.22\linewidth]{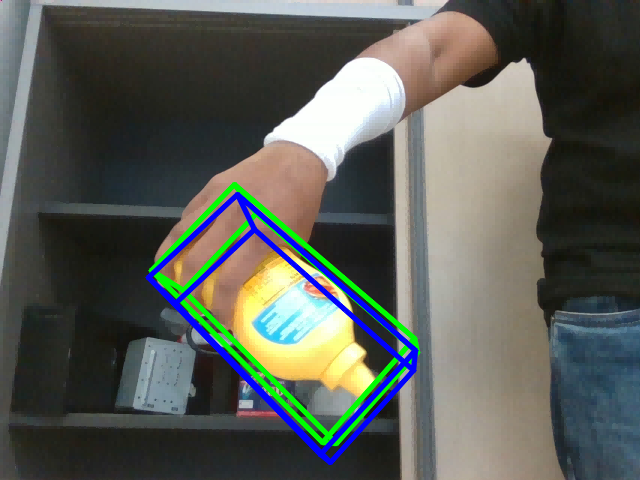} \\

        \includegraphics[width=0.22\linewidth]{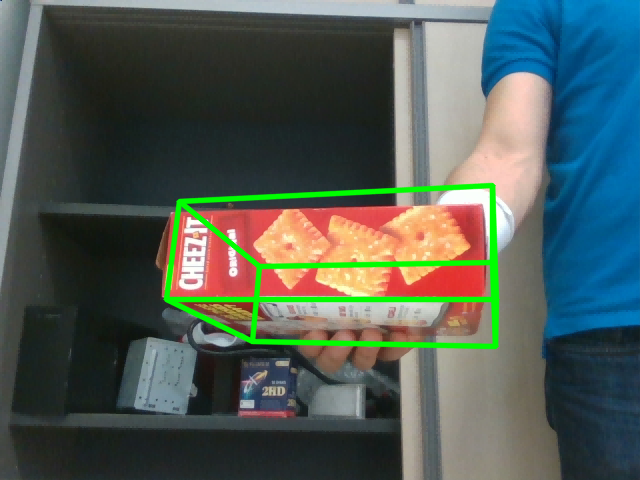} &
        \includegraphics[width=0.22\linewidth]{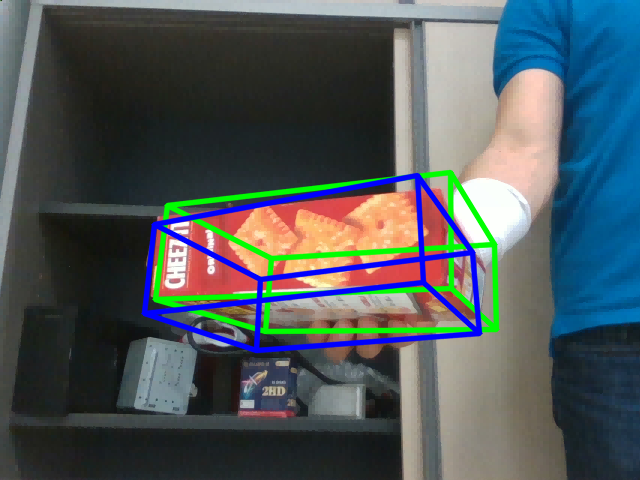} &
        \includegraphics[width=0.22\linewidth]{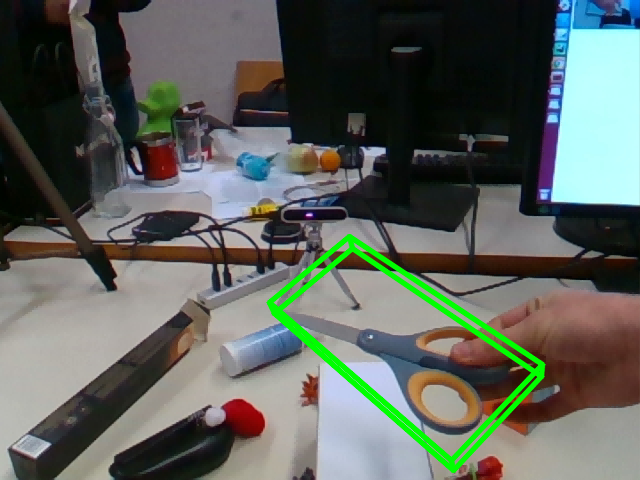} &
        \includegraphics[width=0.22\linewidth]{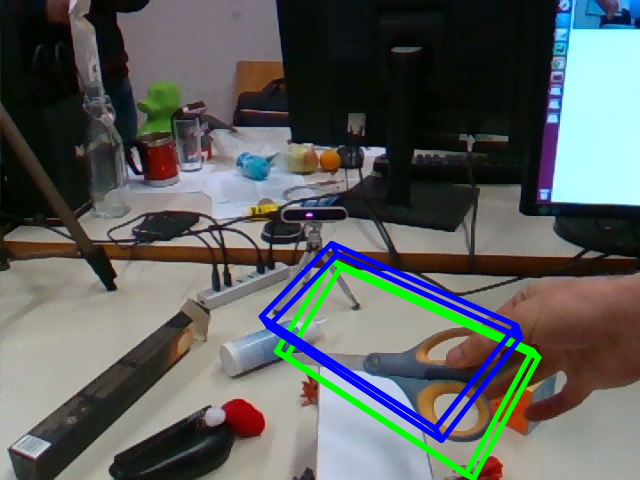} \\
        
    \end{tabular}
    \caption{\textbf{Relative 6D object pose estimation on HO3D~\cite{hampali2020honnotate}}. Ground-truth object poses are drawn in green, while the estimated poses are drawn in blue.}
    \label{fig:qualitative-ho3d}
\end{figure}

\vspace{-2mm}
\cref{fig:qualitative-scene} shows the qualitative results of camera-to-world pose estimation by SRPose on Matterport~\cite{chang2017matterport3d} and ScanNet~\cite{dai2017scannet}. And \cref{fig:qualitative-ho3d} shows the qualitative results of object-to-camera pose estimation by our framework on HO3D~\cite{hampali2020honnotate}.

\vspace{-3mm}
\section{Visualization}
\label{sec:vis}
\begin{figure}[htbp]
	\centering
    \begin{tabular}{ccccc}
        \scriptsize
        \rotatebox[origin=l]{90}{\hspace{3mm}Cross-attn.} & 
    	\includegraphics[width=0.22\linewidth]{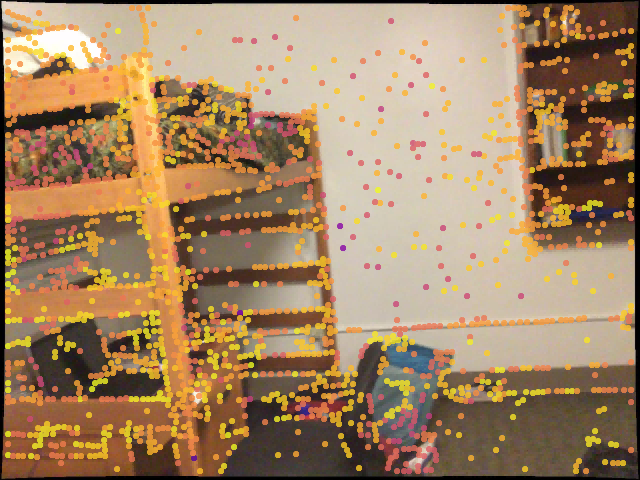} &
        \includegraphics[width=0.22\linewidth]{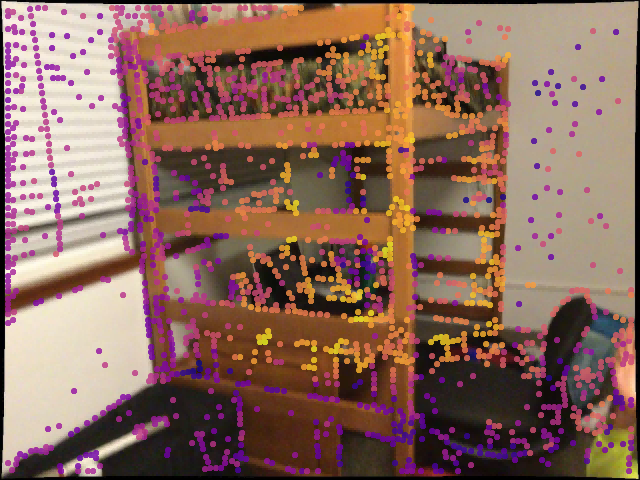} &
        \includegraphics[width=0.22\linewidth]{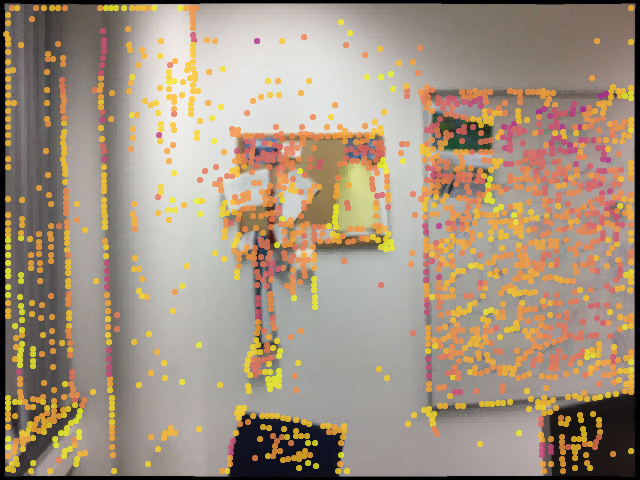} &
        \includegraphics[width=0.22\linewidth]{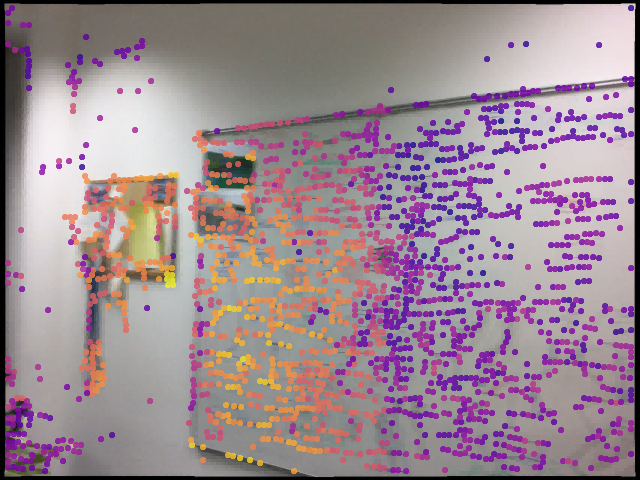} \\

        \scriptsize
        \rotatebox[origin=l]{90}{\hspace{3mm}Similarity} & 
    	\includegraphics[width=0.22\linewidth]{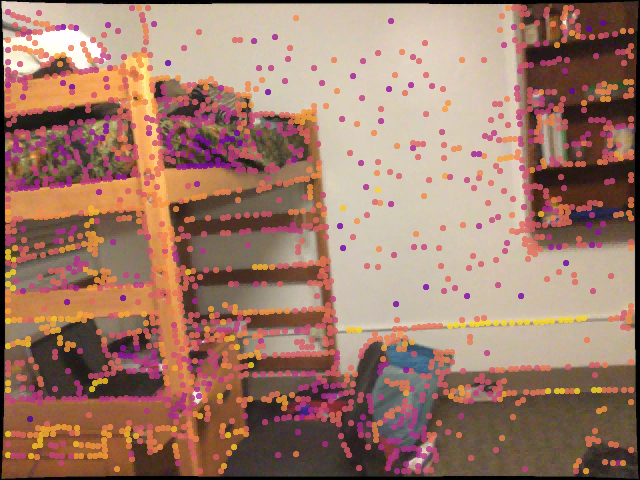} &
        \includegraphics[width=0.22\linewidth]{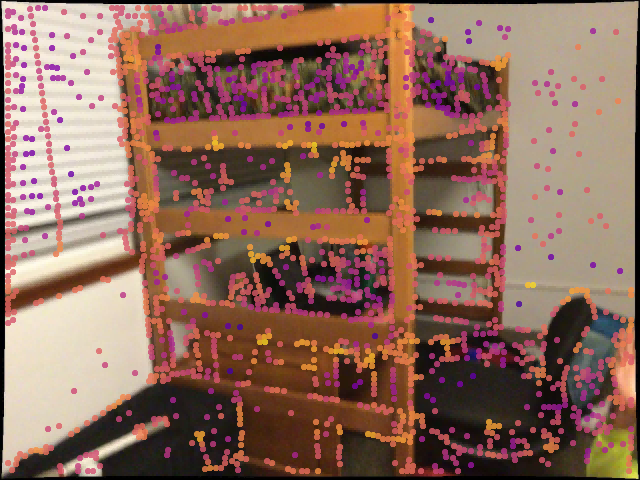} &
        \includegraphics[width=0.22\linewidth]{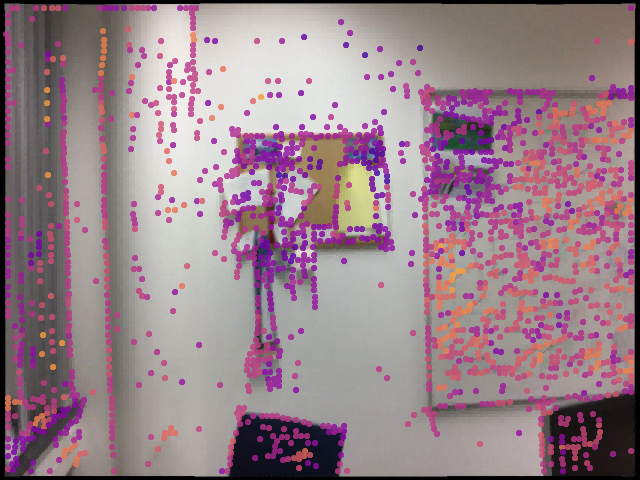} &
        \includegraphics[width=0.22\linewidth]{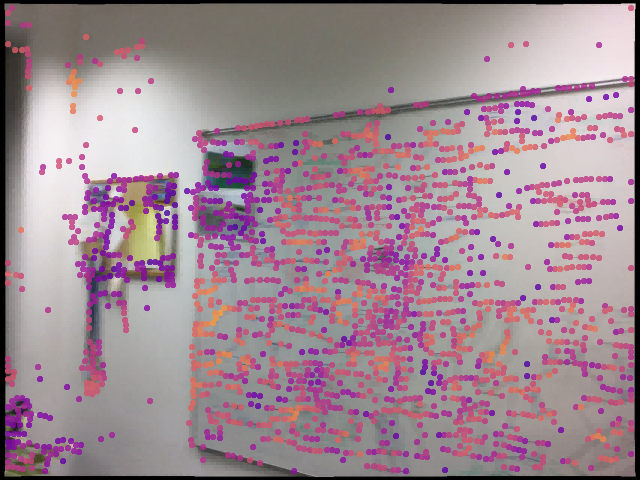} \\

        \scriptsize
        \rotatebox[origin=l]{90}{\hspace{3mm}Cross-attn.} & 
    	\includegraphics[width=0.22\linewidth]{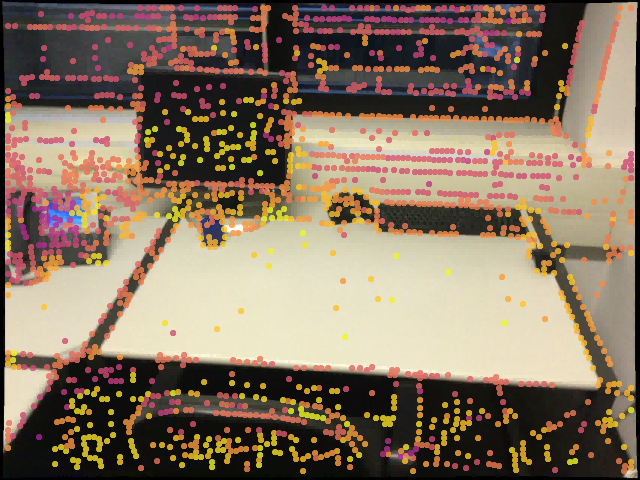} &
        \includegraphics[width=0.22\linewidth]{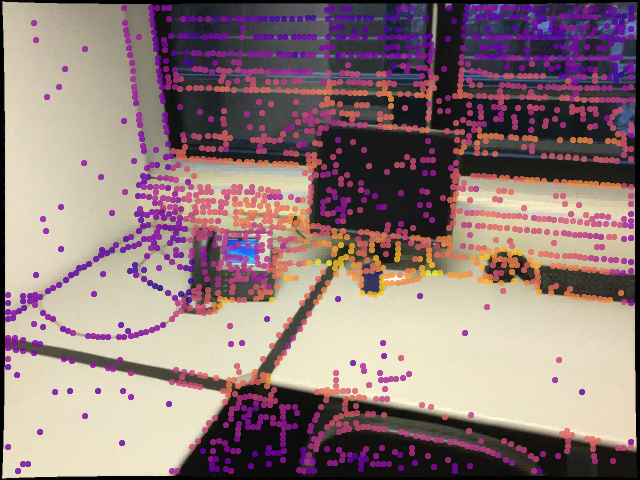} &
        \includegraphics[width=0.22\linewidth]{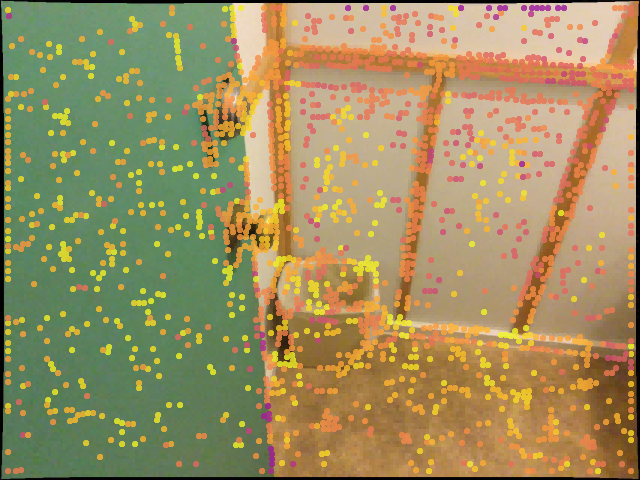} &
        \includegraphics[width=0.22\linewidth]{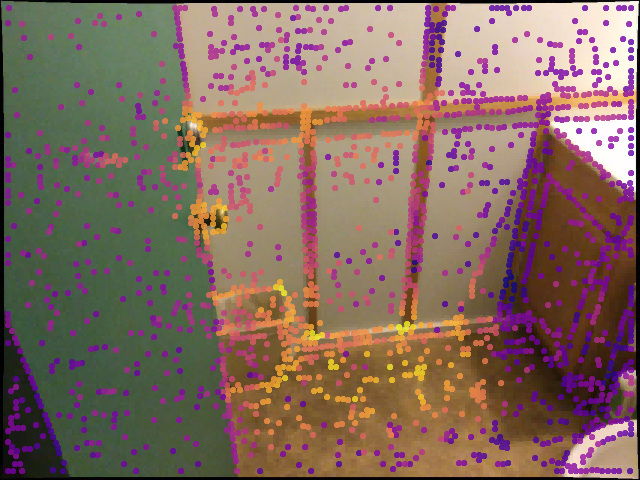} \\

        \scriptsize
        \rotatebox[origin=l]{90}{\hspace{3mm}Similarity} & 
    	\includegraphics[width=0.22\linewidth]{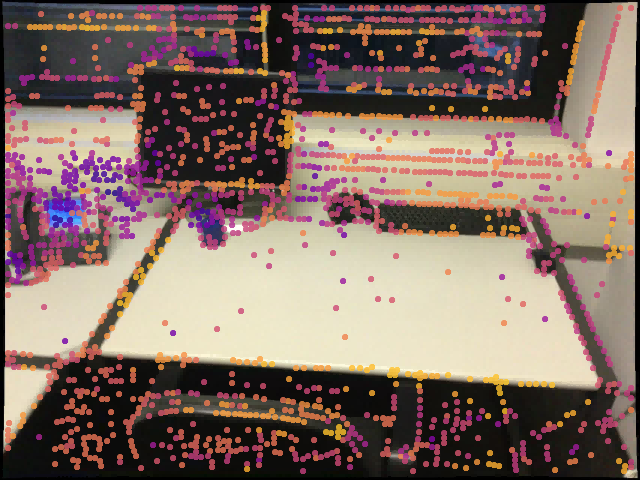} &
        \includegraphics[width=0.22\linewidth]{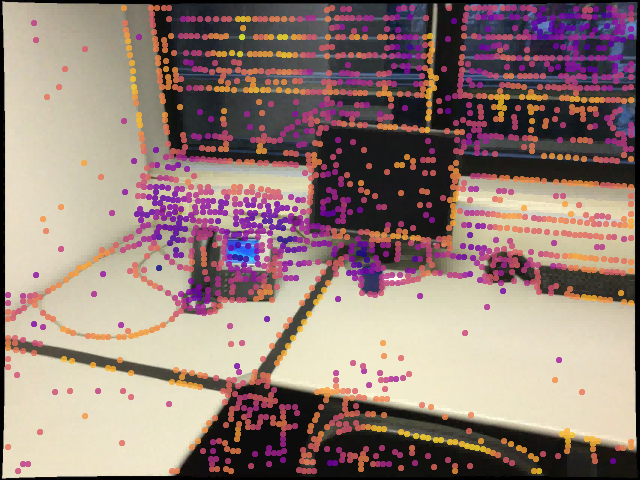} &
        \includegraphics[width=0.22\linewidth]{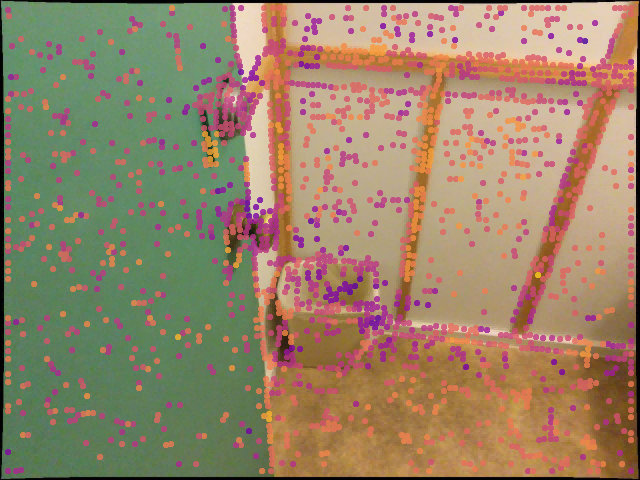} &
        \includegraphics[width=0.22\linewidth]{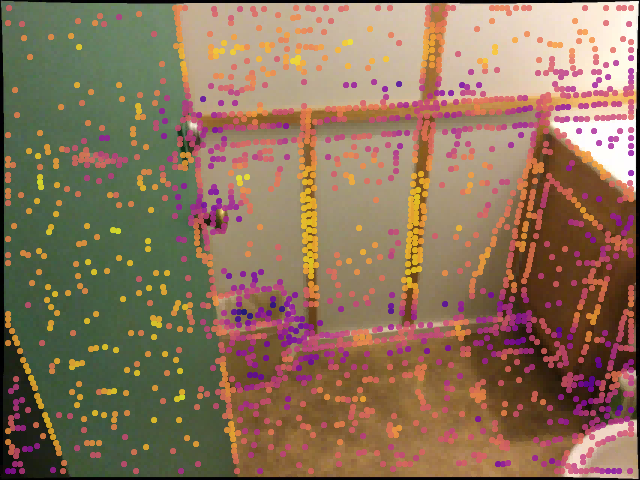} \\

        \scriptsize
        \rotatebox[origin=l]{90}{\hspace{3mm}Cross-attn.} & 
    	\includegraphics[width=0.22\linewidth]{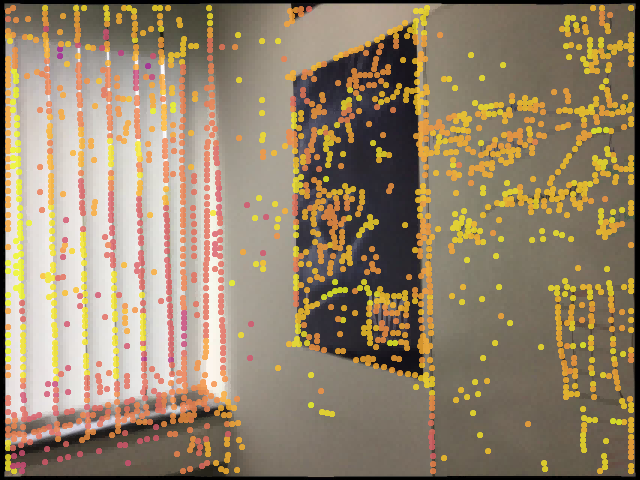} &
        \includegraphics[width=0.22\linewidth]{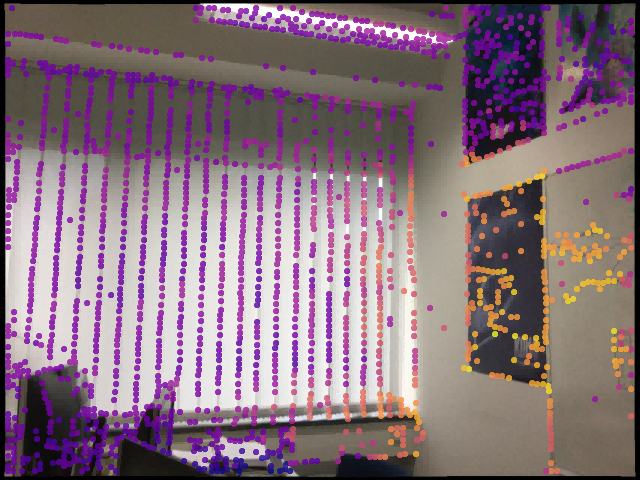} &
        \includegraphics[width=0.22\linewidth]{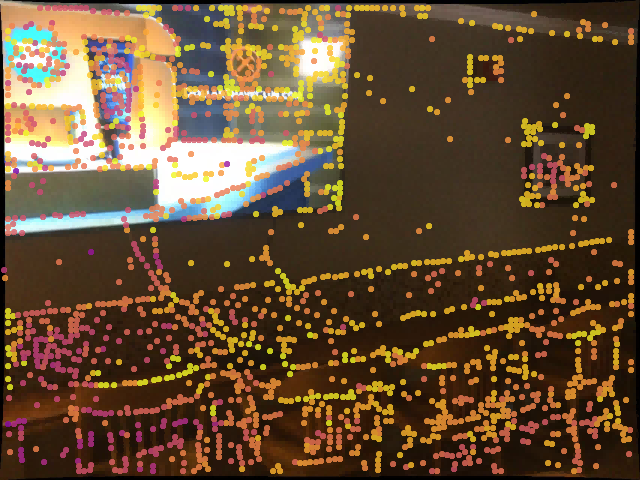} &
        \includegraphics[width=0.22\linewidth]{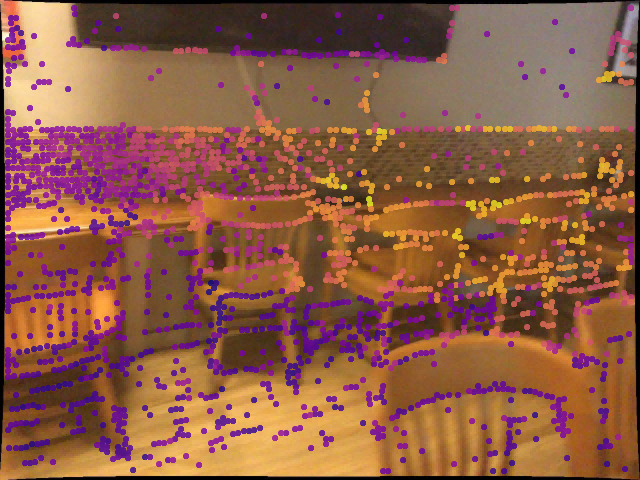} \\

        \scriptsize
        \rotatebox[origin=l]{90}{\hspace{3mm}Similarity} & 
    	\includegraphics[width=0.22\linewidth]{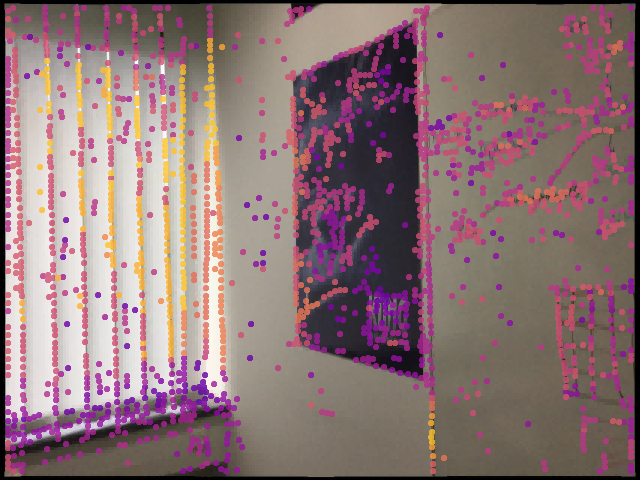} &
        \includegraphics[width=0.22\linewidth]{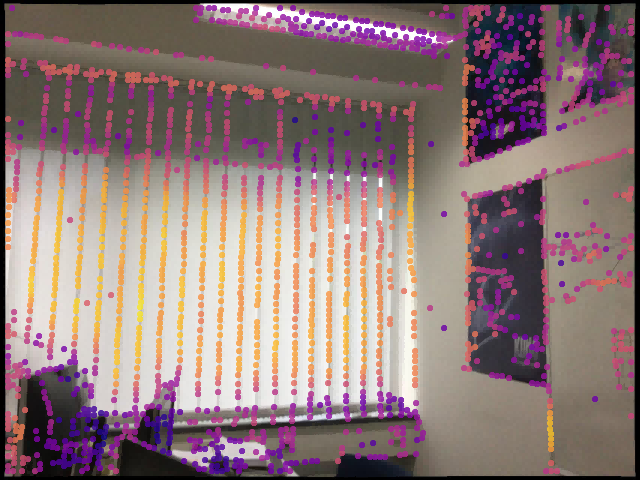} &
        \includegraphics[width=0.22\linewidth]{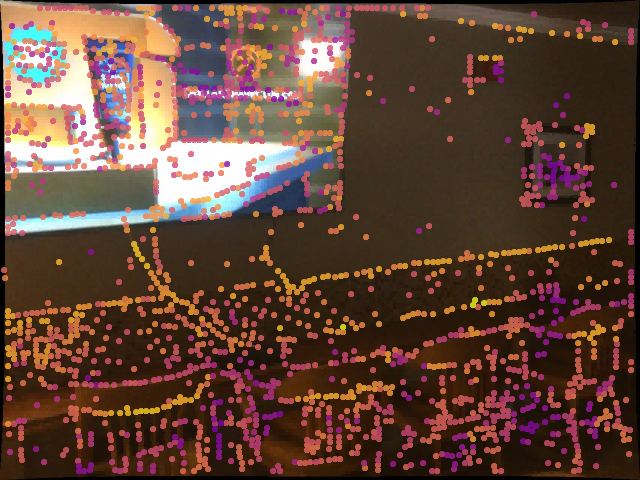} &
        \includegraphics[width=0.22\linewidth]{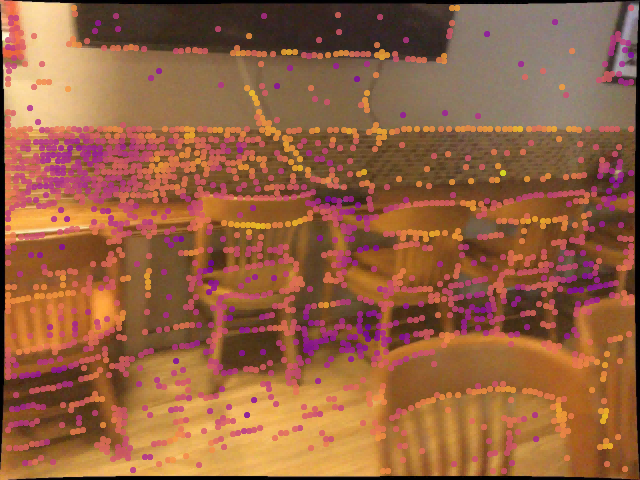} \\

        \scriptsize
        \rotatebox[origin=l]{90}{\hspace{3mm}Cross-attn.} & 
    	\includegraphics[width=0.22\linewidth]{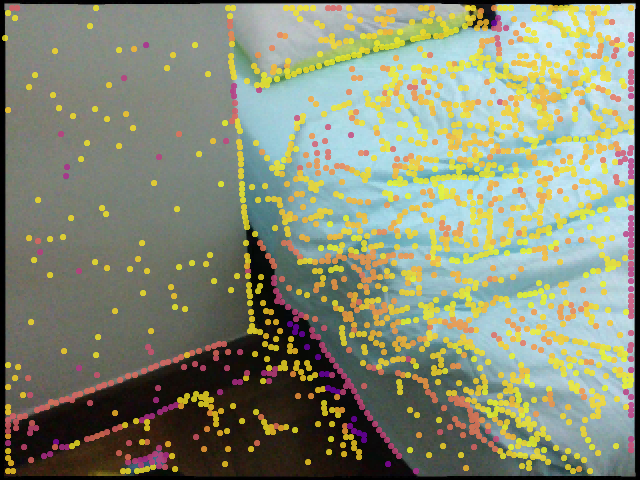} &
        \includegraphics[width=0.22\linewidth]{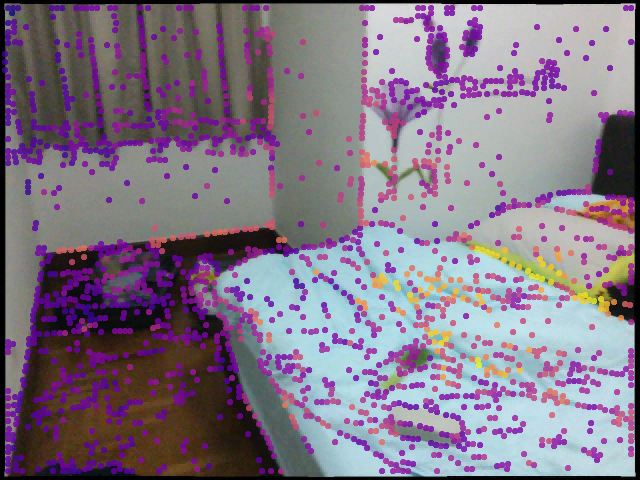} &
        \includegraphics[width=0.22\linewidth]{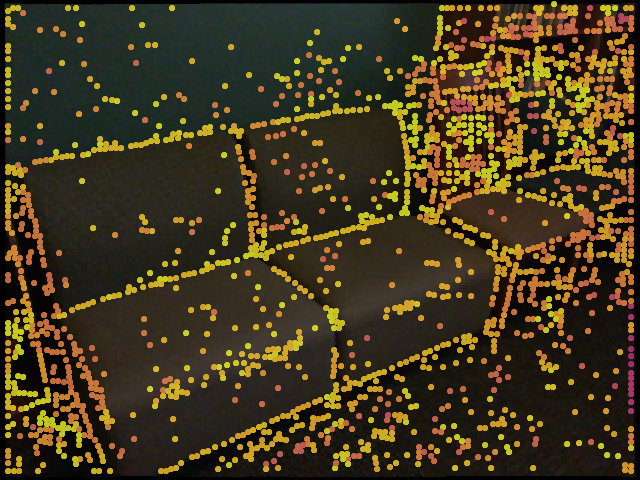} &
        \includegraphics[width=0.22\linewidth]{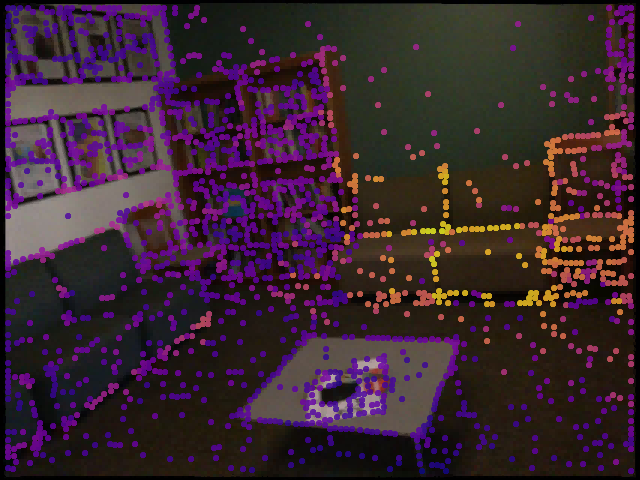} \\

        \scriptsize
        \rotatebox[origin=l]{90}{\hspace{3mm}Similarity} & 
    	\includegraphics[width=0.22\linewidth]{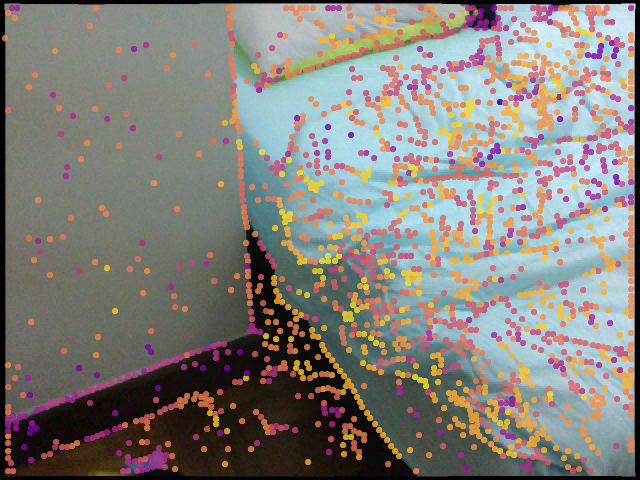} &
        \includegraphics[width=0.22\linewidth]{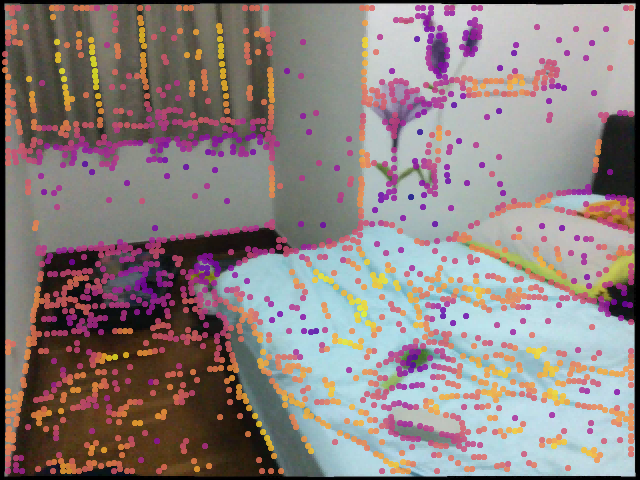} &
        \includegraphics[width=0.22\linewidth]{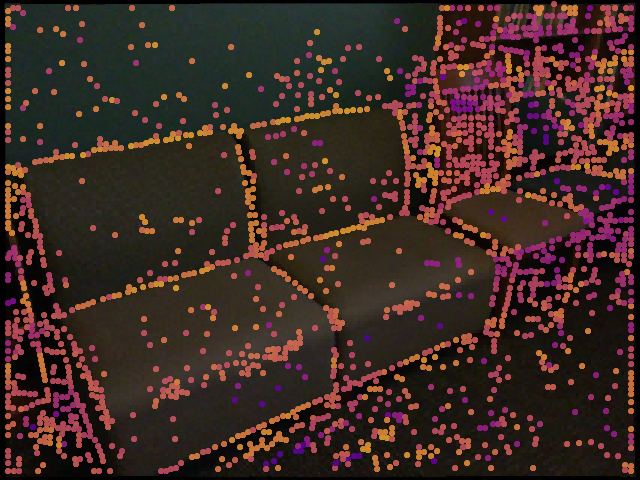} &
        \includegraphics[width=0.22\linewidth]{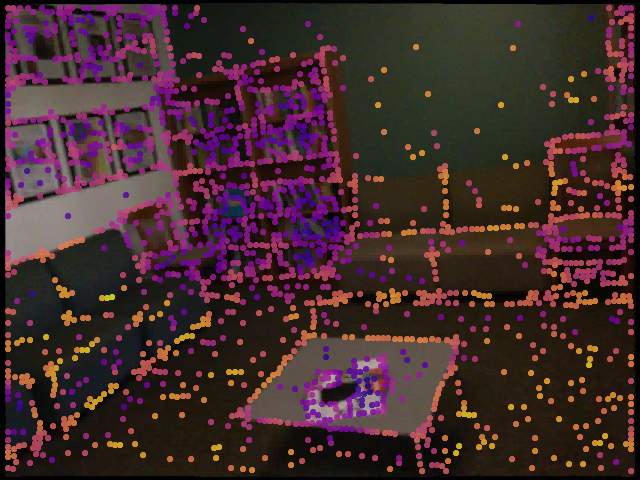} \\

    \end{tabular}
    \caption{\textbf{Visualization of the cross-attention scores and similarity matrices within SRPose.} Brighter dots represent higher values on the keypoints. High attention is shown to the overlapping areas. And the similarity matrices focus more on the informative keypoints on the edges and corners of the scenes.}
    \label{fig:vis-sim}
\end{figure}

\begin{figure}[htbp]
    \scriptsize
	\centering
    \begin{tabular}{cccc}
        Reference & Query & Reference & Query \\
        
        \includegraphics[width=0.22\linewidth]{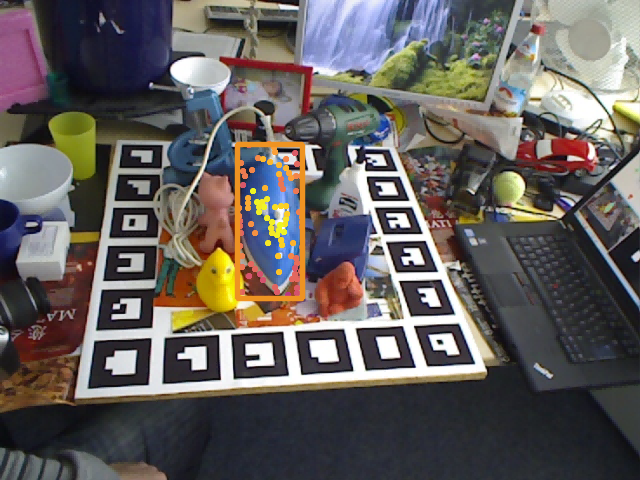} &
        \includegraphics[width=0.22\linewidth]{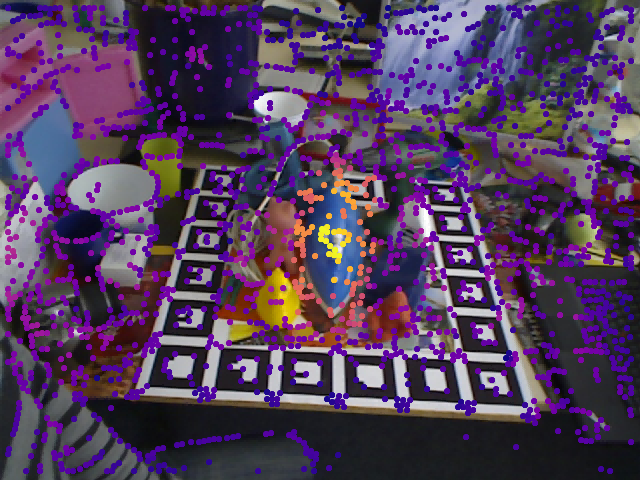} &
        \includegraphics[width=0.22\linewidth]{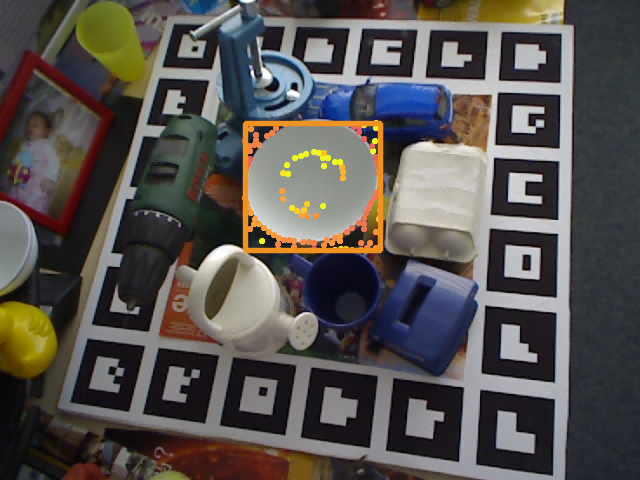} &
        \includegraphics[width=0.22\linewidth]{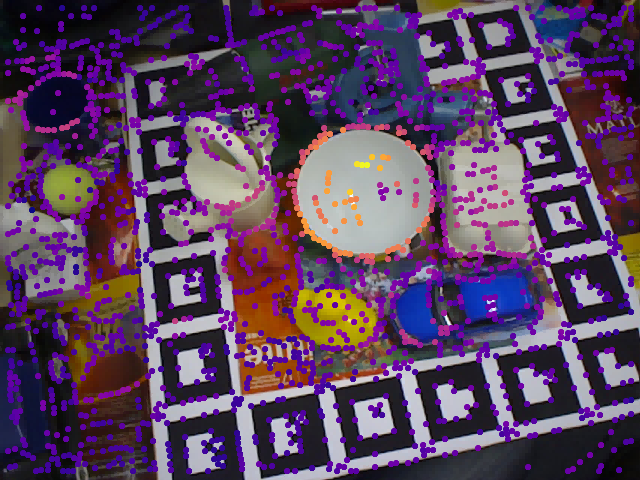} \\

        \includegraphics[width=0.22\linewidth]{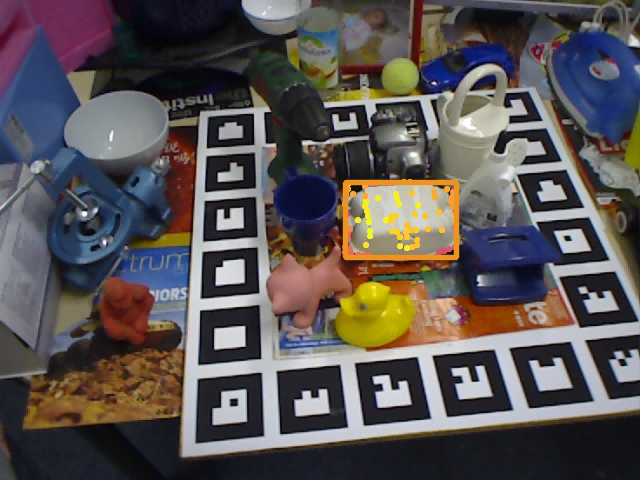} &
        \includegraphics[width=0.22\linewidth]{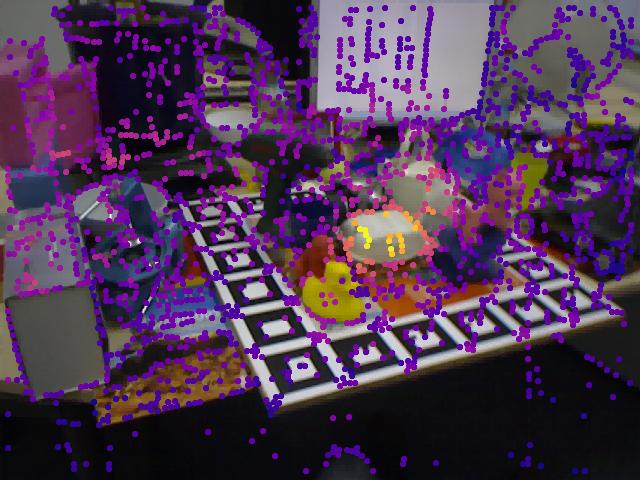} &
        \includegraphics[width=0.22\linewidth]{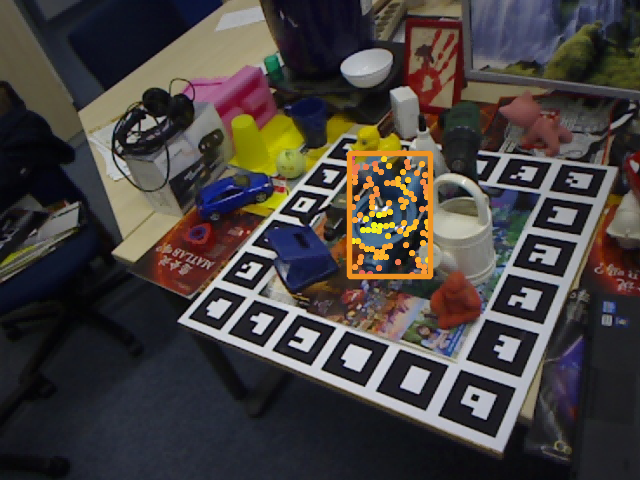} &
        \includegraphics[width=0.22\linewidth]{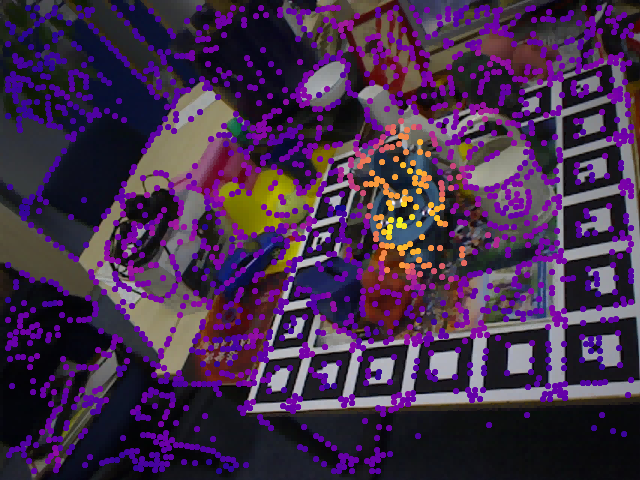} \\

        \includegraphics[width=0.22\linewidth]{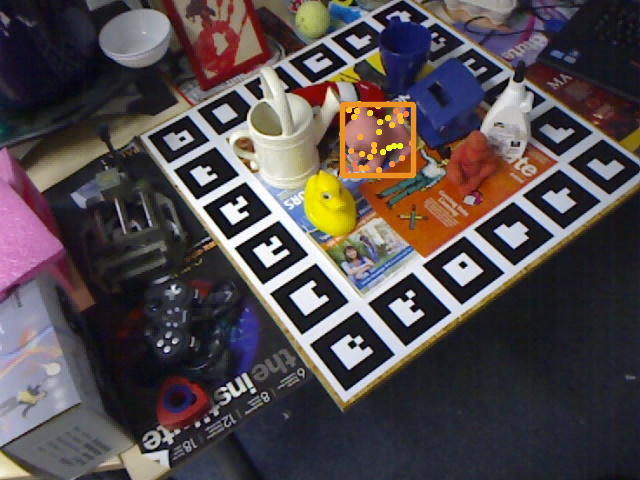} &
        \includegraphics[width=0.22\linewidth]{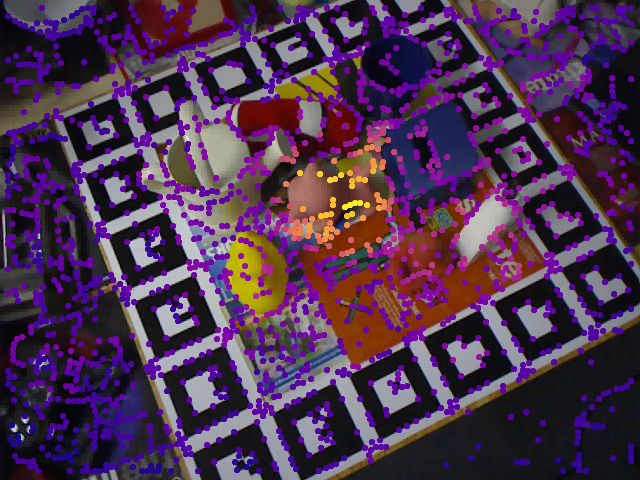} &
        \includegraphics[width=0.22\linewidth]{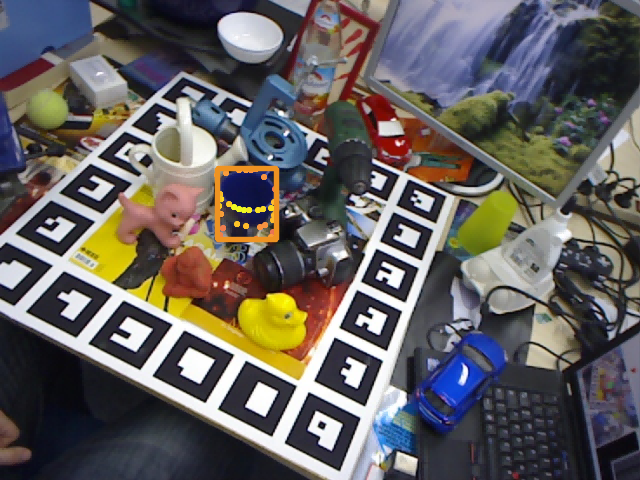} &
        \includegraphics[width=0.22\linewidth]{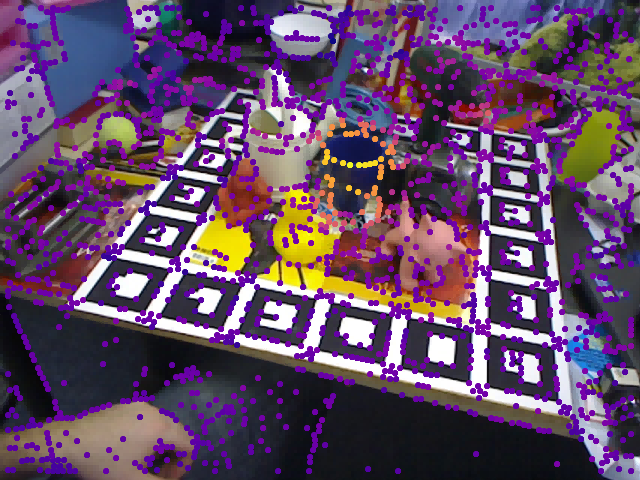} \\

        \includegraphics[width=0.22\linewidth]{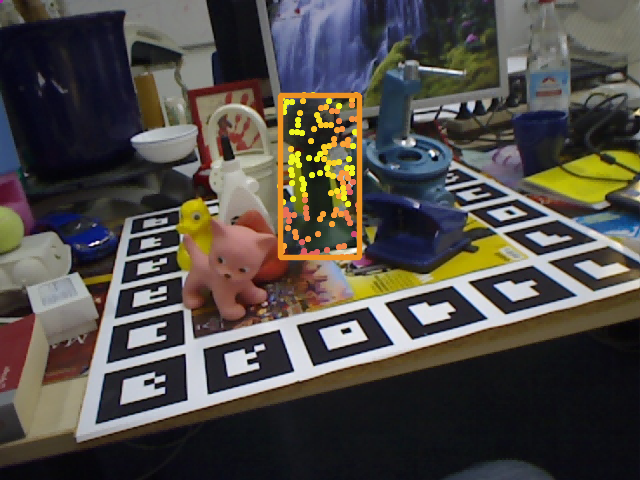} &
        \includegraphics[width=0.22\linewidth]{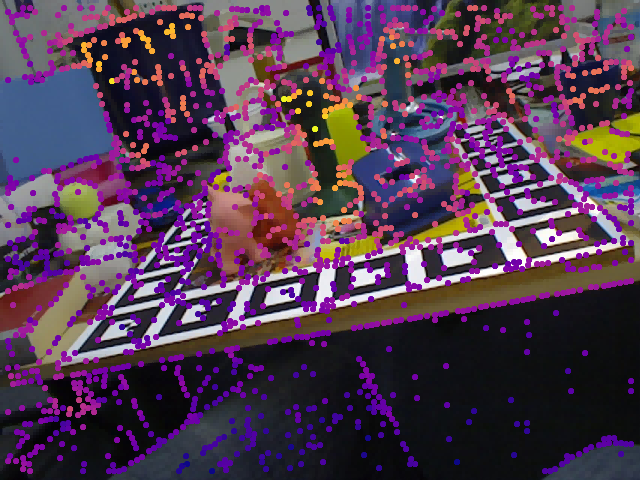} &
        \includegraphics[width=0.22\linewidth]{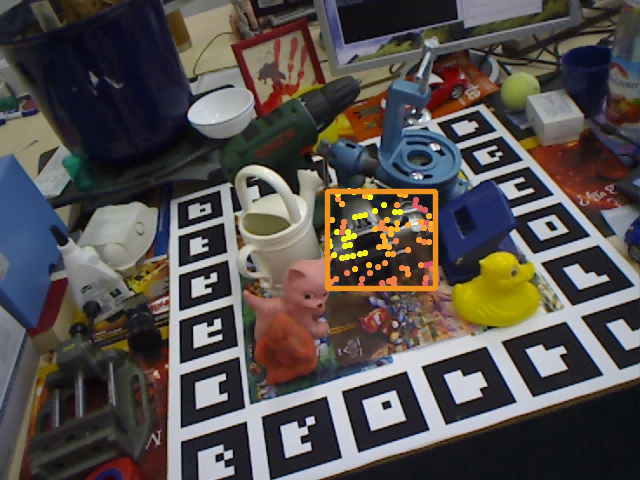} &
        \includegraphics[width=0.22\linewidth]{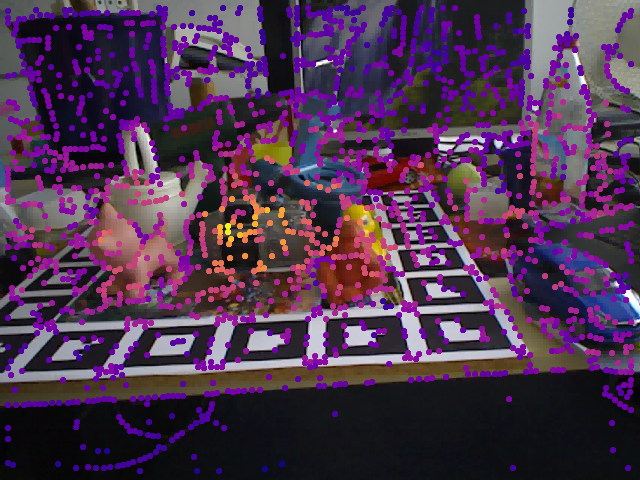} \\

        \includegraphics[width=0.22\linewidth]{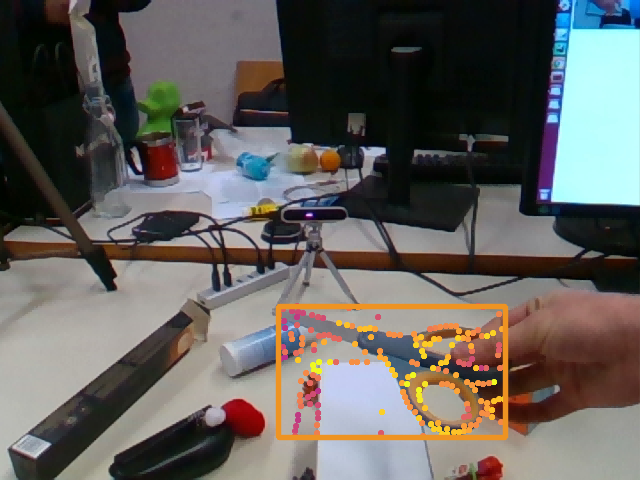} &
        \includegraphics[width=0.22\linewidth]{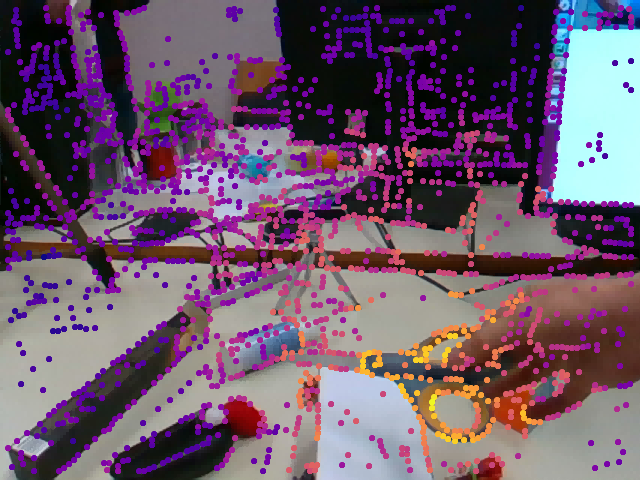} &
        \includegraphics[width=0.22\linewidth]{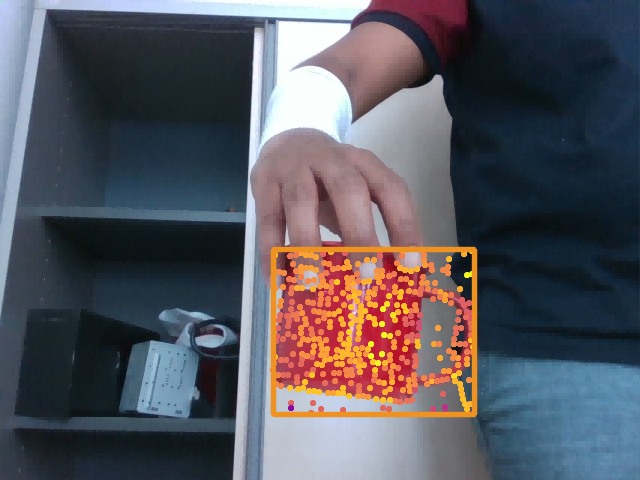} &
        \includegraphics[width=0.22\linewidth]{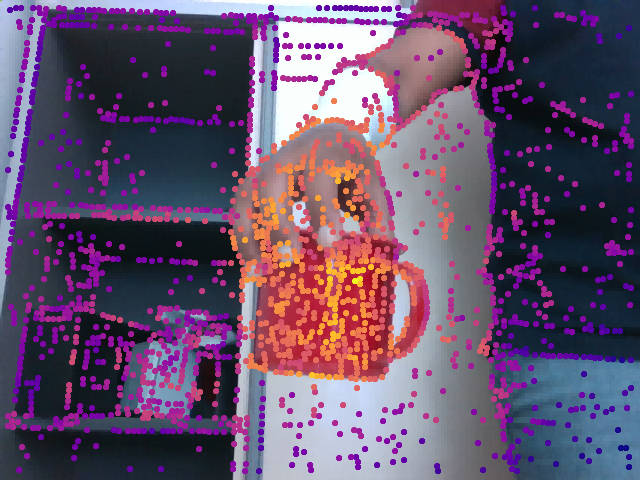} \\

        \includegraphics[width=0.22\linewidth]{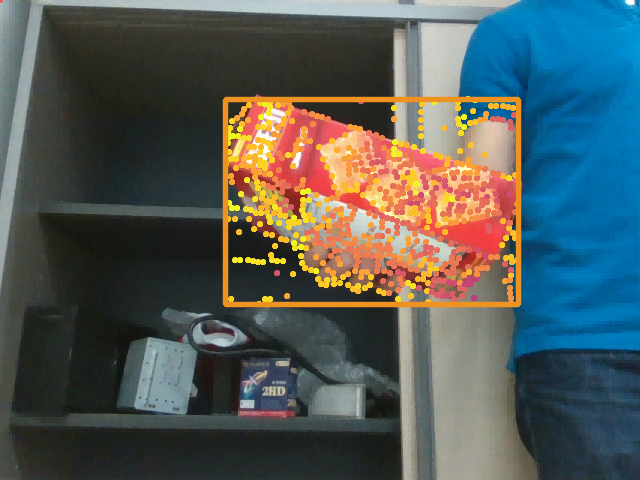} &
        \includegraphics[width=0.22\linewidth]{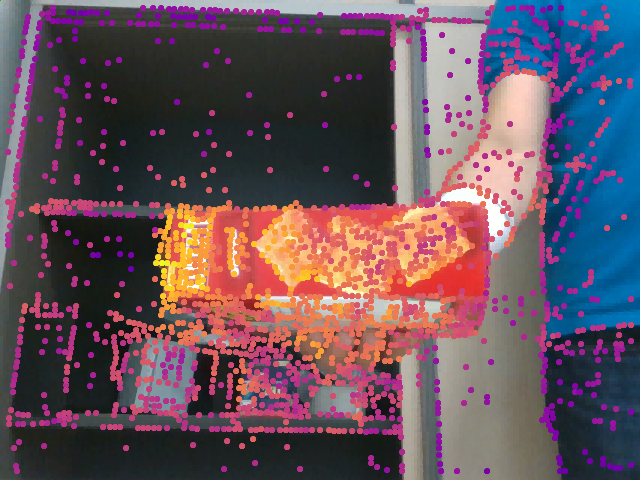} &
        \includegraphics[width=0.22\linewidth]{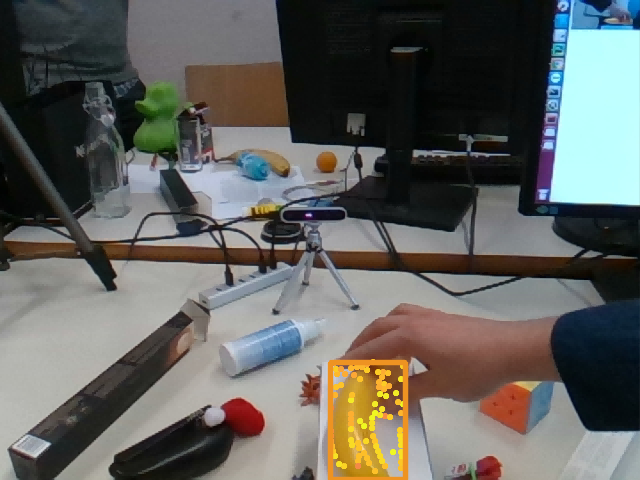} &
        \includegraphics[width=0.22\linewidth]{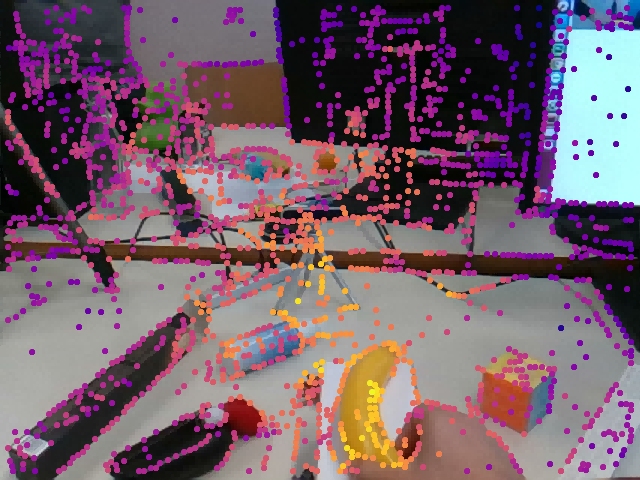} \\

        \includegraphics[width=0.22\linewidth]{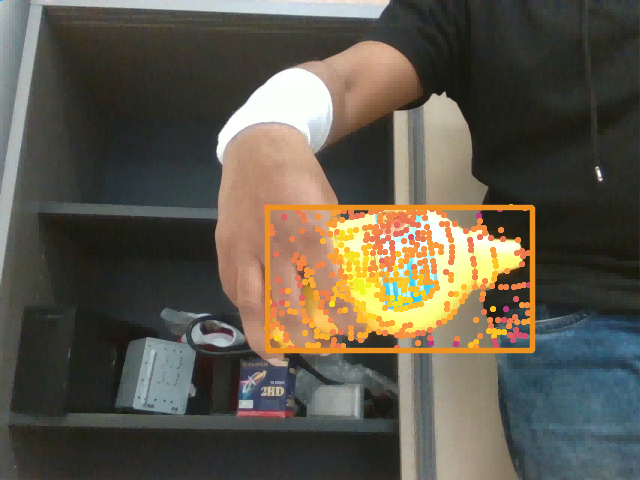} &
        \includegraphics[width=0.22\linewidth]{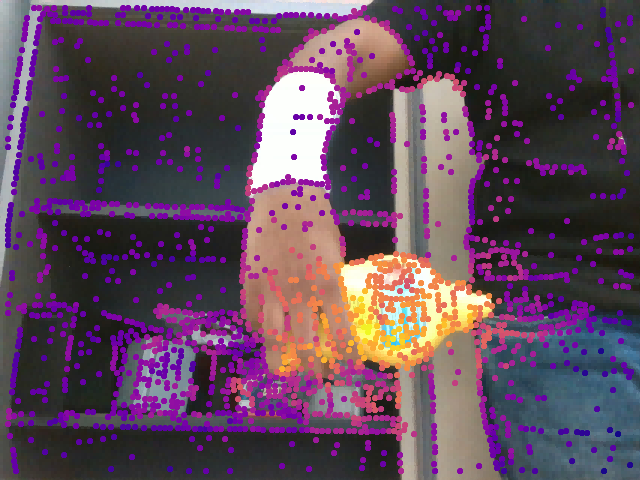} &
        \includegraphics[width=0.22\linewidth]{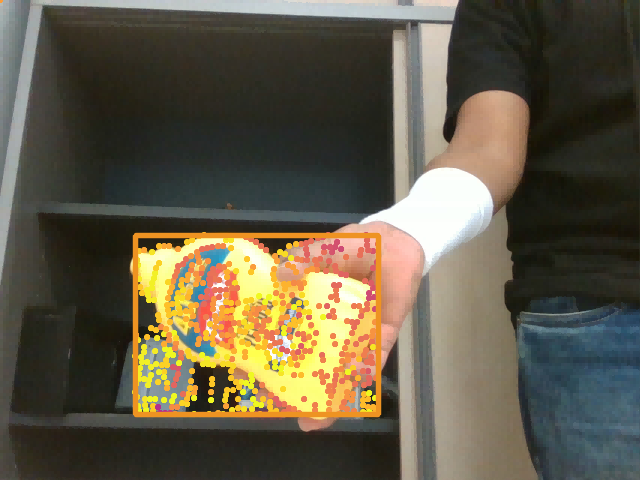} &
        \includegraphics[width=0.22\linewidth]{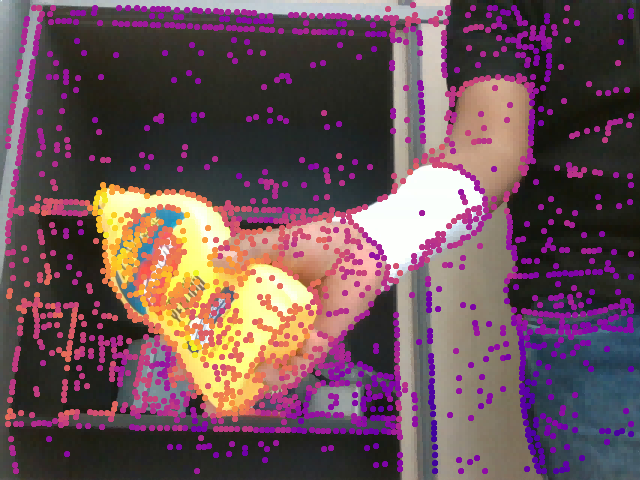} \\

        \includegraphics[width=0.22\linewidth]{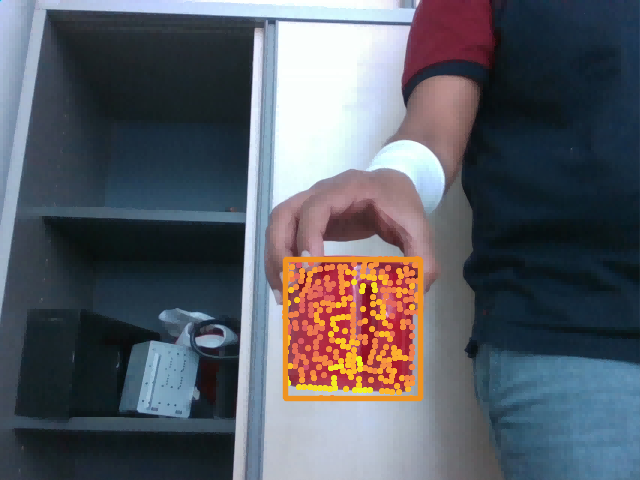} &
        \includegraphics[width=0.22\linewidth]{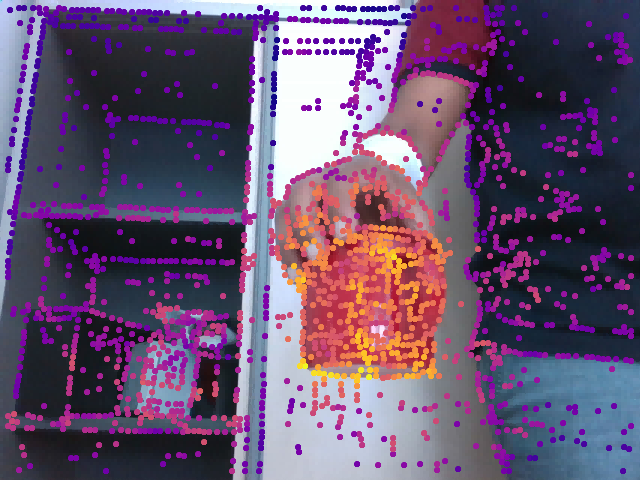} &
        \includegraphics[width=0.22\linewidth]{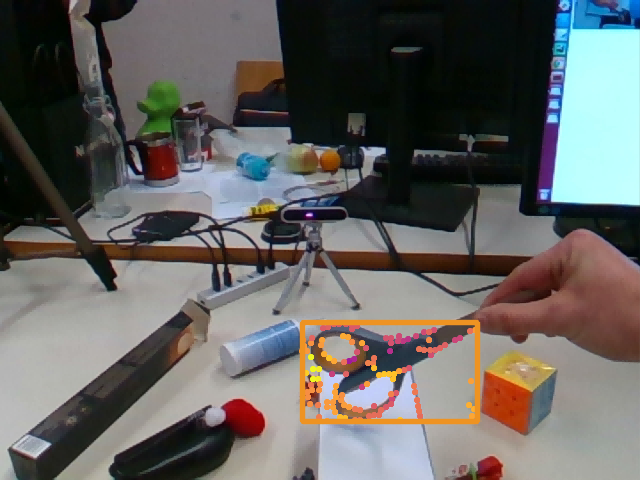} &
        \includegraphics[width=0.22\linewidth]{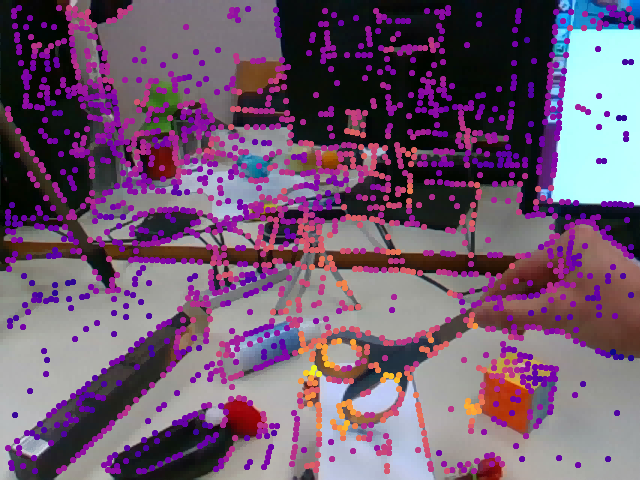} \\
        
    \end{tabular}
    \caption{\textbf{Visualization of the cross-attention scores in the object-to-camera scenario by SRPose.} SRPose utilizes an accessible user-provided object prompt in the reference view to automatically focus on the same target object in the query view.}
    \label{fig:vis-obj-attn}
\end{figure}

\vspace{-2mm}
We present more visualization results of the cross-attention scores in both camera-to-world and object-to-camera scenarios.
\cref{fig:vis-sim} visualizes more cross-attention scores across the two views in ScanNet~\cite{dai2017scannet}. Additionally, we also showcase the similarity matrices within SRPose for each case. As shown in the figure and discussed in the main text, the cross-attention scores exhibit high values on the overlapping areas of the scenes. Notably, the similarity matrices typically display high values near edges or corners in the images, guiding the cross-attention modules to focus on these informative keypoints, and facilitating the establishment of implicit correspondences.
\cref{fig:vis-obj-attn} visualizes the cross-attention scores in the object-to-camera scenarios on Linemod~\cite{hinterstoisser2013model} and HO3D~\cite{hampali2020honnotate}. With the object prompts in the reference images (represented by orange rectangles), SRPose automatically searches the implicitly corresponding keypoints on the same target object in the query view. The use of object prompts enables relative pose estimation in the object-to-camera scenario without object segmentation.

\vspace{-3mm}
\section{Limitations and Future Research}
\label{sec:limit}

\vspace{-3mm}
\subsubsection{Small pose transformation:}
SRPose typically underperforms in estimating small pose transformations, as mentioned in \cref{sec:megadepth}.
It's worth reminding that our framework establishes correspondences and implicitly solves the epipolar constraint equation using neural networks.
SRPose takes in encoded position information, which is used to compute the pose matrix through multiple layers.
However, in this process, the precise position information may be undermined, leading to a certain deterioration in pose estimation accuracy.
In contrast, traditional matcher-based approaches excel at matching local features in highly overlapping image pairs, which often correspond to small transformations.
By explicitly solving the constraint with minimal noise and outliers, these approaches can produce highly accurate results.
This explains why matcher-based baselines typically outperform neural network regressors including SRPose, on MegaDepth, a dataset consisting of small pose transformations, as shown in Table \ref{tab:megadepth}. 
Further quantitative analysis shows the under-performance is also due to SRPose's relatively lower precision at small pose error thresholds.
MegaDepth has a smaller average pose transformation than ScanNet, indicating similar views easier for matcher-based methods to match keypoints.
By direct solution using the epipolar constraint, matchers can yield lower errors on small pose transformations.
While regression methods, including SRPose, approximate solutions through neural networks, leading to lower precision at small thresholds.

\begin{figure}[htbp]
    \vspace{-4mm}
    
    \centering
    \begin{tabular}{cc}
        \includegraphics[height=0.33\linewidth]{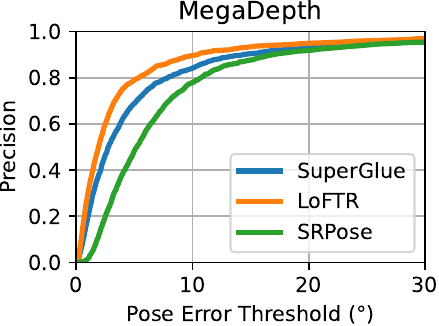} & \includegraphics[height=0.33\linewidth]{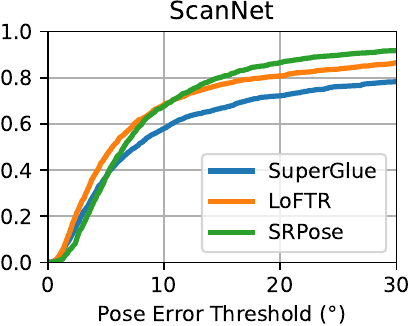}\\
    \end{tabular}
    \vspace{-2mm}
    
    \caption{Precision curve on MegaDepth [\textcolor{NavyBlue}{37}] and ScanNet [\textcolor{NavyBlue}{38}], using which the Area under the Curve (AUC) is computed.}

    \vspace{-8mm}
    \label{fig:curves}
\end{figure}

However, as shown in \cref{fig:curves}, SRPose achieves competitive precision or superior precision at larger thresholds compared to matchers.
Although SRPose underperforms on MegaDepth in the cumulative area under the precision curve (AUC), further analysis still exhibits its effectiveness in terms of precision. 
SRPose leverages the semantic information and connections to implicitly denoise outliers in such difficult cases, leading to higher accuracy.
One area for further research could be minimizing the loss of precision in position information during the propagation through neural network layers.

\vspace{-3mm}
\subsubsection{Relative object pose estimation:}
Estimating relative 6D object pose transformations has always been a challenge for both matcher-based and regressor-based approaches. Current video object pose tracking frameworks~\cite{wen2021bundletrack, wen2023bundlesdf} address the challenge by first estimating coarse poses between two adjacent frames with the assistance of a video object segmentation model.
Then the estimated coarse poses are optimized using global pose graph optimization to greatly improve the overall accuracy.
We believe that SRPose provides a new direction for mask-free object pose tracking. By further incorporating instance detection and global pose optimization, SRPose has the potential to enable pose tracking without relying on video object segmentation models, thereby achieving higher efficiency.

\clearpage

%% file: problem.tex
\section{Problem Definition}
\label{sec:problem}
\vspace{-2mm}
We aim to estimate the relative pose transformation between two views in both camera-to-world and object-to-camera scenarios. 
The estimated relative pose consists of a rotation matrix \(R\in\mathcal{SO}(3)\) and a translation vector \(t\in \mathbb{R}^3\), which maps the set of 3D world points \(P_{\mathrm{w}} \subseteq \mathbb{R}^3\) from the camera coordinate system of the first image \(I_1\) to the second image \(I_2\):
\begin{gather}
    P_1 = K_1 [\mathrm{I} | 0] P_{\mathrm{w}}, \label{pd1} \\
    P_2 = K_2 [R | t] P_{\mathrm{w}}, \label{pd2}
\end{gather}
where \(P_1\) (resp. \(P_2\)) \(\subseteq \mathbb{R}^3\) denotes the points \(P_{\mathrm{w}}\) projected onto the camera space of \(I_1\) (resp. \(I_2\)), and \(K_1, K_2 \in \mathbb{R}^{3\times3}\) denote the camera intrinsics of the two images. In the camera-to-world scenario, when given two overlapping images of a static scene, SRPose estimates the pose transformation of the camera from \(I_1\) to \(I_2\), in which, \(P_{\mathrm{w}}\) represents the set of static 3D points in the scene. On the other hand, in the object-to-camera scenario, when given two images containing multiple objects and an object prompt \(b\) in \(I_1\) that identifies the target object \(o\) to focus on, SRPose estimates the 6D object pose transformation of \(o\) between the two views. In this scenario, we assume the camera of the images is fixed, and \(P_{\mathrm{w}}\) represents the set of points on the moving target object \(o\).